\documentclass{article}

\usepackage[preprint]{neurips_2024}
\usepackage{multicol}

\usepackage[utf8]{inputenc} %
\usepackage[T1]{fontenc}    %
\usepackage{hyperref}       %
\usepackage{url}            %
\usepackage{booktabs}       %
\usepackage{amsfonts}       %
\usepackage{nicefrac}       %
\usepackage{microtype}      %
\usepackage{xcolor}         %
\usepackage{subcaption}
\usepackage{graphicx}
\usepackage{multirow}
\usepackage{adjustbox}
\usepackage[normalem]{ulem}
\usepackage{tabularray}
\usepackage{array}
\usepackage{ragged2e}
\usepackage{amssymb}
\usepackage{amsmath}
\usepackage{appendix}
\usepackage{bm}
\usepackage{wrapfig}

\title{FeDeRA: Efficient Fine-tuning of Language Models in Federated Learning Leveraging Weight Decomposition}

\author{%
  Yuxuan Yan, Shunpu Tang, Zhiguo Shi, Qianqian Yang \\
College of Information Science and Electronic Engineering, Zhejiang University\\
}

\begin{document}

\maketitle

\begin{abstract}
Despite their exceptional performance on various tasks after fine-tuning, pre-trained language models (PLMs) face significant challenges due to growing privacy concerns with data in centralized training methods. We consider federated learning (FL) to fine-tune PLMs in this paper, which trains PLM models on local clients and aggregates weights on a central server without sharing data. However, the substantial number of parameters in PLMs poses significant difficulties for client devices with limited communication and computational resources when fine-tuning all parameters. One promising solution to this is to exploit parameter-efficient fine-tuning (PEFT) into federated learning, which trains a much smaller set of parameters than full parameter fine-tuning (FFT). Although remarkably improving training efficiency, PEFT methods may lead to degraded performance especially when data across different clients are non i.i.d. distributed in FL, as revealed by experimental results. To overcome this, we propose FeDeRA, which extends and improves a widely used PEFT method, i.e., low-rank adaption (LoRA), to a FL setting. FeDeRA follows LoRA by decomposing the weight matrices of the PLMs into low-rank matrices, which allows for more efficient computation and parameter updates during fine-tuning. Different from LoRA which simply initializes these low-rank matrices by random sampling or zeros, the proposed FeDeRA initializes these matrices by the results of performing singular value decomposition (SVD) on the pre-trained weight matrices. Extensive experiments across various tasks and datasets show that FeDeRA outperforms the considered PEFT baselines and is comparable to or even surpasses FFT method within the FL setting in terms of task performance. Moreover, FeDeRA requires only 1\% trainable paramentes compared to FFT, significantly reducing training time costs by more than 90\% to achieve the same task performance level. The experimental results also highlight the robustness of FeDeRA against data heterogeneity, as it maintains stable task performance even as data heterogeneity increases.

\end{abstract}

\section{Introduction}
\label{sec:introduction}
Pre-trained language models (PLMs) have achieved state-of-the-art (SOTA) performances across various NLP tasks, including natural language understanding\cite{kenton2019bert,sanh2019distilbert,liu2019roberta}, text generation\cite{brown2020language, zeng2022glm, jiang2023mistral}, and question answering. However, training these models from scratch demands a significant amount of resources\cite{narayanan2021efficient}, making it unattainable for most. Consequently, fine-tuning models for specific tasks has emerged as the primary method of leveraging large language models. This process typically involves training a pre-trained model on a significantly smaller dataset rather than the original training set. While fine-tuning in a centralized manner is typically preferred, aggregating all data onto a single device raises concerns about data privacy, thus making centralized training increasingly challenging. 

Federated learning (FL) \cite{konevcny2016federated,mcmahan2017communication,wang2022asynchronous,tang2022computational} has emerged as a promising approach in machine learning to address data privacy concerns by training a model collaboratively across decentralized clients without sharing raw local data. In FL, clients periodically compute and send model information, such as parameters or gradients, to a central server for aggregation, resulting in a global model. However, fine-tuning PLMs in an FL setting encounters significant challenges. Firstly, FL requires frequent exchange of model parameters or gradients between the server and clients. The massive number of parameters in PLMs, often in the tens or hundreds of billions, leads to substantial communication overheads. Additionally, devices involved in FL often communicate over bandwidth-limited networks, causing significant delays during data transmission and reducing training efficiency. Moreover, fine-tuning language models demands substantial memory and computational resources, which many edge devices may not adequately meet.

Recently, parameter-efficient fine-tuning (PEFT) techniques such as BitFit\cite{zaken2022bitfit}, adapter tuning\cite{narayanan2021efficient,pfeiffer2021adapterfusion}, prefix tuning\cite{li2021prefix} and LoRA\cite{hu2021lora} have garnered significant attention for their ability to update only a small portion of a pre-trained model's parameters, thus offering advantages in memory and computational efficiency. Existing works have demonstrated that these methods match or even surpass the performance of traditional full-parameter fine-tuning (FFT) methods in centralized training. Adopting PEFT in the FL setting holds promise in addressing the aforementioned challenges by effectively reducing the number of parameters transmitted between clients and the server, as well as lowering the memory and computation costs associated with model training on each client. 

\begin{figure}[!t]
\centering
\begin{subfigure}{.46\textwidth}
  \centering
  \includegraphics[width=1\linewidth]{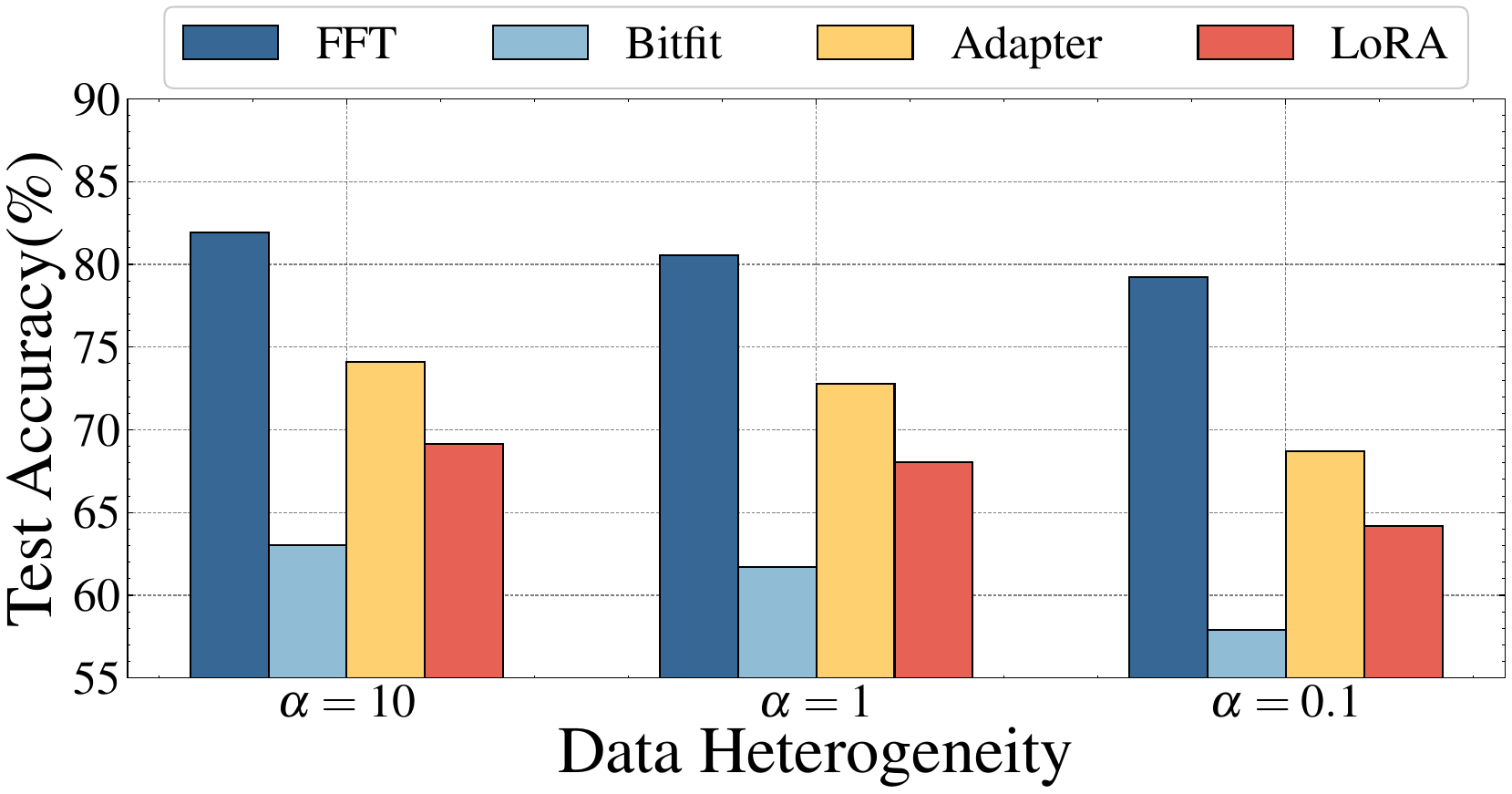}
  \caption{Comm. rounds 200}
\end{subfigure}%
\begin{subfigure}{.46\textwidth}
  \centering
  \includegraphics[width=1\linewidth]{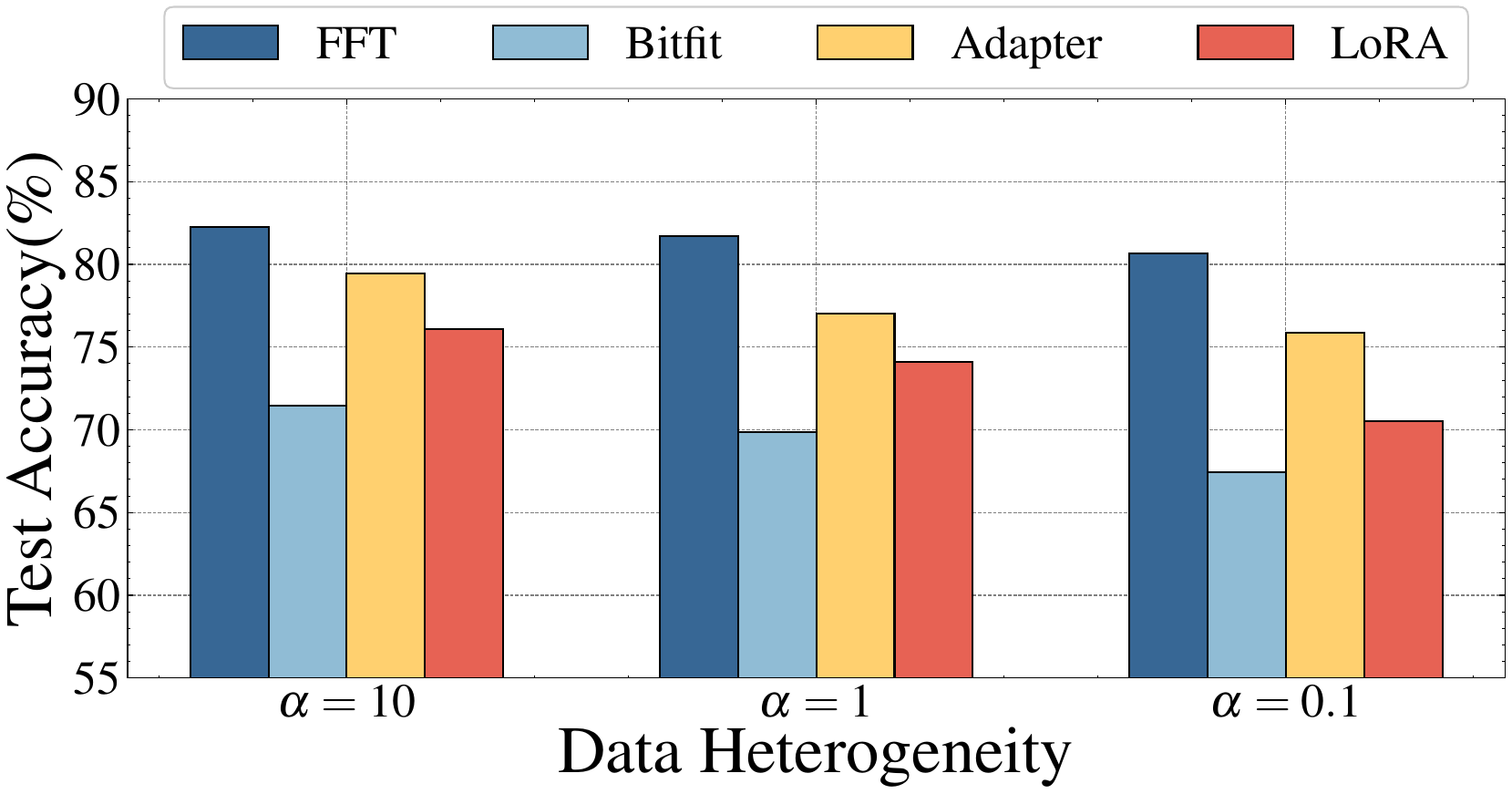}
  \caption{Comm. rounds 500}
\end{subfigure}
\caption{The performance of adopting PEFT methods within an FL setting is evaluated at varying levels of data heterogeneity using DistilBERT fine-tuned on the 20Newsgroup dataset. Heterogeneous data is generated based on a Dirichlet distribution, where the hyperparameter $\alpha$ determines the degree of data heterogeneity—a lower $\alpha$ value indicates higher data heterogeneity\cite{lin2022fednlp}.}
\label{peft in fl}
\end{figure}

In this work, we investigate PEFT methods within a FL setting and find that the heterogeneity of clients' data results in reduced performance and slower convergence rates for PEFT. As shown in \autoref{peft in fl}, the performance gap between  PEFT methods and FFT widens as data heterogeneity increases. To address this issue, we propose FeDeRA, building upon LoRA's approach of decomposing pre-trained model weight matrices into low-rank matrices. FeDeRA innovatively leverages singular value decomposition (SVD) on pre-trained weight matrices to extract principal components for initializing low-rank matrices, i.e., the adaptors, while preserving and freezing remaining components in the original matrices. This simple yet effective solution addresses the remarkable performance decline observed in LoRA within federated settings, particularly when confronted with significantly non-IID data across clients. Experimental outcomes reveal that FeDeRA achieves comparable or superior task performance to FFT across various tasks, significantly surpassing other PEFT methods. Moreover, FeDeRA reduces training time by over 95\% compared to FFT, while maintaining consistent task performance. Furthermore, experimental results underscore the robustness of FeDeRA against data heterogeneity, exhibiting much lower performance degradation compared to other PEFT methods.

\vspace{-0.2cm}
\section{Related Works}
\vspace{-0.2cm}
\subsection{Federated Learning with Non-IID Data}
Federated learning (FL) is widely adopted for distributed learning, especially in tasks requiring significant privacy considerations.  However, it suffers unneglectable performance degradation when data across different clients are non-independently and Identically Distributed (non-IID). Research efforts have surged to mitigate the effects of non-IID data on FL performance\cite{Non_IID_ICLR}. Some studies address this issue directly on the training data of each client through methods such as sharing a small portion of local data\cite{zhao2018federated}, sharing model outputs\cite{itahara2021distillation}, enhancing data with external datasets\cite{jeong2018communication},  generating new data from encoded and protected data of other clients\cite{shin2020xor}, or selecting data carefully\cite{cai2020dynamic,wang2020optimizing}. Other approaches involves carefully designing the model training process, such as using adaptive learning rates on clients\cite{ma2021fedsa} or employing a control variable to prevent excessive drift in local updates\cite{karimireddy2020scaffold}. Additionally, some studies have combined FL with other algorithms like Meta-Learning\cite{jiang2019improving,zhang2021fedpd}, Lifelong Learning\cite{shoham2019overcoming,kopparapu2020fedfmc}, and Knowledge Distillation\cite{itahara2021distillation,zhang2022fine}. Existing research suggests that using pre-trained models in FL, rather than training models from scratch, can effectively mitigate the impact of non-IID data on performance\cite{weller2022pretrained,nguyen2022begin,chen2022importance}. However, as discussed in Section~\ref{sec:introduction}, the substantial computational demands and communication overheads associated with PLMs are significant challenges. Thus, optimizing the fine-tuning of PLMs within an FL setting for both computation and communication is urgently needed.

\subsection{PEFT in Federated Learning}
PEFT methods aim to freeze most or all of the weights in PLMs, fine-tuning only a small subset or newly introduced weights. This approach reduces computational resource requirements and significantly decreases communication overhead. Some studies have incorporated various PEFT methods in FL, demonstrating their efficiency in training and evaluating the comparative advantages and disadvantages of different PEFT approaches in terms of performance, resource demands, and privacy aspects\cite{sun2022exploring,chen2022fedtune,zhang2023fedpetuning}. Additionally, some other studies focus on exploring the potential of PEFT techniques within FL, such as using LoRA for enhanced heterogeneous personalized learning in FL\cite{yi2023fedlora,lu2024hyperflora}, and incrementally enhancing adapter configurations during FL training to accelerate convergence\cite{cai2023efficient}. However, it has been increasingly recognized that PEFT methods experience significant performance degradation when dealing with highly non-IID data distributions in FL\cite{zhang2023fedpetuning,babakniya2023slora}. Despite this growing awareness, most prior research has either overlooked this critical issue or failed to evaluate PEFT methods under severe non-IID data conditions.

\section{Preliminaries}

\subsection{Federated Learning}
FL is a distributed approach to collaboratively train a global model across multiple clients, denoted by $\mathcal{M}=[1,2\ldots M]$, governed by a central server. We use $\mathcal{D}_m$ to denote the local training set at client $m\in \mathcal{M}$, and $|\mathcal{D}_m|$ to represent the number of training samples in $\mathcal{D}_m$. The goal of FL is to optimize the global model parameters $\omega$ by minimizing the following objective function:
\begin{equation}
\min_{\omega}  F(\omega) \triangleq \sum_{m=1}^{M}p_{m}F_{m}(\omega),
\end{equation}
where $p_m\geq 0$ is the weight assigned to the client $m$, with $\sum_{m=1}^{M}p_m=1$. Typically, $p_m = \frac{|\mathcal{D}m|}{\sum{m=1}^{M} |\mathcal{D}_m|}$, ensuring that each training sample contributes equally to the optimization process. Additionally, $F_{m}(\cdot)$ represents the local objective, which can be expressed as:
\begin{equation}
F_{m}(\omega) \triangleq \mathcal{L}_{m}(\omega ;\mathcal{D}_m),
\end{equation}
where $\mathcal{L}(\cdot)$ is the specific local loss function at client $m$.

One of the most commonly used methods to solve this problem is the well-known FedAvg algorithm \cite{mcmahan2017communication}. Specifically, in each training round, the server selects a subset of clients, denoted as $\mathcal{S}\in \mathcal{M}$, to participate in the FL training. $S=|\mathcal{S}|$ represents the number of selected clients. For a selected client $s\in \mathcal{S}$, the local training process begins by initializing its local model using the global model parameters from the previous round, i.e., $\omega_{s} = \omega^{t-1}$. The client then updates its local model parameters $\omega_{s}$ using its local dataset according to the following rule:
\begin{equation}
    \omega_{s} \leftarrow \omega_{s}- \eta \nabla \mathcal{L}_{s}(\omega_{s}; \mathcal{D}_{s}),
\end{equation}
where $\eta$ is the learning rate. This process is repeated for a specified number of iterations, resulting in the updated local weights $\omega_{s}^t$.

The selected clients then upload their updated weights, $\omega_{s}^t$, to the server. The server aggregates these received model parameters to update the global model parameters as follows:
\begin{equation}
\omega^{t} \leftarrow \frac{\sum_{s \in \mathcal{S}} |\mathcal{D}_s| \omega^{t}_{s}}{\sum_{s \in \mathcal{S}} |\mathcal{D}_s|}.
\end{equation}
However, it is important to note that in real-world scenarios, the data across clients are often highly non-IID. This can lead to significant divergence among the local model parameters from different clients, making it challenging for the global optimization to converge.

\subsection{Low-Rank Adaptation}
\begin{wrapfigure}{R}{0.46\textwidth}
  \vspace{-1cm}
  \setlength{\belowcaptionskip}{-0.2cm}
\begin{center}
    \begin{subfigure}{.46\textwidth}
        \includegraphics[width=\linewidth]{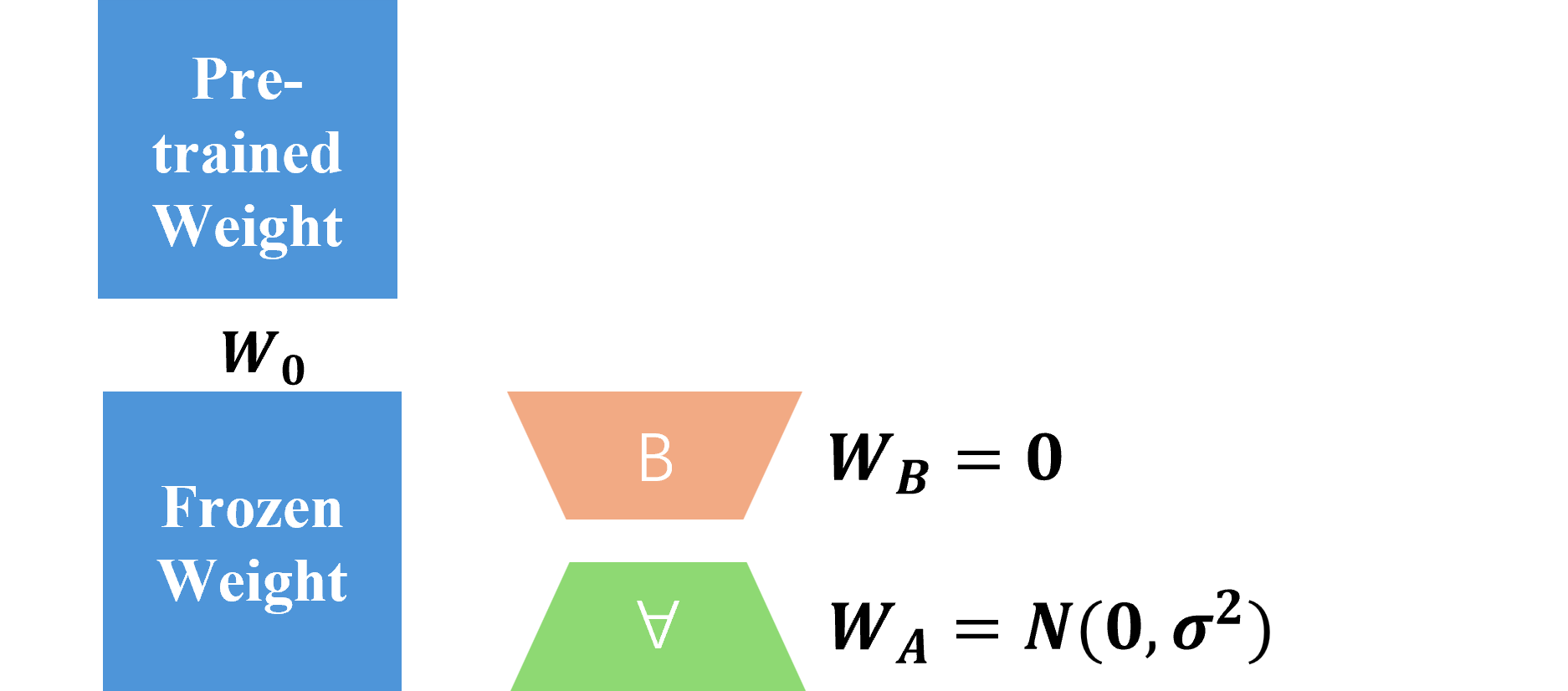}
        \caption{LoRA}
        \label{subfig:lora}
    \end{subfigure} 
    \\ 
    \vspace{0.25cm}
    \begin{subfigure}{.46\textwidth}
        \includegraphics[width=\linewidth]{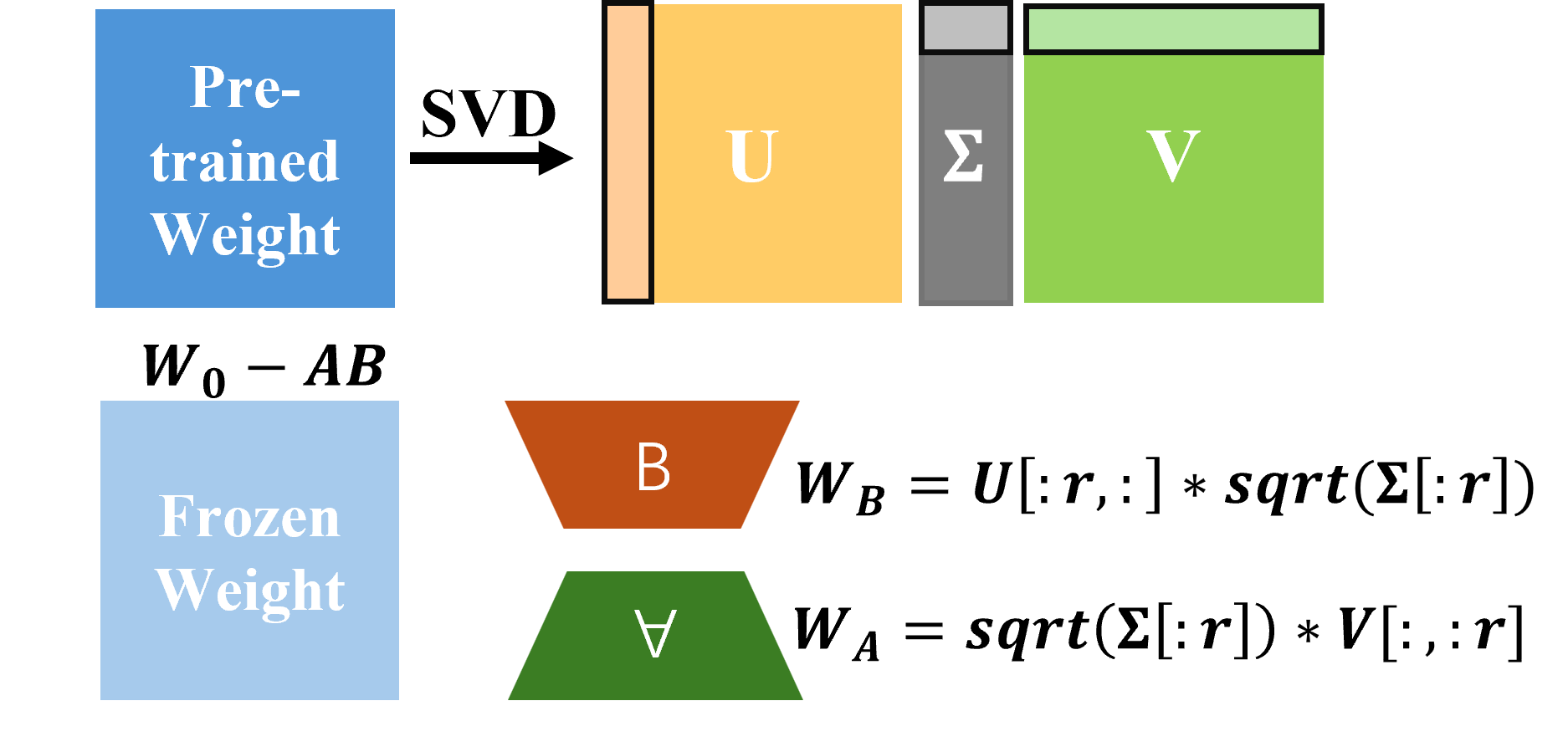}
        \caption{FeDeRA}
        \label{subfig:federa}
    \end{subfigure}\hfill
    \caption{Illustration of LoRA and FeDeRA.}
\end{center}
\end{wrapfigure}
LoRA is one of the most widely used PEFT method of PLMs.  This method is based on the observation that when PLMs are adapted to specific tasks, their weight update matrices have a lower intrinsic dimension than their dimensions. This implies that projecting these matrices into a lower dimension does not lead to significant information loss. The key idea behind LoRA is to avoid fine-tuning an entire pre-trained weight matrix $W_0 \in \mathbb{R}^{d\times k}$. Instead, as illustrated in Figure. \ref{subfig:lora}, LoRA introduces a weight update mechanism based on matrix decomposition:  $W_0+\Delta W=W_0+AB$. Here, $B\in \mathbb{R}^{d\times r}$ and $A\in\mathbb{R}^{r\times k}$ are trainable matrices, with rank $r\ll \min\{d,k\}$. During training, matrix $A$ is initialized by sampling from a random Gaussian distribution, and matrix $B$ is initialized to zeros, ensuring that initially $AB=0$. The matrix $W_0$ remains frozen throughout the training, while $AB$ are updated during the fine-tuning process. The output of a layer after implementing LoRA is given by:
\begin{equation}
h = W_0x+\frac{\beta}{r}BAx,
\end{equation}
where $x$ is the input, and $\beta$ is designed to eliminate the need of re-adjusting hyperparameters when changing $r$.

\section{Methods}

\subsection{FeDeRA}
Although LoRA has shown superior performance across various tasks in FL, it still suffers from a slower convergence rate and degraded performance when the data across clients is highly non-IID. To address this issue, we introduce FeDeRA, a novel FL approach for fine-tuning PLMs based on LoRA that performs SVD on a pre-trained weight matrix $W_0$ to initialize the corresponding weight parameters of $A$ and $B$. As illustrated in \autoref{subfig:federa},  the procedure begins with the SVD decomposition of $W_0$, which can be expressed as:
\begin{equation}
    W_0\stackrel{\text{SVD}}{=} U\Sigma V,
\end{equation}
where
\begin{align*}
     U = [u_1, u_2, \ldots, u_{d}] \in \mathbb{R}^{d\times d},  \\  V = [v_1, v_2, \ldots, v_{k}]^T \in \mathbb{R}^{k \times k}
\end{align*}
denotes the matrix of left and right singular vectors with rank $r$, respectively, and 
\begin{align}
     \Sigma = \text{diag}(\sigma_1, \sigma_2, \ldots, \sigma_{k}) \in \mathbb{R}^{d \times k},
\end{align}
is the diagonal matrix containing the singular values.
We then initialize the weight matrices $A$ and $B$ as follows with rank $r$:
\begin{equation}
A=\sqrt{\Sigma[:r]}V[:,:r]=[\sqrt{\sigma_1}v_1, \sqrt{\sigma_2}v_2,\ldots, \sqrt{\sigma_r}v_r]^T
\end{equation}
\begin{equation}
B=U[:r,:]\sqrt{\Sigma[:r]}=[\sqrt{\sigma_1}u_1,\sqrt{\sigma_2}u_2,\ldots,\sqrt{\sigma_r}u_r].
\end{equation}
To maintain the original model output unchanged, we adjust and freeze the initial matrix by:
\begin{equation}
W_0 \xleftarrow{} W_0-BA.
\end{equation}

\subsection{Analysis}
\label{subsec:analysis}
We elucidate the superiority of FeDeRA by delving into the process of model updating.  Firstly, we define magnitude vector $M_w\in \mathbb{R}^{1\times k}$ and direction matrix$D_w \in \mathbb{R}^{d\times k}$ for a weight matrix $W\in \mathbb{R}^{d\times k}$ as\cite{liu2024dora}:
\begin{equation}
M_w=[m_w^1,m_w^2\ldots m_w^k]=[||w^1||_2,||w^2||_2\ldots ||w^k||_2],
\end{equation}
\begin{equation}
D_w=[d_w^1,d_w^2\ldots d_w^k]=[\frac{w^1}{m_w^1},\frac{w^2}{m_w^2}\ldots \frac{w^k}{m_w^k}],
\end{equation}
where $w^j$ is the is the $j$-th column vector of $W$. Define magnitude variation $\Delta M_w(\cdot)$ and direction variation $\Delta D_w(\cdot)$ between two weights matrics $W_1\in \mathbb{R}^{d\times k}$ and $W_2\in \mathbb{R}^{d\times k}$as:
\begin{equation}
\Delta M_w(W_1,W_2)=\frac{\Sigma_{n=1}^k|m_{w1}^n-m_{w2}^n|}{k},
\end{equation}
\begin{equation}
\Delta D_w(W_1,W_2)=\frac{\Sigma_{n=1}^k(1-\text{CosineSimilarity}(d_{w1}^n,d_{w2}^n))}{k}.
\end{equation}

\begin{figure}[!t]
    \centering
    \begin{subfigure}{.24\textwidth}
        \includegraphics[width=\linewidth]{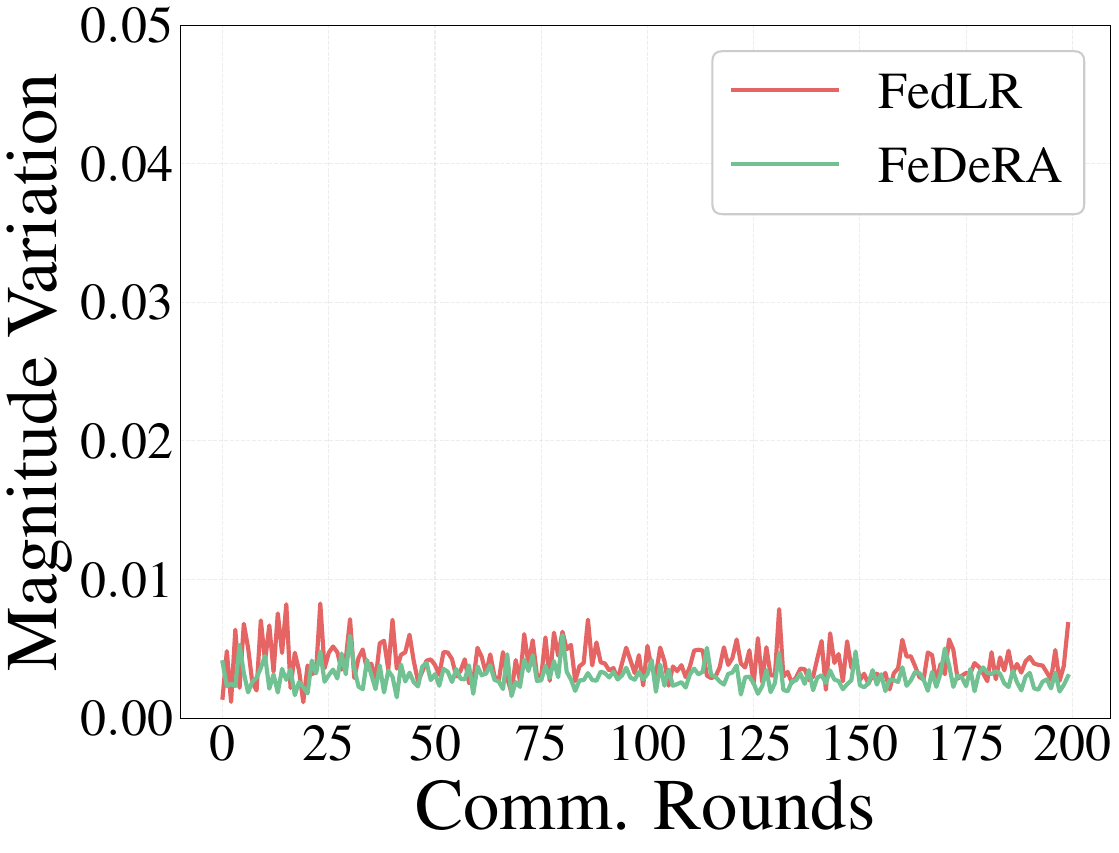}
        \caption{layer.0.q.lora\_A}
    \end{subfigure}\hfill
    \begin{subfigure}{.24\textwidth}
        \includegraphics[width=\linewidth]{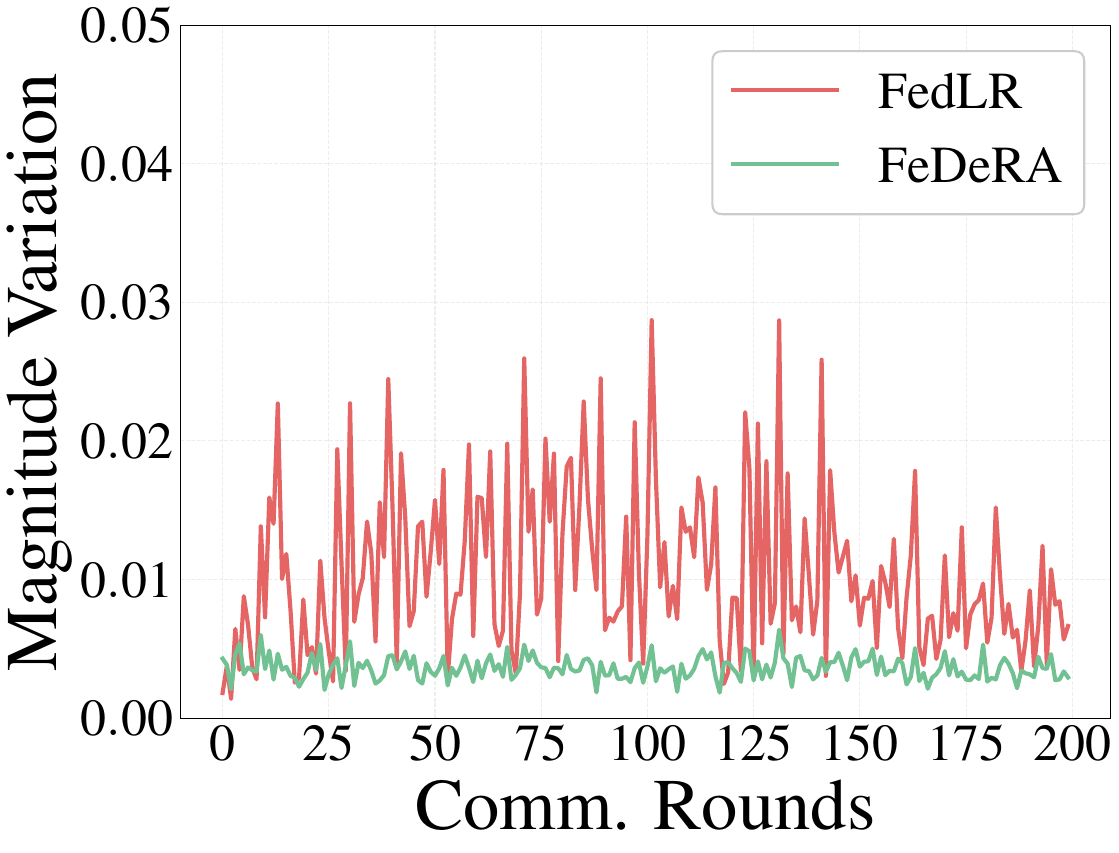}
        \caption{layer.0.v.lora\_A}
    \end{subfigure}\hfill
    \begin{subfigure}{.24\textwidth}
        \includegraphics[width=\linewidth]{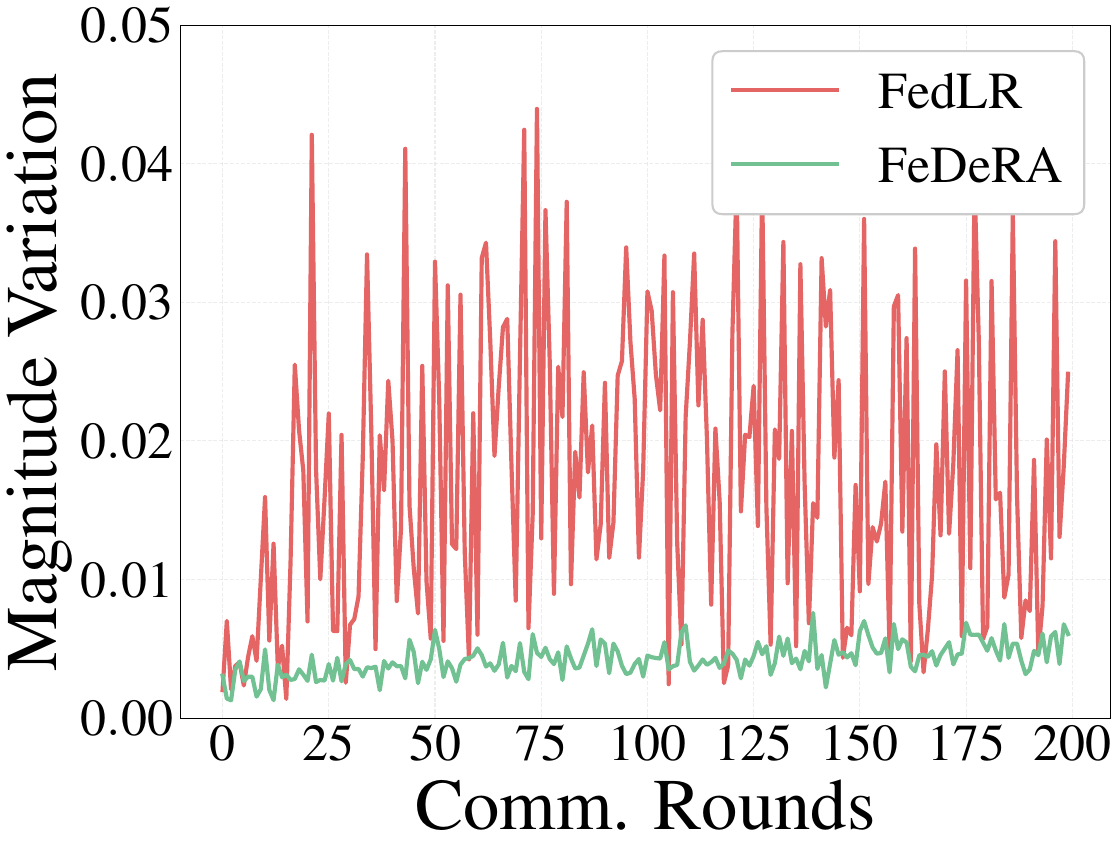}
        \caption{layer.5.q.lora\_A}
    \end{subfigure}\hfill
    \begin{subfigure}{.24\textwidth}
        \includegraphics[width=\linewidth]{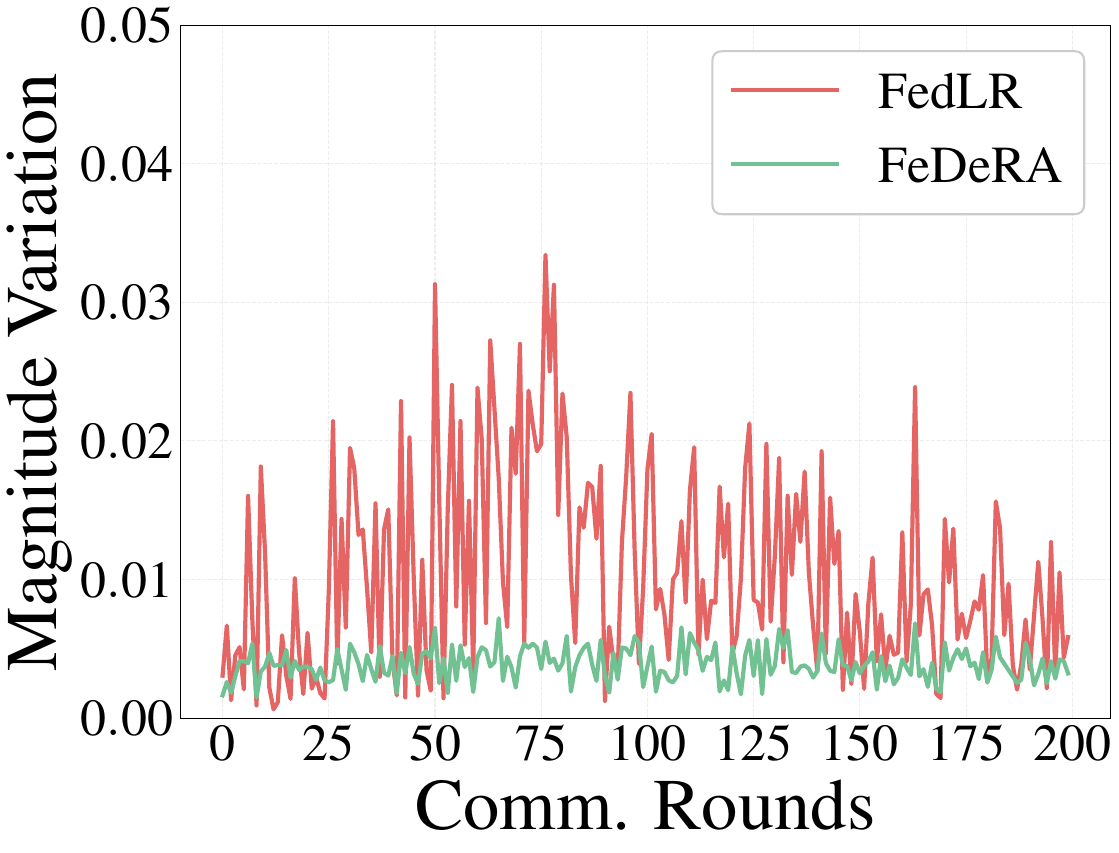}
        \caption{layer.5.v.lora\_A}
    \end{subfigure}

    \begin{subfigure}{.24\textwidth}
        \includegraphics[width=\linewidth]{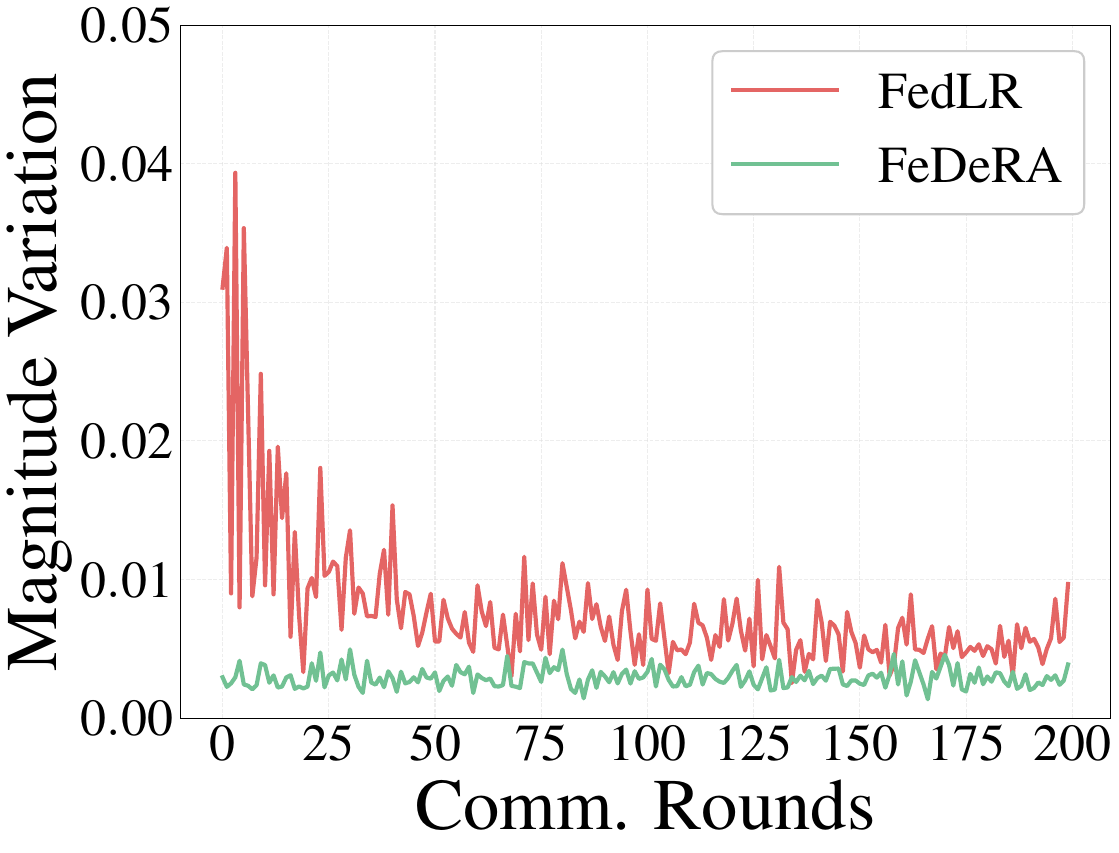}
        \caption{layer.0.q.lora\_B}
    \end{subfigure}\hfill
    \begin{subfigure}{.24\textwidth}
        \includegraphics[width=\linewidth]{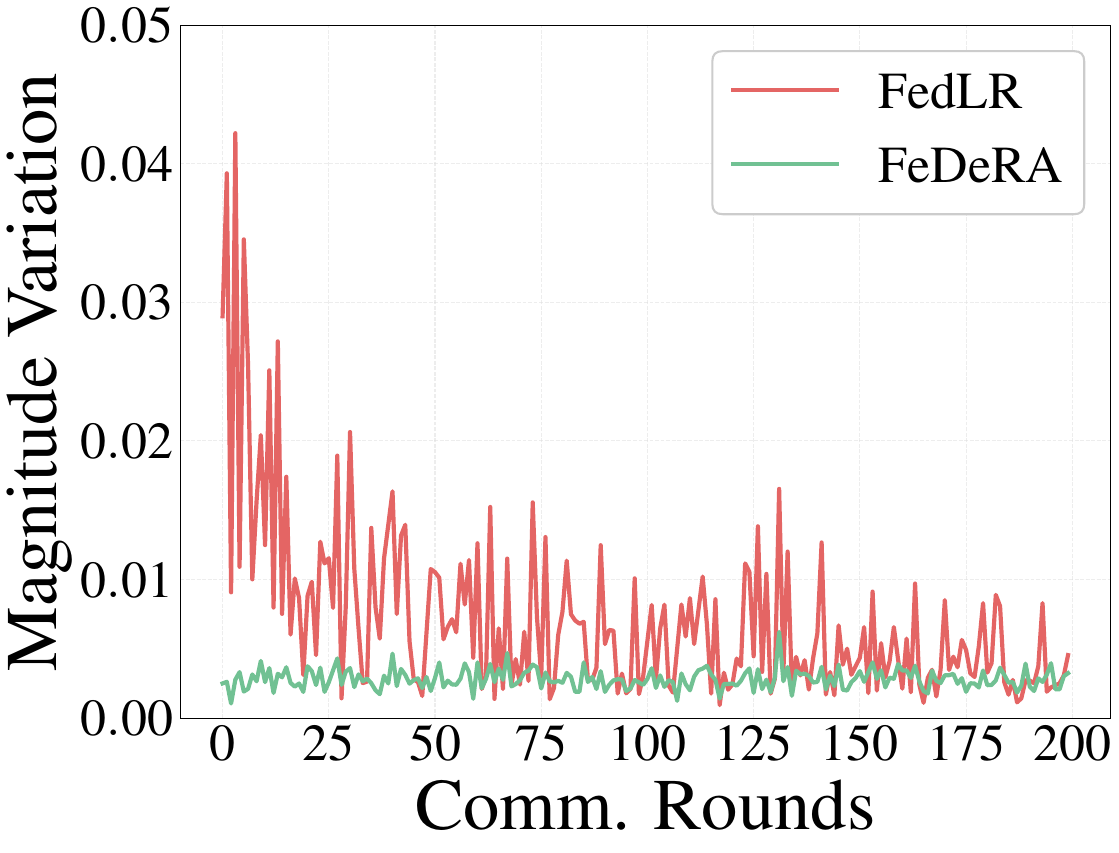}
        \caption{layer.0.v.lora\_B}
    \end{subfigure}\hfill
    \begin{subfigure}{.24\textwidth}
        \includegraphics[width=\linewidth]{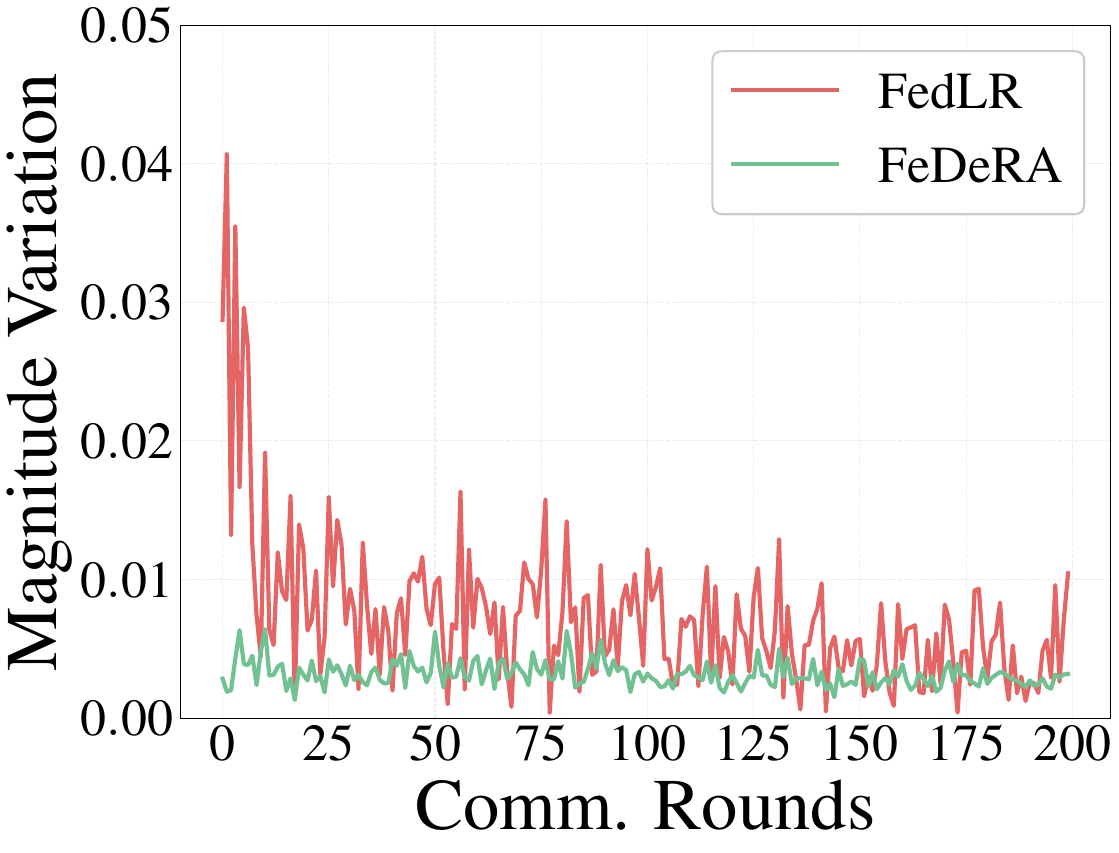}
        \caption{layer.5.q.lora\_B}
    \end{subfigure}\hfill
    \begin{subfigure}{.24\textwidth}
        \includegraphics[width=\linewidth]{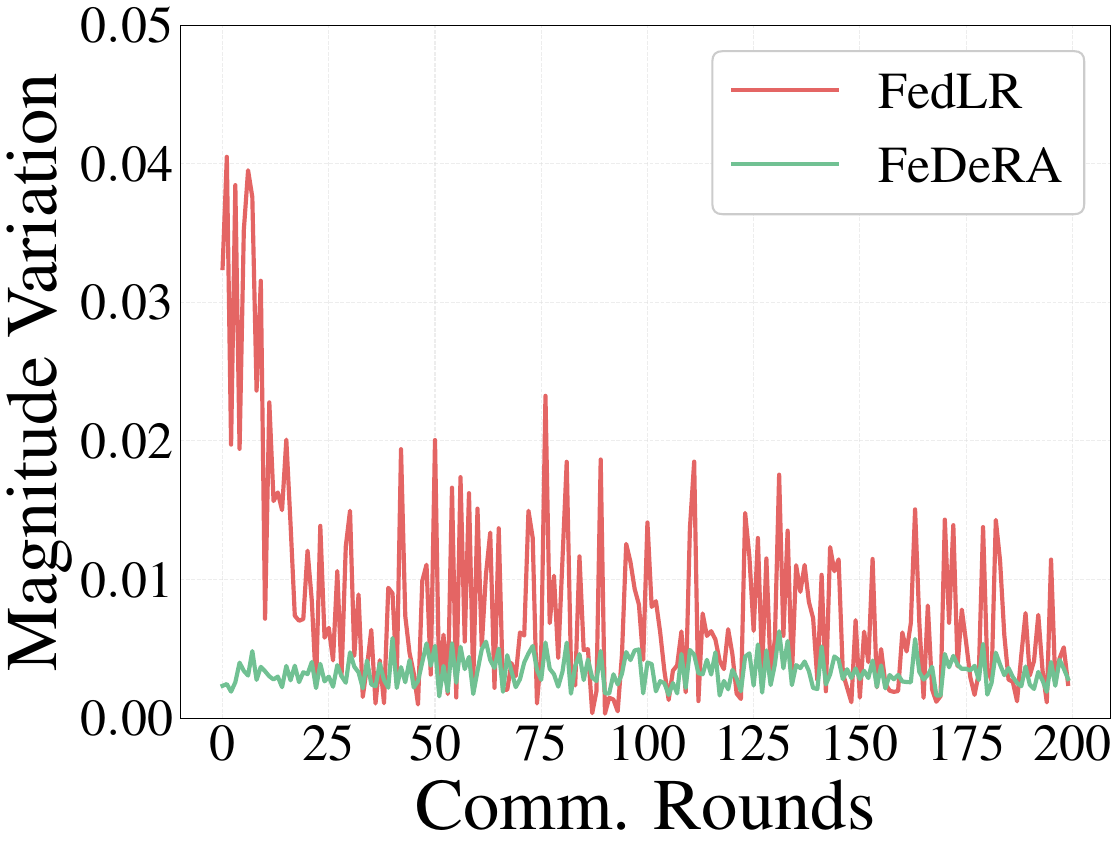}
        \caption{layer.5.v.lora\_B}
    \end{subfigure}

    \caption{Magnitude variation in consecutive global weight updates by FeDeRA and FedLR fine-tuning DistilBERT over 200 Communication Rounds on the 20 Newsgroups Dataset.}
    \label{fig:mag and dir variation}
\end{figure}

\begin{figure}[!htb]
    \centering
    \begin{subfigure}{.24\textwidth}
        \includegraphics[width=\linewidth]{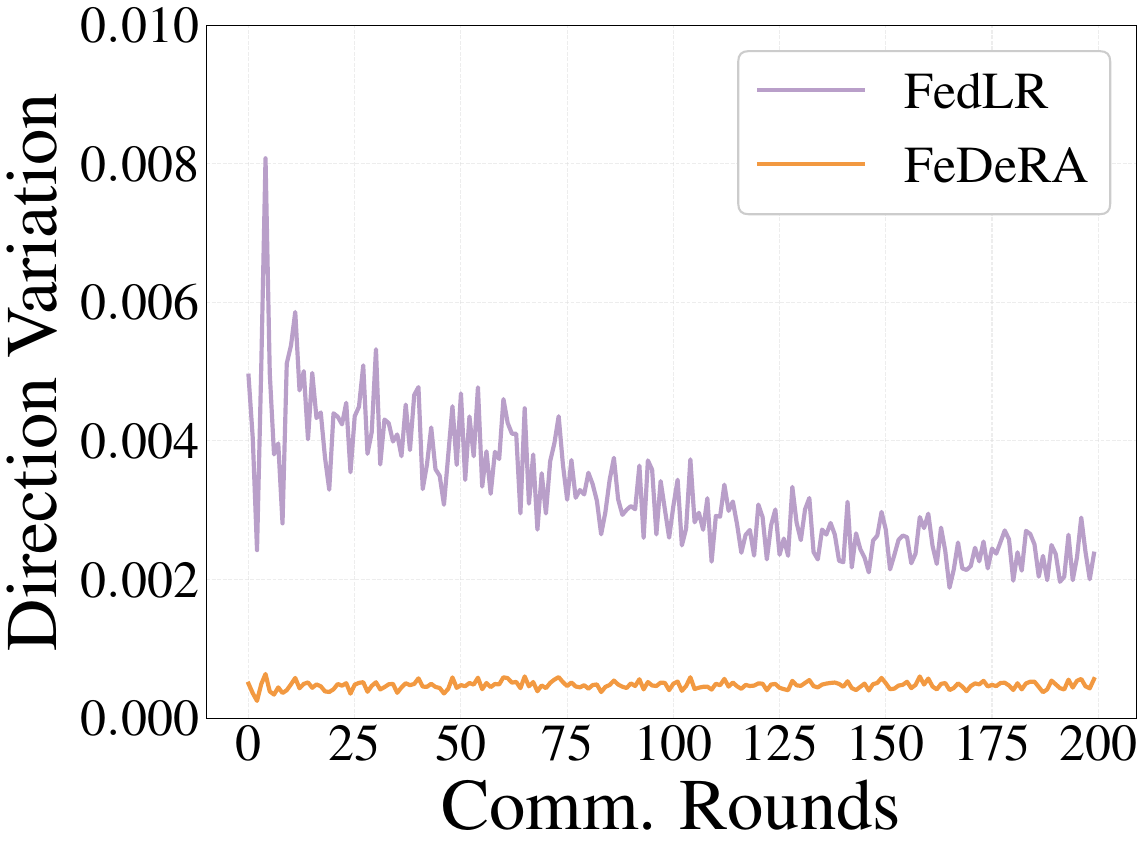}
        \caption{layer.0.q.lora\_A}
    \end{subfigure}\hfill
    \begin{subfigure}{.24\textwidth}
        \includegraphics[width=\linewidth]{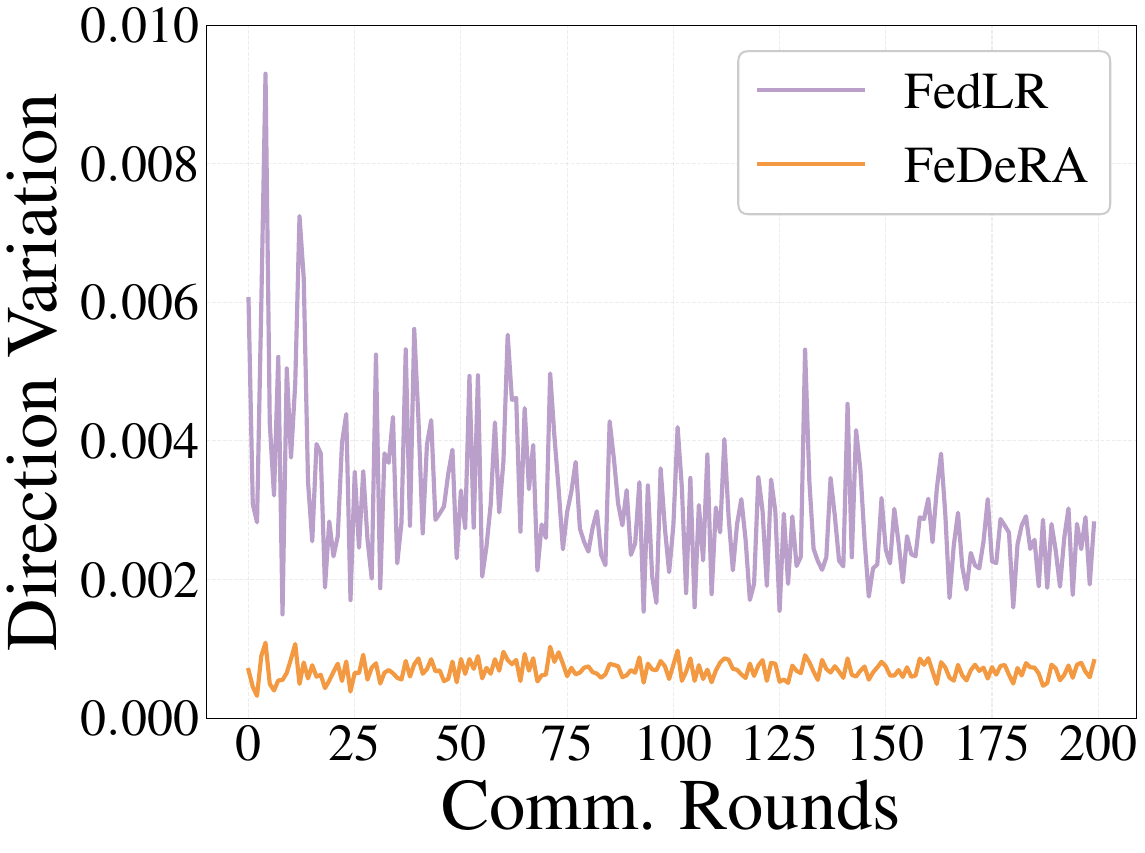}
        \caption{layer.0.v.lora\_A}
    \end{subfigure}\hfill
    \begin{subfigure}{.24\textwidth}
        \includegraphics[width=\linewidth]{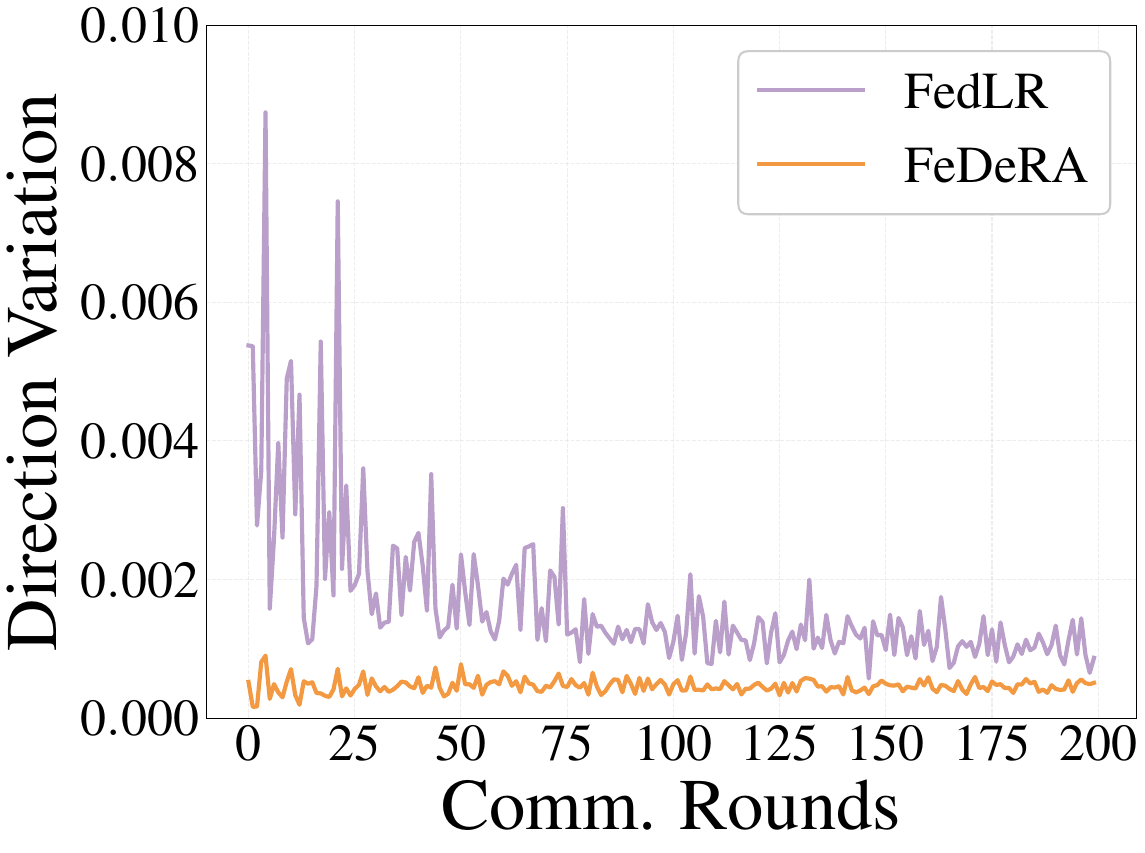}
        \caption{layer.5.q.lora\_A}
    \end{subfigure}\hfill
    \begin{subfigure}{.24\textwidth}
        \includegraphics[width=\linewidth]{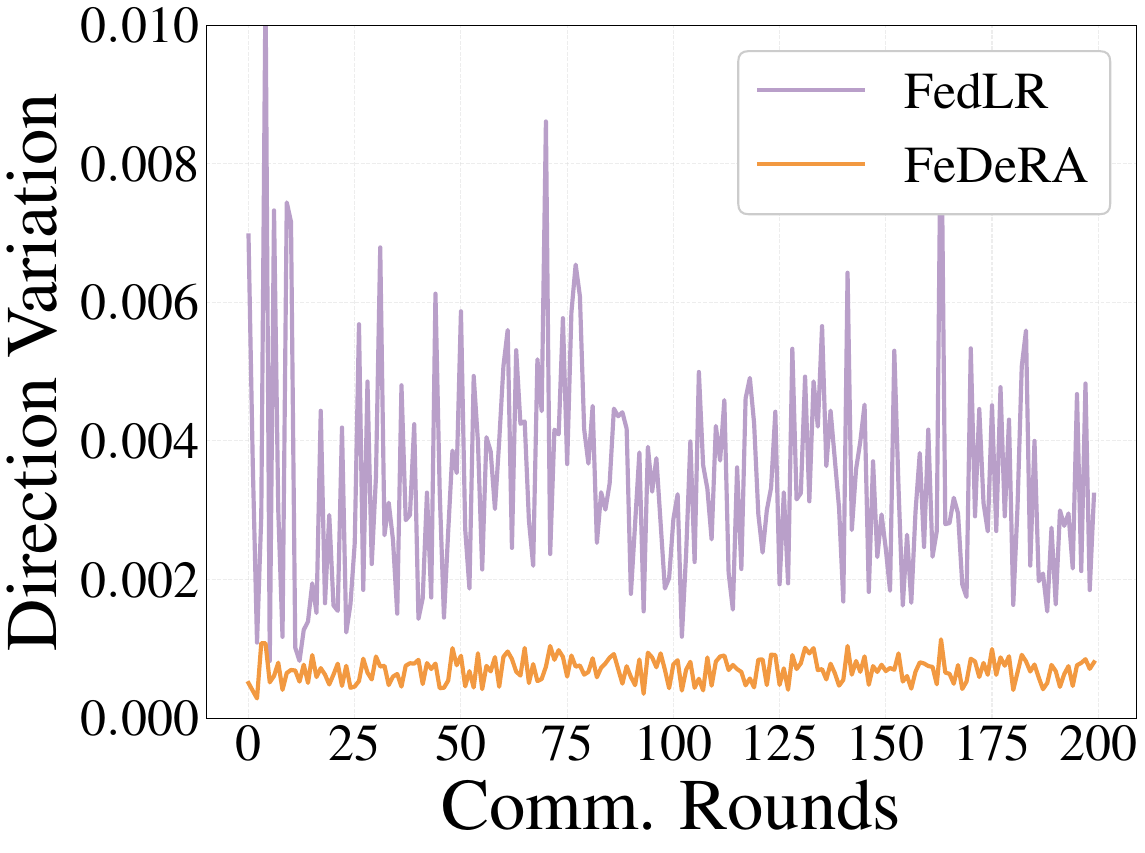}
        \caption{layer.5.v.lora\_A}
    \end{subfigure}

    \begin{subfigure}{.24\textwidth}
        \includegraphics[width=\linewidth]{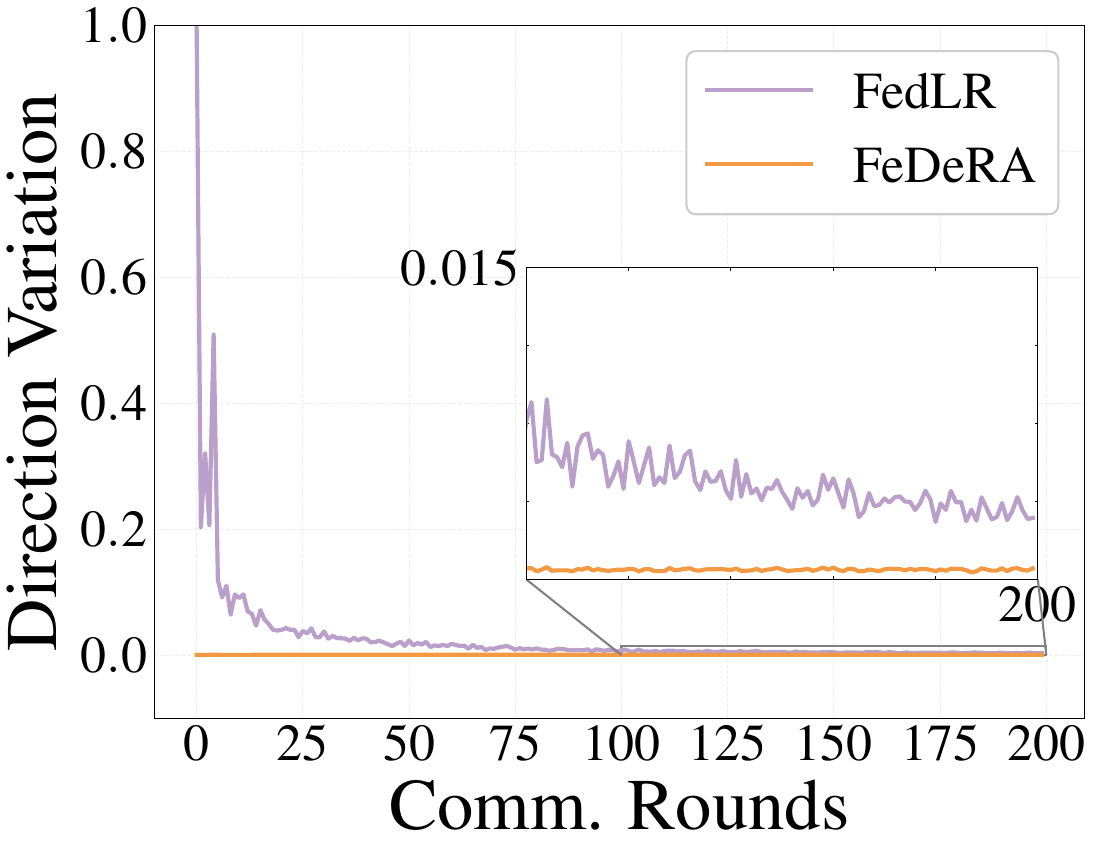}
        \caption{layer.0.q.lora\_B}
        \label{subfig:0_q_b}
    \end{subfigure}\hfill
    \begin{subfigure}{.24\textwidth}
        \includegraphics[width=\linewidth]{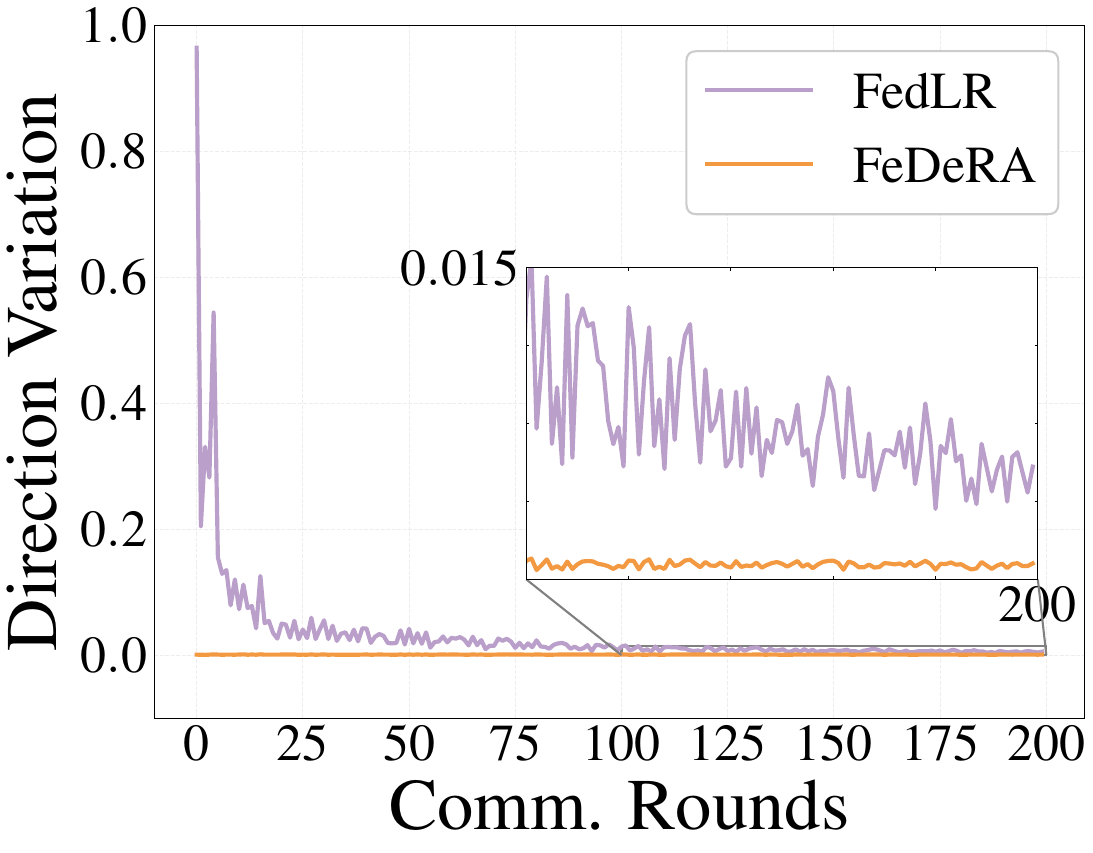}
        \caption{layer.0.v.lora\_B}
        \label{subfig:0_v_b}
    \end{subfigure}\hfill
    \begin{subfigure}{.24\textwidth}
        \includegraphics[width=\linewidth]{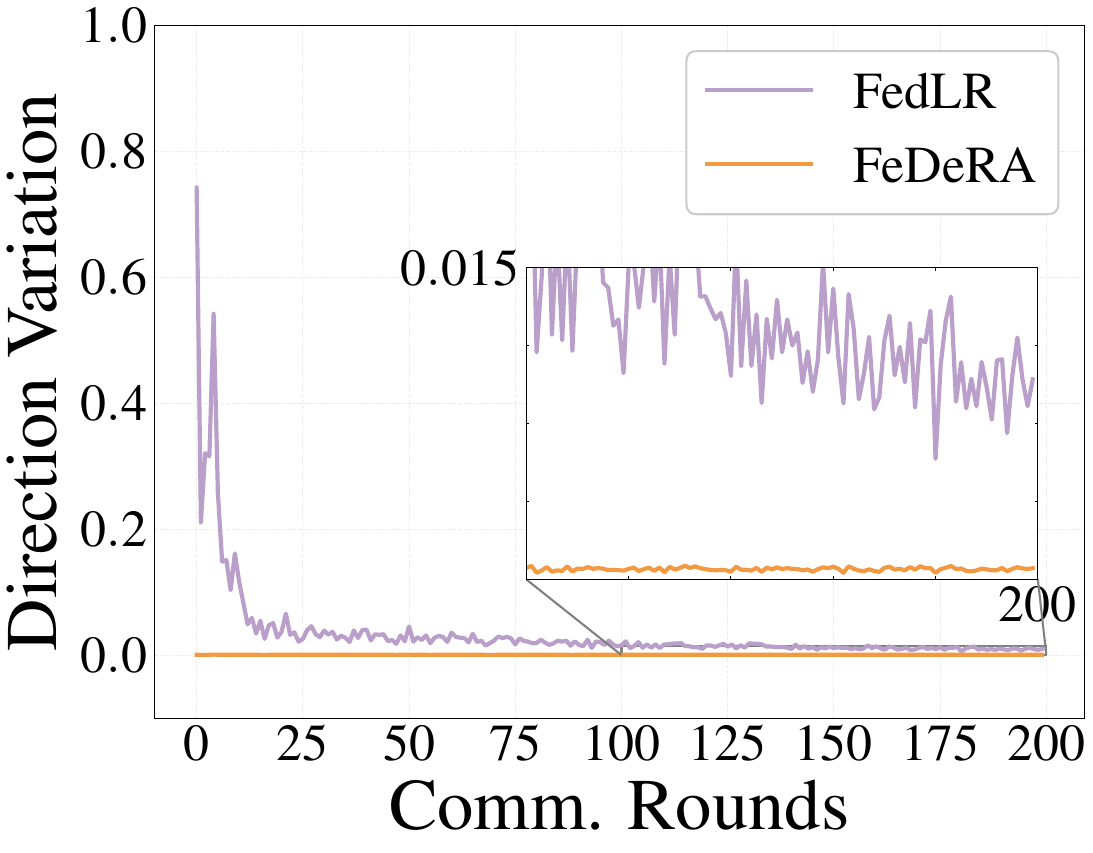}
        \caption{layer.5.q.lora\_B}
        \label{subfig:5_q_b}
    \end{subfigure}\hfill
    \begin{subfigure}{.24\textwidth}
        \includegraphics[width=\linewidth]{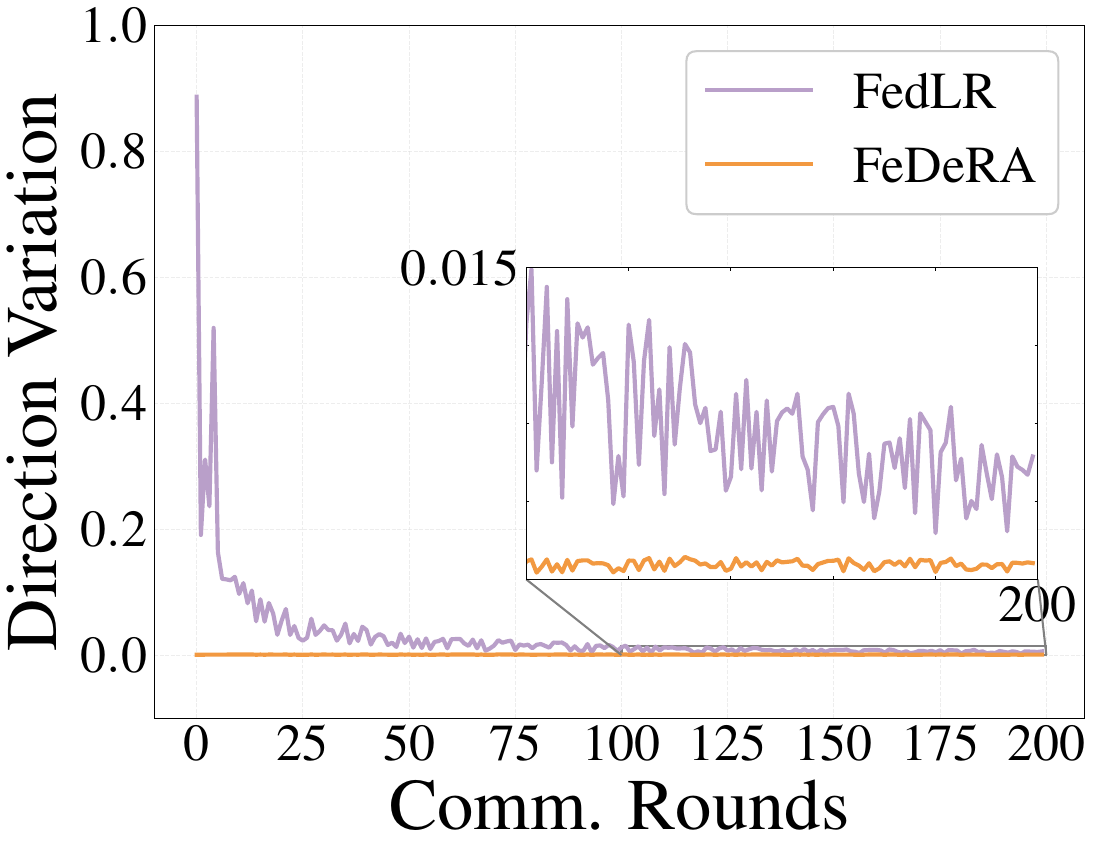}
        \caption{layer.5.v.lora\_B}
        \label{subfig:5_v_b}
    \end{subfigure}

    \caption{Direction variation in consecutive global weight updates by FeDeRA and FedLR fine-tuning DistilBERT over 200 communication rounds on the 20Newsgroups dataset.}
\end{figure}

We then compare the magnitude variation and direction variation in successive global weight updates between the proposed FeDeRA and the original LoRA approach adapted to a FL setting, referred to as FedLR. Specifically, we calculate $\Delta M_w(\omega^t, \omega^{t-1})$ and $\Delta D_w(\omega^t, \omega^{t-1})$ for both FedLR and FeDeRA, as shown in \autoref{fig:mag and dir variation} for $t$ ranging from 0 to 200. The results indicate that the original LoRA method exhibits more dramatic changes in the parameter updates of matrices $A$ and $B$, with magnitude and direction variation in successive updates being several to hundreds of times greater than those by FeDeRA. This is particularly noticeable for the $B$ matrix at the beginning of the training, likely due to its initialization to zero. As illustrated in \autoref{subfig:0_q_b}, \autoref{subfig:0_v_b}, \autoref{subfig:5_q_b}, and \autoref{subfig:5_v_b}, the initial weight updates tend to be orthogonal or opposite to their consecutive updates. This supports our assertion that LoRA's initialization under highly heterogeneous data training leads to more severe weight update drift, making convergence more challenging. In contrast, our proposed FeDeRA methodology exhibits stability in both magnitude variation and direction changes, facilitating faster convergence and effectively mitigating the impact of data heterogeneity in FL.

\section{Evaluation}
In this section, we present numerical results to evaluate the performance of the proposed FeDeRA using the FedML\cite{he2020fedml} and FedNLP\cite{lin2022fednlp} frameworks. Specifically, we first compare the performance of FeDeRA with various baselines across six datasets and three NLP tasks, using highly non-IID training data generated according to a Dirichlet distribution. We then implement FeDeRA and the baseline methods on practical devices and access time cost of different approaches to achieve a target accuracy to evalute the training efficiency.

\subsection{Baselines}
\label{subsec:baselines}
We adopt existing FFT and PEFT methods within the FL setting as baselines. To ensure a fair comparison, the experimental settings for these baselines are consistent with those in their original works, unless explicitly stated otherwise. The details of the baselines are as follows:

\textbf{FedFT:} The conventional approach which updates all the model's parameters throughout the fine-tuning process.

\textbf{FedBF:} Each client updates only the bias terms while keeping the other parameters fixed.

\textbf{FedAP:} This approach incorporates adapter-tuning by inserting some trainable adapter layers into the model while keeping the parameters of the original model frozen. Specifically, adapters are inserted after the attention and feed-forward modules, following the methodology outlined in \cite{houlsby2019parameter}. The adapter module is implemented using the adapters library \cite{poth-etal-2023-adapters,pfeiffer2020AdapterHub}.

\textbf{FedLR:} This approach integrates LoRA into FL, , applying the LoRA module solely to the query and value modules, similar to the proposed FeDeRA. LoRA implementation is carried out using the PEFT library \cite{peft}.

\subsection{Setup}
\textbf{Models and Datasets.} We adopt RoBERTa-base \cite{liu2019roberta} and DeBERTaV3-base \cite{he2022debertav3} as the backbone models in our experiments, implementing them using the transformers library\cite{wolf-etal-2020-transformers}. We consider three NLP tasks: text classification, named entity recognition, and question answering. Specifically,  we use the 20Newsgroups\cite{lang1995newsweeder} and SemEval-2010Task8\cite{hendrickx2010semeval} datasets for text classification, WNUT2017\cite{derczynski2017results} and PLONER\cite{fu2020rethinking} datasets for named entity recognition, SQuADv1.1\cite{rajpurkar2016squad} and MRQA\cite{fisch2019mrqa} datasets for question answering, respectively. Additional details about the datasets and parameter configurations can be found in \autoref{tab:datasets details}. It is noteworthy that we select as small $\alpha$ as possible to ensure extreme data heterogeneity without significantly impeding the convergence of all methods. Further information on this choice is available in \autoref{appendix:data partition}.

\begin{table}[htbp]
\centering
\caption{Details of the used dataset and  Non-IID settings. $*$ indicates that the dataset are divided into 80\% and 20\%  for training and test, respectively. \# of class denotes the number of classes into which we split the dataset. $\alpha$ is used to control the magnitude of data heterogeneity.\vspace{1mm}}
\label{tab:datasets details}
\resizebox{1\textwidth}{!}{
\begin{tabular}{cccccccc} 
\toprule[2pt]
\multicolumn{1}{c}{\textbf{Dataset}} & \textbf{Task}                                              & \textbf{\#Train} & \textbf{\#Test} & \textbf{\# of clients} & \textbf{\# selected per round} & \textbf{\textbf{\# of classes}} & \textbf{$\alpha$}  \\ 
\midrule
20NEWS                               & Text Classification                                        & 11.3k            & 7.5k            & 100                    & 10                             & 20                            & 0.1       \\
SEMEVAL                          & Text Classification                                        & 8k               & 2.7k            & 100                    & 10                             & 19                            & 1         \\
WNUT                             & Named Entity Recognition & 3.4k             & 1.2k            & 30                     & 5                              & 37                            & 0.1       \\
$\text{PLONER}^*$                               & Named Entity Recognition & 14k              & 3.5k            & 100                    & 10                             & 49                            & 0.01      \\
SQuADv1.1                            & Question Answering                                         & 87.5             & 34.7            & 300                    & 15                             & 30                            & 0.1       \\
$\text{MRQA}^*$                                 & Question Answering                                         & 45.5k            & 11.3k           & 300                    & 15                             & 6                             & 0.01      \\
\bottomrule[2pt]
\end{tabular}}
\end{table}

\textbf{Training settings.} The training batch size is configured to 16, with each local training epoch set to 1. We select the optimal learning rate from the range [$1\times10^{-3}$, $5\times10^{-4}$, $1\times10^{-4}$, $5\times10^{-5}$, $1\times10^{-5}$] for each backbone model and dataset settings. Additionally, the maximum sequence length is set to 64 for SemEval-2010Task8, WNUT2017, and PLONER, 256 for 20Newsgroup, 384 for SQuADv1.1, and 512 for MRQA. For FedLR and FeDeRA, the rank $r$ is configured as 32, and $\beta$ is set to be equal to $r$.

\subsection{Task Performance}
\label{subsec:performance evaluation}

\autoref{Tab:tc result}, \autoref{Tab:ner result} and \autoref{Tab:qa result} presents the performance of the proposed FeDeRA and baselines in text classification, named entity recognition and question answering, respectively. The results indicate that the proposed FeDeRA consistently outperforms FedBF, FedAP and FedLR across these tasks. Moreover, we can also observe that the proposed FeDeRA achieves comparable performance to FedFT, and even performs better than FedFT on 20NEWS and WNUT datasets with only about 1\% trainable parameters. These results demonstrate the superiority of the proposed method.

\begin{table}[ht]
\centering
\caption{Evaluation results on text classification tasks. The data marked with \textbf{bold} and \uline{underlined} indicates the best and second-best results achieved by the five methods under each configuration. Each result is derived by averaging over five experiments with different random seeds. This applies to \autoref{Tab:ner result} and \autoref{Tab:qa result} as well.\vspace{1mm}}
\label{Tab:tc result}
\resizebox{1\textwidth}{!}{
\begin{tabular}{c|l|c|llllll} 
\toprule[2pt]
\multirow{3}{*}{\textbf{Model}} & \multicolumn{1}{c|}{\multirow{3}{*}{\textbf{Method}}} & \multirow{3}{*}{\begin{tabular}[c]{@{}c@{}}\textbf{\#Trainable}\\\textbf{ Parameters}\end{tabular}} & \multicolumn{3}{c}{\textbf{20NEWS(Acc.)}}                                    & \multicolumn{3}{c}{\textbf{SEMEVAL(Acc.)}}                                    \\ 
\cline{4-9}
                                & \multicolumn{1}{c|}{}                                 &                                                                                                     & \multicolumn{3}{c}{\textbf{Comm. rounds}}                                    & \multicolumn{3}{c}{\textbf{Comm. rounds}}                                     \\ 
                                & \multicolumn{1}{c|}{}                                 &                                                                                                     & \multicolumn{1}{c}{200} & \multicolumn{1}{c}{500} & \multicolumn{1}{c}{1000} & \multicolumn{1}{c}{200} & \multicolumn{1}{c}{500} & \multicolumn{1}{c}{1000}  \\ 
\cmidrule{1-9}
\multirow{5}{*}{RoBERTa}        & FedFT                                                 & \multicolumn{1}{c|}{125M}                                                                           & \textbf{79.56$_{\pm0.6}$}  & \textbf{82.71$_{\pm0.2}$}  & \textbf{83.67$_{\pm0.3}$}   & \uline{80.34$_{\pm1.6}$}  & \textbf{82.99$_{\pm0.4}$}  & \uline{83.53$_{\pm0.8}$}  \\ 
                                & FedBF                                                 & \multicolumn{1}{c|}{0.1M}                                                                           & 65.41$_{\pm0.9}$           & 70.32$_{\pm0.3}$           & 72.21$_{\pm0.2}$            & 72.03$_{\pm1.1}$          & 77.28$_{\pm0.5}$           & 78.66$_{\pm0.1}$             \\
                                & FedAP                                                 & \multicolumn{1}{c|}{1.8M}                                                                           & 74.73$_{\pm0.1}$           & 78.46$_{\pm0.3}$           & 80.78$_{\pm0.1}$            & 72.98$_{\pm0.5}$           & 79.39$_{\pm0.1}$           & 80.66$_{\pm0.2}$             \\
                                & FedLR                                                 & \multicolumn{1}{c|}{1.2M}                                                                           & 73.17$_{\pm0.6}$           & 78.03$_{\pm0.2}$           & 80.25$_{\pm0.3}$            & 74.81$_{\pm1.7}$          & 79.26$_{\pm0.5}$           & 80.37$_{\pm0.8}$             \\
                                & \textbf{FeDeRA}                                       & \multicolumn{1}{c|}{1.2M}                                                                           & \uline{77.24$_{\pm0.2}$}   & \uline{80.33$_{\pm0.2}$}   & \uline{82.21$_{\pm0.1}$}    & \textbf{80.74$_{\pm0.6}$}  & \uline{82.91$_{\pm0.5}$}   & \textbf{83.78$_{\pm0.4}$}    \\ 
\midrule
\multirow{5}{*}{DeBERTaV3}        & FedFT                                                 & 184M                                                                                                & \uline{76.47$_{\pm1.2}$}  & \uline{82.25$_{\pm0.6}$}  & \uline{83.99$_{\pm0.4}$}    & \textbf{79.57$_{\pm2.5}$} & \textbf{83.66$_{\pm1.2}$} & \textbf{84.42$_{\pm1.1}$}   \\
                                & FedBF                                                 & 0.1M                                                                                                & 38.93$_{\pm1.3}$          & 55.72$_{\pm0.2}$          & 61.37$_{\pm1.3}$           & 31.44$_{\pm1.9}$          & 64.37$_{\pm0.9}$           & 73.79$_{\pm0.9}$             \\
                                & FedAP                                                 & 1.8M                                                                                                & 58.75$_{\pm1.7}$          & 73.54$_{\pm0.8}$          & 79.09$_{\pm0.4}$            & 64.05$_{\pm1.1}$          & 78.08$_{\pm0.7}$           & 80.46$_{\pm0.2}$             \\
                                & FedLR                                                 & 1.2M                                                                                                & 50.41$_{\pm1.3}$          & 73.02$_{\pm0.9}$           & 80.16$_{\pm0.2}$            & 39.76$_{\pm2.3}$          & 71.11$_{\pm1.9}$          & 80.13$_{\pm1.4}$            \\
                                & \textbf{FeDeRA}                                       & 1.2M                                                                                                & \textbf{76.91$_{\pm1.1}$} & \textbf{82.57$_{\pm0.5}$} & \textbf{84.38$_{\pm0.4}$}   & \uline{72.45$_{\pm2.5}$}  & \uline{83.09$_{\pm0.8}$}   & \uline{84.36$_{\pm0.5}$}     \\
\bottomrule[2pt]
\end{tabular}}
\end{table}

\begin{table}[htbp]
\centering
\caption{Evaluation results on named entity recognition tasks .\vspace{1mm}}
\label{Tab:ner result}
\resizebox{1\textwidth}{!}{
\begin{tabular}{c|l|c|llllll} 
\toprule[2pt]
\multicolumn{1}{c|}{\multirow{3}{*}{\textbf{Model}}} & \multicolumn{1}{c|}{\multirow{3}{*}{\textbf{Method}}} & \multirow{3}{*}{\begin{tabular}[c]{@{}c@{}}\textbf{\#Trainable}\\\textbf{ Parameters}\end{tabular}}
& \multicolumn{3}{c}{\textbf{WNUT(F1)}}                                   & \multicolumn{3}{c}{\textbf{PLONER(F1)}}                                 \\ 
\cline{4-9}
\multicolumn{1}{c|}{}                                & \multicolumn{1}{c|}{}                                 &                                    & \multicolumn{3}{c}{\textbf{Comm. rounds}}                                   & \multicolumn{3}{c}{\textbf{Comm. rounds}}                               \\ 
\multicolumn{1}{c|}{}                                & \multicolumn{1}{c|}{}                                 &                                    & \multicolumn{1}{c}{100} & \multicolumn{1}{c}{200} & \multicolumn{1}{c}{500} & \multicolumn{1}{c}{50}                     & \multicolumn{1}{c}{100}                   & \multicolumn{1}{c}{150}                    \\ 
\midrule
\multirow{5}{*}{RoBERTa}                             & FedFT                                                 & 125M                               & \textbf{50.03$_{\pm.5}$}  & \uline{51.53$_{\pm.9}$}   & \uline{52.31}$_{\pm.2}$   & \textbf{86.01$_{\pm.6}$} & \textbf{88.01$_{\pm.1}$} & \textbf{89.19$_{\pm.1}$}  \\
                                                     & FedBF                                                 & 0.1M                               & 35.71$_{\pm2.9}$          & 42.45$_{\pm.9}$           & 45.23$_{\pm1.1}$          & 75.95$_{\pm.6}$           & 79.81$_{\pm.1}$          & 80.78$_{\pm.1}$           \\
                                                     & FedAP                                                 & 1.8M                               & 47.81$_{\pm1.1}$          & 50.11$_{\pm.1}$           & 50.25$_{\pm.2}$           & 79.23$_{\pm.1}$           & 83.21$_{\pm.6}$          & 85.72$_{\pm.2}$           \\
                                                     & FedLR                                                 & 1.2M                               & 46.39$_{\pm2.4}$          & 49.71$_{\pm.3}$           & 50.15$_{\pm.6}$           & 81.14$_{\pm.1}$           & 85.29$_{\pm.1}$          & 86.88$_{\pm.4}$           \\
                                                     & \textbf{FeDeRA}                                                   & 1.2M                               & \uline{49.14$_{\pm1.3}$}  & \textbf{52.28$_{\pm.7}$}  & \textbf{52.73}$_{\pm.8}$  & \uline{84.74$_{\pm.5}$}   & \uline{87.14$_{\pm.1}$}  & \uline{88.44$_{\pm.1}$}   \\ 
\midrule
\multirow{5}{*}{DeBERTaV3}                          & FedFT                                                 & 184M                               & \textbf{50.26}$_{\pm.7}$  & \textbf{51.04}$_{\pm.8}$  & \textbf{51.45}$_{\pm.6}$  & \textbf{81.41$_{\pm.4}$}  & \textbf{86.24$_{\pm.3}$} & \textbf{87.06$_{\pm.1}$}   \\
                                                     & FedBF                                                 & 0.1M                               & 42.79$_{\pm.5}$           & 46.25$_{\pm.3}$           & 47.54$_{\pm.4}$           & 70.44$_{\pm.1}$           & 75.22$_{\pm.2}$          & 77.54$_{\pm.3}$           \\
                                                     & FedAP                                                 & 1.8M                               & 43.29$_{\pm1.3}$          & 45.45$_{\pm1.9}$          & 47.68$_{\pm.5}$           & 77.46$_{\pm.2}$           & 82.69$_{\pm.5}$          & 84.82$_{\pm.6}$           \\
                                                     & FedLR                                                 & 1.2M                               & 43.75$_{\pm1.3}$          & 46.78$_{\pm1.7}$          & 48.82$_{\pm.6}$           & 76.61$_{\pm.3}$           & 82.55$_{\pm.5}$          & 83.88$_{\pm.2}$           \\
                                                     & \textbf{FeDeRA}                                                  & 1.2M                               & \uline{50.02}$_{\pm1.2}$  & \uline{50.51}$_{\pm1.1}$  & \uline{51.14}$_{\pm.6}$   & \uline{81.02$_{\pm.2}$}   & \uline{84.59$_{\pm.1}$} & \uline{86.11$_{\pm.1}$}   \\
\bottomrule[2pt]
\end{tabular}}
\end{table}

\begin{table}[!t]
\centering
\caption{Evaluation results on question answering tasks.\vspace{1mm}}
\label{Tab:qa result}
\resizebox{1\textwidth}{!}{
\begin{tabular}{c|l|c|llll} 
\toprule[2pt]
\multirow{3}{*}{\textbf{Model}} & \multicolumn{1}{c|}{\multirow{3}{*}{\textbf{Method}}} & \multirow{3}{*}{\begin{tabular}[c]{@{}c@{}}\textbf{\#Trainable}\\\textbf{ Parameters}\end{tabular}} & \multicolumn{2}{c}{\textbf{SQuADv1.1(EM/F1)}}                                                          & \multicolumn{2}{c}{\textbf{MRQA(EM/F1)}}                                                              \\ 
\cline{4-7}
                                & \multicolumn{1}{c|}{}                                 & \multicolumn{1}{l|}{}                                   & \multicolumn{2}{c}{\textbf{Comm. rounds}}                                                             & \multicolumn{2}{c}{\textbf{Comm. rounds}}                                                             \\ 
                                & \multicolumn{1}{c|}{}                                 & \multicolumn{1}{l|}{}                                   & \multicolumn{1}{c}{50}                            & \multicolumn{1}{c}{100}                           & \multicolumn{1}{c}{100}                          & \multicolumn{1}{c}{200}                            \\ 
\midrule
\multirow{5}{*}{RoBERTa}        & FedFT                                                 & 125M                                                    & \textbf{63.16$_{\pm.3}$} \slash\ \textbf{77.84$_{\pm.3}$} & \uline{65.27$_{\pm.2}$} \slash\ \uline{79.62$_{\pm.1}$}   & \textbf{51.45$_{\pm.5}$} \slash\ \uline{60.78$_{\pm.1}$} & \textbf{52.44$_{\pm.2}$} \slash\ \textbf{63.24$_{\pm.3}$}  \\
                                & FedBF                                                 & 0.1M                                                    & 43.87$_{\pm2.4}$ \slash\ 60.91$_{\pm2.1}$                 & 50.09$_{\pm.7}$ \slash\ 66.53$_{\pm.8}$                   & 25.67$_{\pm1.5}$ \slash\ 37.88$_{\pm1.1}$                & 33.69$_{\pm.1}$ \slash\ 45.55$_{\pm.3}$                    \\
                                & FedAP                                                 & 1.8M                                                    & 62.02$_{\pm.1.2}$ \slash\ 76.84$_{\pm1.2}$                & 65.14$_{\pm.2}$ \slash\ 79.42$_{\pm.4}$                   & 47.38$_{\pm.1}$ \slash\ 59.21$_{\pm.3}$                  & 51.01$_{\pm.1}$ \slash\ 62.25$_{\pm.4}$                    \\
                                & FedLR                                                 & 1.2M                                                    & 57.37$_{\pm.2}$ \slash\ 72.68$_{\pm.1}$                   & 60.87$_{\pm.5}$ \slash\ 75.91$_{\pm.4}$                   & 46.85$_{\pm.3}$ \slash\ 58.4$_{\pm.2}$                   & 49.88$_{\pm.5}$ \slash\ 61.16$_{\pm.3}$                    \\
                                & \textbf{FeDeRA}                                       & 1.2M                                                    & \uline{62.53$_{\pm.2}$} \slash\ \uline{77.51$_{\pm.2}$}   & \textbf{65.31$_{\pm.3}$} \slash\ \textbf{79.98$_{\pm.2}$}  & \uline{49.73$_{\pm.5}$} \slash\ \textbf{61.27$_{\pm.4}$} & \uline{52.05$_{\pm.3}$} \slash\ \uline{63.11$_{\pm.5}$}    \\ 
\midrule
\multirow{5}{*}{DeBERTaV3}     & FedFT                                                 & 184M                                   & \textbf{65.81$_{\pm.3}$} \slash\ \textbf{81.12$_{\pm.2}$}                 & \textbf{66.39$_{\pm.3}$} \slash\ \textbf{81.15$_{\pm.3}$}                 & \textbf{55.11$_{\pm.2}$} \slash\ \textbf{66.46$_{\pm.1}$}                 & \textbf{56.15$_{\pm.1}$} \slash\ \textbf{67.34$_{\pm.1}$}                                                  \\
                                & FedBF                                                 & 0.1M                                           & 53.51$_{\pm.4}$ \slash\ 71.32$_{\pm.6}$                   & 57.49$_{\pm.1}$ \slash\ 74.81$_{\pm.3}$                   & 23.03$_{\pm2.3}$ \slash\ 34.19$_{\pm.6}$                   & 30.72$_{\pm1.4}$ \slash\ 43.34$_{\pm.5}$                                                  \\
                                & FedAP                                                 & 1.8M                                                    & \uline{63.32$_{\pm.2}$} \slash\ 79.21$_{\pm.1}$        & 65.35$_{\pm.1}$ \slash\ \uline{80.76$_{\pm.1}$}        & 52.32$_{\pm.3}$ \slash\ 64.31$_{\pm.1}$        & 55.56$_{\pm.5}$ \slash\ 66.65$_{\pm.3}$                                                   \\
                                & FedLR                                                 & 1.2M                                           & 61.05$_{\pm.2}$ \slash\ 77.78$_{\pm.3}$                   & 63.84$_{\pm.3}$ \slash\ 79.25$_{\pm.2}$                   & 50.45$_{\pm.2}$ \slash\ 62.56$_{\pm.2}$                   & 53.71$_{\pm.1}$ \slash\ 65.31$_{\pm.1}$                                                  \\
                                & \textbf{FeDeRA}                                       & 1.2M                                           & 63.24$_{\pm.4}$ \slash\ \uline{79.27$_{\pm.4}$}           & \uline{65.54$_{\pm.4}$} \slash\ 80.59$_{\pm.1}$                   & \uline{53.64$_{\pm.1}$} \slash\ \uline{65.96$_{\pm.1}$}                  & \uline{55.92$_{\pm.4}$} \slash\ \uline{67.14$_{\pm.2}$}                                                  \\
\bottomrule[2pt]
\end{tabular}}
\end{table}

\subsection{Training Efficiency}
\label{subsec:efficiency evaluation} 
We then evaluate the training efficiency of our proposed method in terms of time cost to reach a targeted accuracy in a practical scenario, where we use Jetson AGX Orin as clients, and a server equipped with 8$\times$NVIDIA RTX A6000 GPU and 2$\times$64-Core AMD EPYC 7763 CPU as the central node. These devices are connected via WiFi with bandwidth ranging from 20-30 Mbps. We measure the time cost by different methods to reach 90\%, 95\%, and 99\% of target metrics, as shown in Table \ref{Tab:efficiency}. From this table, we can observe that the proposed FeDeRA achieves fastest convergence, reducing training time by up to 96.9\% compared to that of FedFT. These results demonstrate that FeDeRA can significantly improve training efficiency in practical FL systems.

\begin{table}[htbp]
\caption{Comparison on training efficiency by different methods, measured by time required to achieve 90\%, 95\%, and 99\% of the target metrics, which are accuracy of 0.8 for text classification on 20Newsgroup, F1 score of 0.5 for named entity recognition on WNUT2017, EM score of 0.65 and F1 score of 0.8 for question answering on SQuADv1.1. ``-'' indicates that the target metric cannot be achieved by the corresponding method under the considered settings. The time cost is measured in hours. \vspace{1mm}}
\label{Tab:efficiency}
\resizebox{1\textwidth}{!}{
\begin{tabular}{c|c|ccccccccc} 
\toprule[2pt]
\multirow{2}{*}{\textbf{Model}} & \multirow{2}{*}{\textbf{Method}} & \multicolumn{3}{c}{\textbf{20NEWS(Acc.)}}                       & \multicolumn{3}{c}{\textbf{WNUT(F1)}}                                    & \multicolumn{3}{c}{\textbf{SQuADv1.1(EM/F1)}}                                            \\
                                &                                  & \textbf{90\%} & \textbf{\textbf{95\%}} & \textbf{\textbf{99\%}} & \textbf{\textbf{90\%}} & \textbf{\textbf{95\%}} & \textbf{\textbf{99\%}} & \textbf{\textbf{90\%}}      & \textbf{\textbf{95\%}}      & \textbf{\textbf{99\%}}       \\ 
\midrule
\multirow{5}{*}{RoBERTa}        & FedFT                            & 5.60          & 6.53                   & 13.19                  & 5.22                   & 7.04                   & 10.38                  & 1.39 \slash\ 0.84                 & 2.31\slash\ 1.96                  & 5.02\slash\ 5.64                  \\
                                & FedBF                            & 0.78          & -                    & -                     & 0.89                   & -                      & -                     & - \slash\ -                         & - \slash\ -                         & - \slash\ -                          \\
                                & FedAP                            & 0.30          & 0.49                   & 0.99                   & 0.30                   & 0.44                   & 0.95                   & 0.21 \slash\ 0.12                   & 0.39 \slash\ 0.34                   & 0.65 \slash\ 0.73                    \\
                                & FedLR                            & 0.33          & 0.59                   & 1.21                   & 0.27                   & 0.39                   & 0.72                   & 0.45 \slash\ 0.24                   & 0.85 \slash\ 0.84                   & 1.43 \slash\ 1.79                    \\
                                & FeDeRA                           & \textbf{0.17} & \textbf{0.30}          & \textbf{0.54}          & \textbf{0.21}          & \textbf{0.28}          & \textbf{0.32}          & \textbf{0.16} \slash\ 
\textbf{0.09} & \textbf{0.31} \slash\ \textbf{0.28} & \textbf{0.51} \slash\ \textbf{0.52}  \\ 
\midrule
\multirow{5}{*}{DeBERTaV3}      & FedFT                            & 13.27         & 16.91                  & 23.84                  & 6.48                   & 9.14                   & 12.96                  & 0.67 \slash\ 0.48                   & 0.96 \slash\ 0.67                   & 1.62 \slash\ 1.34                    \\
                                & FedBF                            & -            & -                     & -                      & 0.63                   & 1.27                   & -                      & 1.13 \slash\  0.50                   & - \slash\ -                          & - \slash\ -                           \\
                                & FedAP                            & 1.13          & 1.74                   & -                      & 0.69                   & 1.40                   & -                      & 0.22 \slash\ 0.15                   & 0.36 \slash\ 0.25                   & 0.67 \slash\ 0.49                    \\
                                & FedLR                            & 1.03          & 1.43                   & 1.80                   & 0.81                   & 0.85                   & -                      & 0.33 \slash\ 0.18                   & 0.55 \slash\ 0.34                   & 1.12 \slash\ 0.77                    \\
                                & FeDeRA                           & \textbf{0.36} & \textbf{0.45}          & \textbf{0.64}          & \textbf{0.16}          & \textbf{0.32}          & \textbf{0.45}          & \textbf{0.14} \slash\ \textbf{0.09} & \textbf{0.24} \slash\ \textbf{0.19} & \textbf{0.49} \slash\  \textbf{0.31}  \\
\bottomrule[2pt]
\end{tabular}}
\end{table}

\subsection{Impact of Data Heterogeneity}
\begin{figure}[!t]
\centering
\begin{subfigure}{.33\textwidth}
  \centering
  \includegraphics[width=1\linewidth]{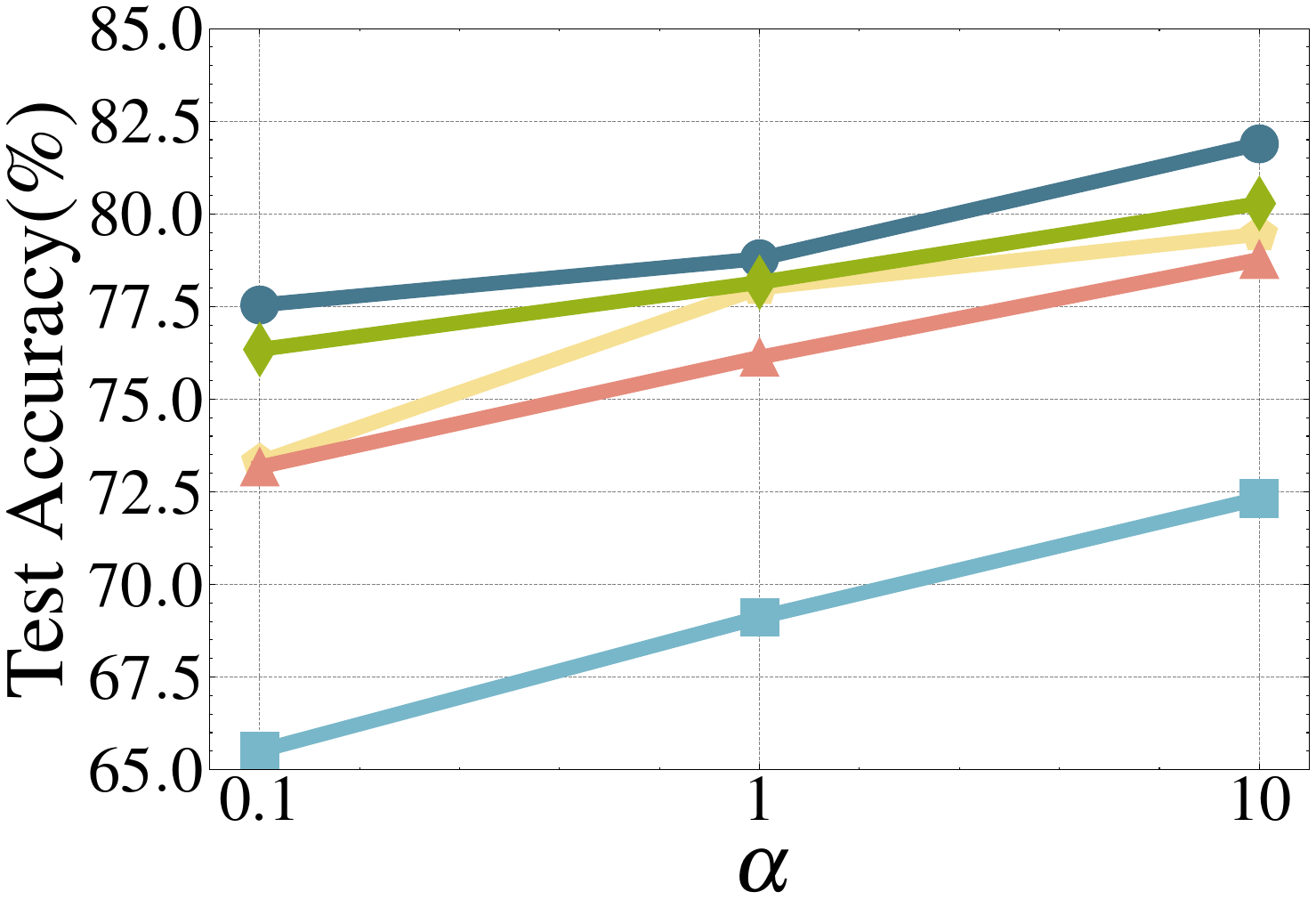}
  \caption{Comm. rounds 200}
\end{subfigure}%
\begin{subfigure}{.33\textwidth}
  \centering
  \includegraphics[width=1\linewidth]{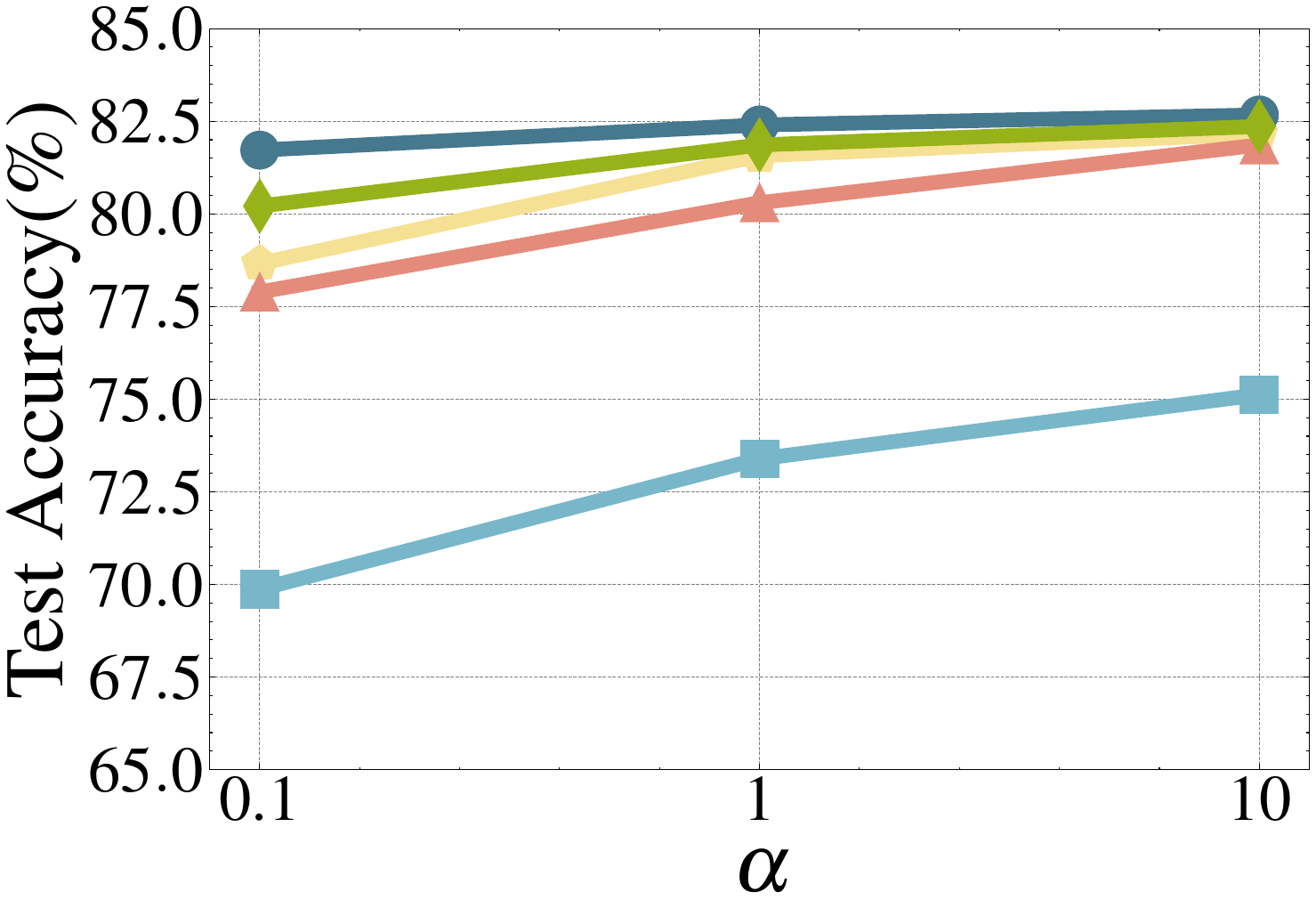}
  \caption{Comm. rounds 500}
\end{subfigure}
\begin{subfigure}{.33\textwidth}
  \centering
  \includegraphics[width=1\linewidth]{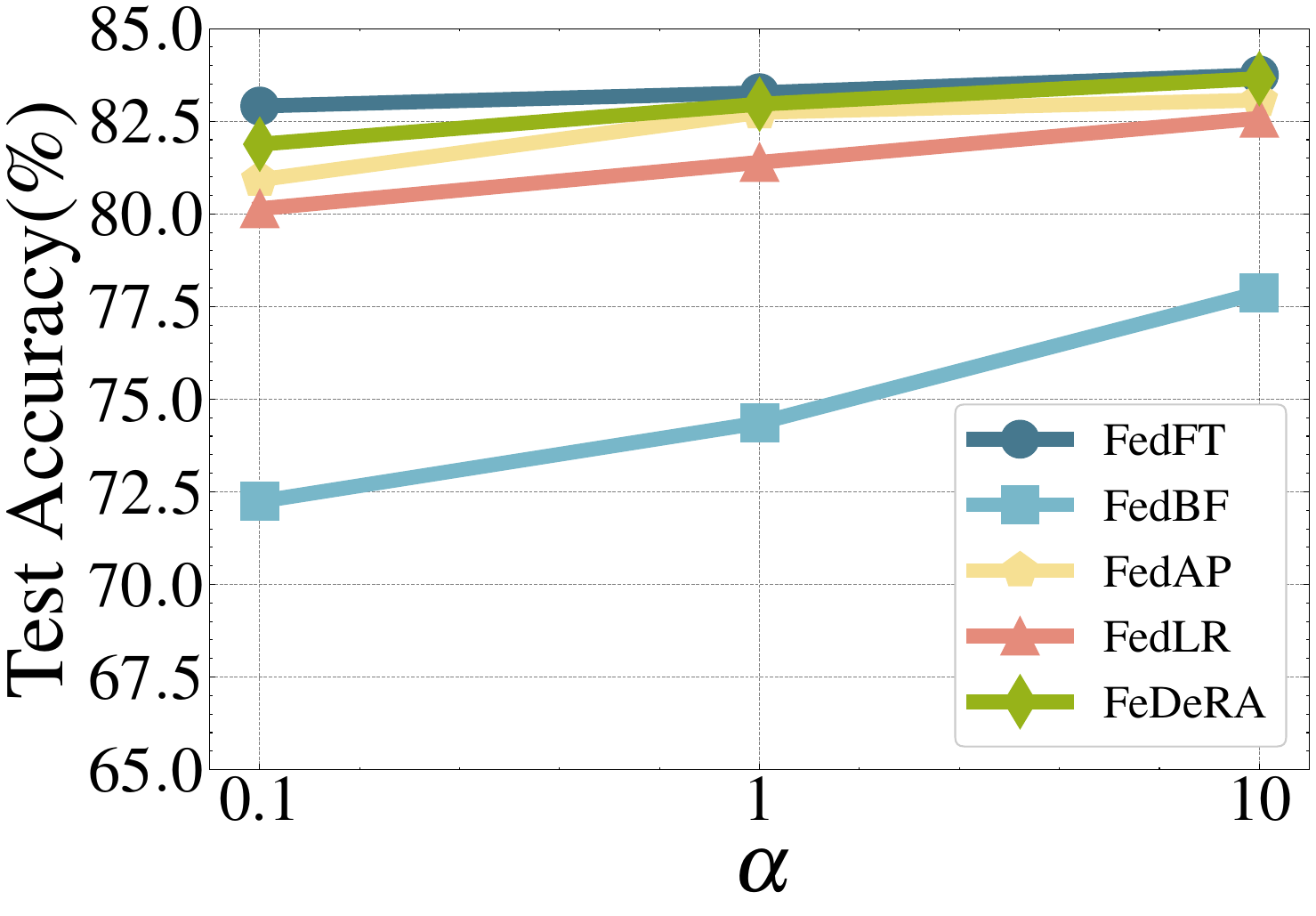}
  \caption{Comm. rounds 1000}
\end{subfigure}
\caption{Performance comparison on test accuracy with regards to different levels of data heterogeneity. 
\vspace{-3mm}}
\label{fig:non-iid}
\end{figure}

 In this section, we investigate the impact of non-IID on the training accuracy by changing the value of $\alpha$, as shown in \autoref{fig:non-iid}. We use RoBERTa-base model and the 20Newsgroup dataset. It can be observed that as $\alpha$ decreases, i.e., the data heterogeneity increases, the performance of all methods deteriorates. However, the performance of FedFT stays stable compared to other approach, showing its robustness to data heterogeneity in a FL setting by updating all parameters. The proposed FeDeRA also suffers much less performance degradation compared to others with significantly fewer trainable parameters than that of FedFT. These results further demonstrate the effectiveness of the proposed FeDeRA in handling data heterogeneity.

\subsection{Impact of Trainable Parameters Budgets}
\begin{figure}[htbp]
\centering
\begin{subfigure}{.38\textwidth}
  \centering
  \includegraphics[width=1\linewidth]{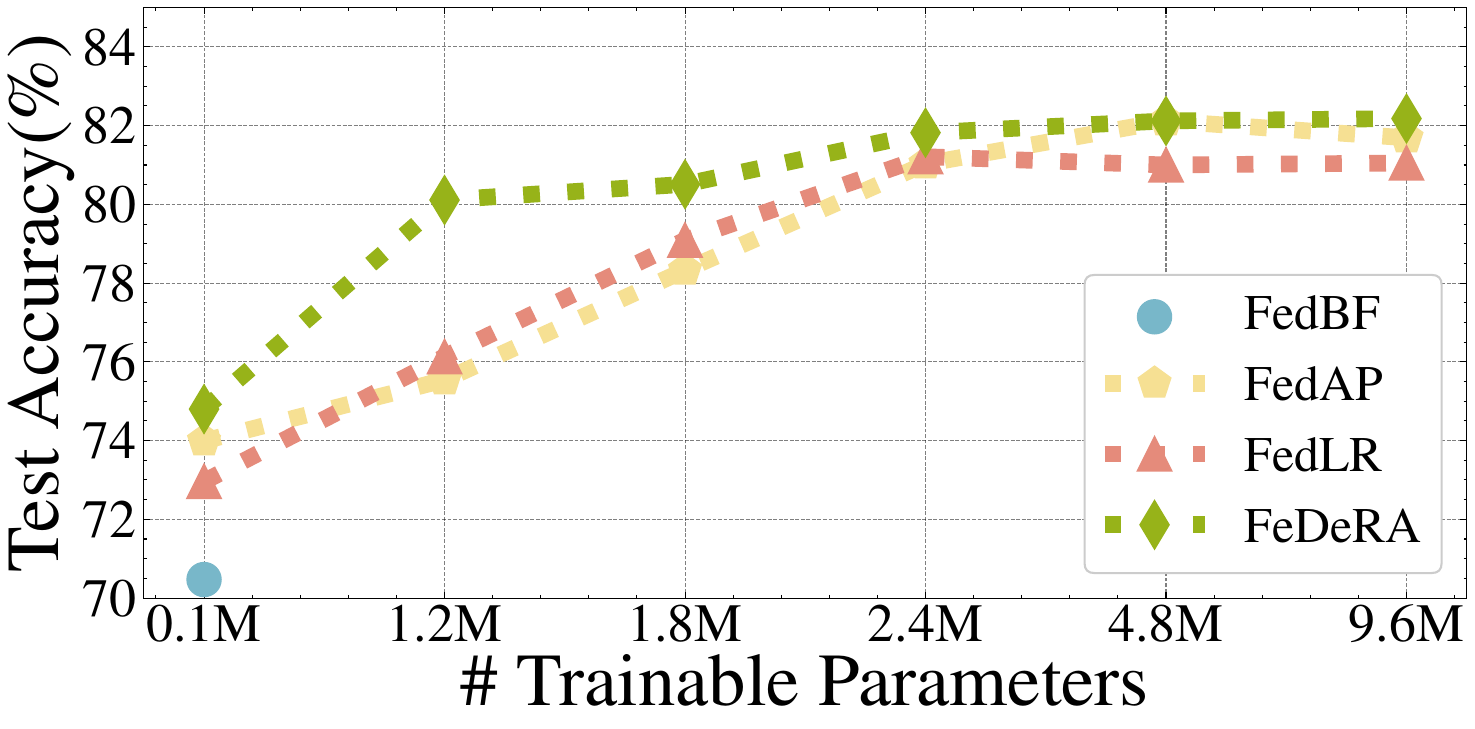}
  \caption{Acc. under various parameters budgets}
  \label{subfig:params}
\end{subfigure}%
\begin{subfigure}{.285\textwidth}
  \centering
  \includegraphics[width=1\linewidth]{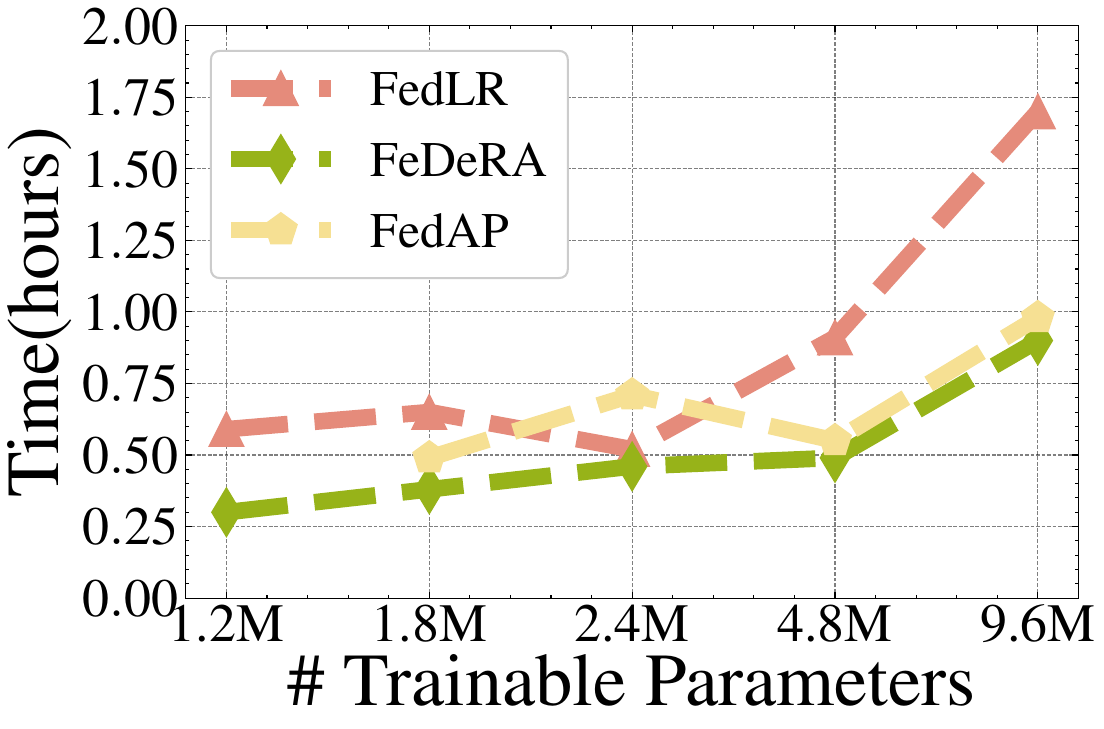}
  \caption{Target accuracy 0.76}
  \label{subfig:params_0.76}
\end{subfigure}
\begin{subfigure}{.285\textwidth}
  \centering
  \includegraphics[width=1\linewidth]{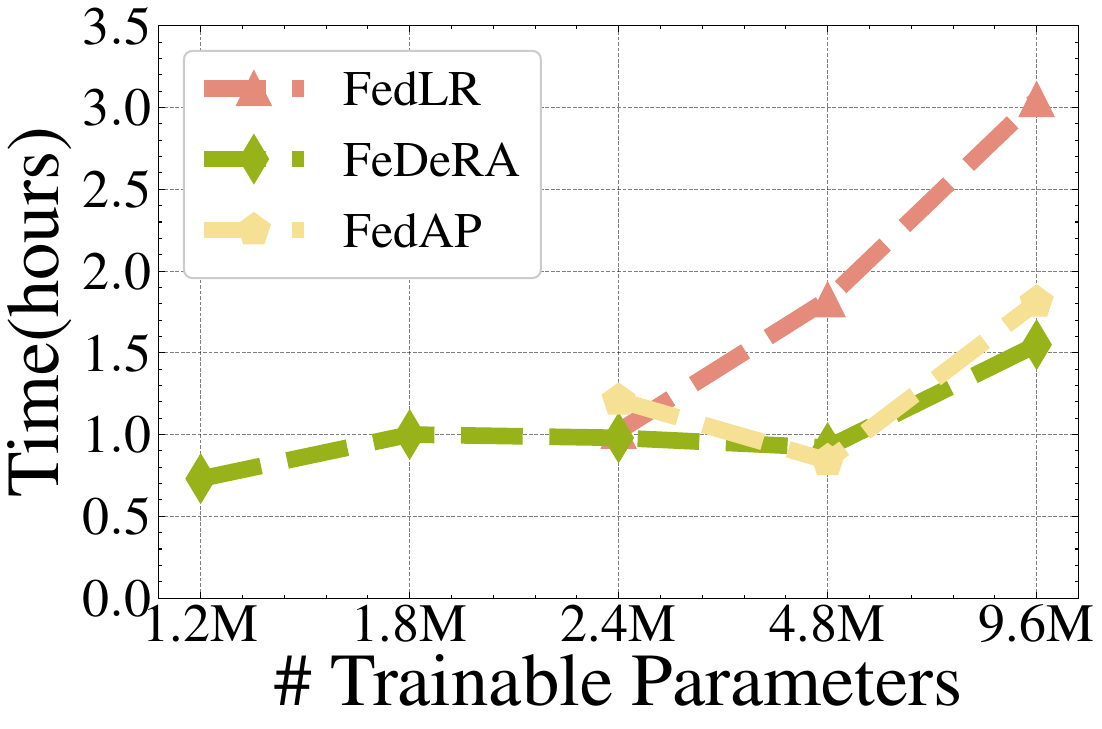}
  \caption{Target accuracy 0.8}
  \label{subfig:params_0.8}
\end{subfigure}
\caption{Performance comparison on test accuracy and training time cost with regards to different trainable parameter budgets. \vspace{-3mm}}
\label{fig:params}
\end{figure}

We further evaluate the performance of different methods in terms of test accuracy and training time with regards to different trainable parameters budgets, as shown in \autoref{fig:params}. We use the RoBERTa-base model and the 20Newsgroup dataset,  and set $\alpha$ to 1 and total training round to 500 for the experiments presented in \autoref{subfig:params}. It can be seen from \autoref{subfig:params} that the proposed FeDeRA notably outperforms the three benchmarks in terms of test accuracy across a wide range of trainable parameters. Moreover, in \autoref{subfig:params_0.76} and \autoref{subfig:params_0.8}, we set the target accuracy to $0.76$ and $0.8$, respectively, and observe that the proposed FeDeRA also requires  less training time under various trainable parameters budgets. This further demonstrates the effectiveness of the proposed FeDeRA in the FL setting.

\section{Conclusions}
In this paper, we introduced FeDeRA, a method that extends LoRA to a federated learning setting, and performs Singular Value Decomposition on the pre-trained weight matrices to initialize the low-rank adapter at each client. We analyzed how FeDeRA accelerates and improves fine-tuning of transformer-based models by showing the magnitude and direction curves of weight updates during training. Our empirical findings demonstrated that FeDeRA not only outperforms the existing PEFT method in terms of task performance and training efficiency but also achieved comparable or even better performance than full-parameter fine-tuning. Furthermore, FeDeRA is more robust to data heterogeneity than all other PEFT methods, showing it is more applicable to federated learning systems. 

\section{Limitations and Future Works}
Because of privacy constraints, we couldn't gather naturally heterogeneous datasets (like personal conversations) from the open world for experimentation. Instead, we relied on simulated data distributions created from public datasets. Additionally, simulating the prevalent device heterogeneity found in real-world environments is challenging, making it difficult to examine the efficacy and superior performance of our proposed approach in a heterogeneous FL system.

\bibliographystyle{unsrt}
\bibliography{Ref.bib}
\clearpage
\appendix
\section{Non-IID Data Partition}
\label{appendix:data partition}

\subsection{Dirichlet Distribution}
The Dirichlet distribution, commonly referred to as the multivariate Beta distribution, represents a class of high-dimensional continuous distributions. Its support is based on the standard simplex within the realm of positive real numbers. This distribution serves as an extension to the Beta distribution, generalizing it to higher dimensions. The Dirichlet distribution, delineated by a parameter measure denoted as $\mathbf{u}$, can be articulated through its corresponding probability density function as:
\begin{equation*}
    f(\boldsymbol{x};\boldsymbol{u})=\frac{1}{B(\boldsymbol{u})}\prod_{i=1}^{K}x_i^{u_i-1},
\end{equation*}
where $B(\boldsymbol{u})$  is the multivariate beta function:
\begin{equation*}
    B(\boldsymbol{u})=\frac{\prod_{i=1}^{K}\Gamma(u_i)}{\Gamma(\sum_{i=1}^{K}u_i)},
\end{equation*}
and $\Gamma(\cdot)$ means Gamma function. In which, $\boldsymbol{u}$ is typically expressed as $\boldsymbol{u}=\alpha\boldsymbol{m},\sum_{i=1}^{K}m_i=1,m_i>0$. If random variable $X\sim Dir(\alpha\boldsymbol{m})$, $\boldsymbol{m}$ is the mathematic expectation of $X$:
\begin{equation*}
    \mathbb{E}(X)=\int f(\boldsymbol{x};\boldsymbol{u})\boldsymbol{x} \, \mathrm{d}\boldsymbol{x}=\boldsymbol{m}.
\end{equation*}
The magnitude of $\alpha$ will determine the degree of similarity between the distribution obtained from each sampling and the original distribution; a larger $\alpha$ will yield a distribution that is more similar to the original.

\subsection{Classes of Datasets}
We defined a specific number of classes for each dataset to facilitate the partition of the datasets. The following is the detailed information.

\textbf{20Newsgroup.} The 20Newsgroups dataset comprises around 18000 newsgroups posts on 20 topics and the topics cover a range of subjects from politics to religion to sports to science. The 20 Newsgroups dataset is employed for text classification tasks; consequently, its original labels are utilized as its classes.

\textbf{SemEval2010Task8.}  SemEval2010-Task8 is aimed at addressing the challenge of Multi-Way Classification of Semantic Relations Between Pairs of Nominals. The task involves identifying and categorizing the semantic relations between specific pairs of nominals within given sentences. It delineates nine distinct types of semantic relations, with each relation type being subject to inversion, thereby totaling 19 potential labels when including an additional 'Other' category for relations that do not conform to any of the predefined types. We also utilize the original labels as classes.

\textbf{WNUT2017.} The WNUT2017 dataset refers to the data collection used for the Workshop on Noisy User-generated Text (WNUT) in its 2017 edition. This workshop series focuses on processing and understanding noisy user-generated text, often encountered in social media, online forums, and other web sources. WNUT2017 specifically aimed to address several key tasks in noisy text processing. One of the primary tasks was Named Entity Recognition (NER), which involved identifying and classifying proper names within text into predefined categories such as persons, organizations, locations, etc. The token labels comprise seven parts of speech, further divided into begin and intermediate positions, along with an additional category that does not belong to any of the aforementioned parts of speech, totaling thirteen labels in all. We observe that each training sample may comprise a unique set of token labels, thereby contributing to the learning capability across diverse token labels. Consequently, we catalogue the total types of token labels encompassed within each training sample as its class, as detailed in \autoref{Tab:examples of WNUT2017}.

\begin{table}
\centering
\caption{Examples of constructing classes for the WNUT2017 dataset. \vspace{2mm}}
\label{Tab:examples of WNUT2017}
\resizebox{1\textwidth}{!}{
\begin{tabular}{llc} 
\toprule[2pt]
\multicolumn{1}{c}{\textbf{Tokens}}                                                                                                                                                                                                                                                                                                                                                                                     & \multicolumn{1}{c}{\textbf{Token Labels}}                                                                                                                                                                                                                                                                                                                                                                                                                                                                                                                                                                                                                                                                                                                                                                                                           & \textbf{Class}                                                                                                                                                                             \\ 
\midrule
\textcolor[rgb]{0.067,0.094,0.153}{[ "today", "is", "my", "last", "day", "at", "the", "office", "." ]}                                                                                                                                                                                                                                                                                                                  & \textcolor[rgb]{0.067,0.094,0.153}{[ O, O, O, O, O, O, O, O, O ]}                                                                                                                                                                                                                                                                                                                                                                                                                                                                                                                                                                                                                                                                                                                                                                                   & O                                                                                                                                                                                          \\ 
\midrule
\begin{tabular}[c]{@{}l@{}}\textcolor[rgb]{0.067,0.094,0.153}{[ "Pxleyes", "Top", "50", "Photography", "Contest", }\\\textcolor[rgb]{0.067,0.094,0.153}{"Pictures", "of",~ "August", "2010", "...", "http://bit.ly/bgCyZ0", }\\\textcolor[rgb]{0.067,0.094,0.153}{"\#photography" ]}\end{tabular}                                                                                                                       & \begin{tabular}[c]{@{}l@{}}\textcolor[rgb]{0.067,0.094,0.153}{[}\textcolor[rgb]{0.122,0.161,0.216}{B-corporation}\textcolor[rgb]{0.067,0.094,0.153}{, O, O, O, O, O, }\\\textcolor[rgb]{0.067,0.094,0.153}{O, O, O, O, O, O ]}\end{tabular}                                                                                                                                                                                                                                                                                                                                                                                                                                                                                                                                                                                                         & \textcolor[rgb]{0.122,0.161,0.216}{B-corporation}                                                                                                                                        \\ 
\midrule
{[} "Toy", "story", "3", "tonight", "on", "the", "lawn", "!" ]                                                                                                                                                                                                                                                                                                                                                          & \begin{tabular}[c]{@{}l@{}}\textcolor[rgb]{0.067,0.094,0.153}{[}\textcolor[rgb]{0.122,0.161,0.216}{B-creative-work}\textcolor[rgb]{0.067,0.094,0.153}{,~}\textcolor[rgb]{0.122,0.161,0.216}{I-creative-work}\textcolor[rgb]{0.067,0.094,0.153}{}\\\textcolor[rgb]{0.067,0.094,0.153}{,}\textcolor[rgb]{0.122,0.161,0.216}{I-creative-work}\textcolor[rgb]{0.067,0.094,0.153}{, O, O, O, O, O ]}\end{tabular}                                                                                                                                                                                                                                                                                                                                                                                                                                        & \textcolor[rgb]{0.122,0.161,0.216}{B-creative-work}\textcolor[rgb]{0.067,0.094,0.153}{-}\textcolor[rgb]{0.122,0.161,0.216}{I-creative-work}                                              \\ 
\midrule
\begin{tabular}[c]{@{}l@{}}\textcolor[rgb]{0.067,0.094,0.153}{[ "(", "via", "POPSUGAR", ")", "Sarah", "Jessica", "Parker", }\\\textcolor[rgb]{0.067,0.094,0.153}{"and", "Gwen", "Stefani", "Wrap", "Up", "Another", "Successful", }\\\textcolor[rgb]{0.067,0.094,0.153}{"New", "York", "Fashion", "Week", ":", "New", "York", "Fa", "...", }\\\textcolor[rgb]{0.067,0.094,0.153}{"http://bit.ly/aMaJNB" ]}\end{tabular} & \begin{tabular}[c]{@{}l@{}}\textcolor[rgb]{0.067,0.094,0.153}{[ O, O, O, O,}\textcolor[rgb]{0.122,0.161,0.216}{B-person}\textcolor[rgb]{0.067,0.094,0.153}{,~}\textcolor[rgb]{0.122,0.161,0.216}{I-person}\textcolor[rgb]{0.067,0.094,0.153}{}\\\textcolor[rgb]{0.067,0.094,0.153}{,~}\textcolor[rgb]{0.122,0.161,0.216}{I-person}\textcolor[rgb]{0.067,0.094,0.153}{, O,}\textcolor[rgb]{0.122,0.161,0.216}{B-person}\textcolor[rgb]{0.067,0.094,0.153}{,~}\textcolor[rgb]{0.122,0.161,0.216}{I-person}\textcolor[rgb]{0.067,0.094,0.153}{, }\\\textcolor[rgb]{0.067,0.094,0.153}{O, O, O, O, O, O, O, O, O,~}\textcolor[rgb]{0.122,0.161,0.216}{B-location}\textcolor[rgb]{0.067,0.094,0.153}{}\\\textcolor[rgb]{0.067,0.094,0.153}{,~}\textcolor[rgb]{0.122,0.161,0.216}{I-location}\textcolor[rgb]{0.067,0.094,0.153}{, O, O, O ]}\end{tabular} & \textcolor[rgb]{0.122,0.161,0.216}{B-person-}\textcolor[rgb]{0.122,0.161,0.216}{I-person-}\textcolor[rgb]{0.122,0.161,0.216}{B-location-I}\textcolor[rgb]{0.122,0.161,0.216}{-location}  \\
\bottomrule[2pt]
\end{tabular}}
\end{table}

\textbf{PLONER.} PLONER, the abbreviation for Person, Location, Organization Named Entity Recognition, aims to assess the cross-domain generalization capability. It involves selecting samples from representative datasets that include at least one of the three entity types: person, location, and organization. The representative datasets encompass CoNLL2003, OntoNotes 5.O, and WNUT2016. The English data of CoNLL2003 is a collection of news wire articles from the Reuters Corpus. The data for WNUT2016 were primarily collected from Twitter. The OntoNotes5.0 corpus is categorized into several types, which include newswire (News), broadcast news (BN), broadcast conversation (BC), telephone conversation (Tele), and web data (Web). The corpora across various themes inherently exhibit heterogeneity. Hence, while adopting the methodology used for class construction in WNUT2017, we also take into account the source dataset of the samples for further distinction, details are shown in \autoref{Tab:examples of PLONER}.

\begin{table}
\centering
\caption{Examples of constructing classes for the PLONER dataset. \vspace{2mm}}
\label{Tab:examples of PLONER}
\resizebox{1\textwidth}{!}{
\begin{tabular}{llcc} 
\toprule[2pt]
\multicolumn{1}{c}{\textbf{Tokens}}                                                                                                                                                                                                                                     & \multicolumn{1}{c}{\textbf{Token Labels}}                                                                                                                                                                                                                                                                                                                                                                                                                                                                                                                  & \textbf{Source Dataset}          & \textbf{Class}                                                                                                                                                                                                               \\ 
\midrule
\textcolor[rgb]{0.067,0.094,0.153}{[ "only", "France", "and", "Britain", "backend", "Fischler", "'s", "proposal", "." ]}                                                                                                                                                & \begin{tabular}[c]{@{}l@{}}\textcolor[rgb]{0.067,0.094,0.153}{[ O,~}\textcolor{blue}{B-LOC}\textcolor[rgb]{0.067,0.094,0.153}{, O,~}\textcolor{blue}{B-LOC}\textcolor[rgb]{0.067,0.094,0.153}{}\\\textcolor[rgb]{0.067,0.094,0.153}{, O,~}\textcolor{blue}{B-PER}\textcolor[rgb]{0.067,0.094,0.153}{, O, O, O ]}\end{tabular}                                                                                                                                                                                                                              & \textcolor{red}{CoNLL2023}       & \textcolor{red}{CoNLL-}\textcolor{blue}{B-LOC}-\textcolor{blue}{B-PER}                                                                                                                                                       \\ 
\midrule
\begin{tabular}[c]{@{}l@{}}\\\textcolor[rgb]{0.18,0.196,0.22}{["Cowboys", "on", "a", "3rd", "and", "10", "finally", "get", "their", "1st", }\\\textcolor[rgb]{0.18,0.196,0.22}{"down", "in", "the", "game", "."]}\end{tabular}                                          & \begin{tabular}[c]{@{}l@{}}\textcolor[rgb]{0.18,0.196,0.22}{[ }\textcolor{blue}{B-ORG}\textcolor[rgb]{0.18,0.196,0.22}{, O, O, O, O, O, }\\\textcolor[rgb]{0.18,0.196,0.22}{O, O, O, O, O, O, O, O, O ]}\end{tabular}                                                                                                                                                                                                                                                                                                                                      & \textcolor{red}{WNUT2016}        & \textcolor{red}{WNUT}-\textcolor{blue}{B-ORG}                                                                                                                                                                                \\ 
\midrule
\textcolor[rgb]{0.18,0.196,0.22}{["As", "you", "know", "Lebanon", "is", "an", "agreement", "-", "based", "democracy", "."]}                                                                                                                                             & \begin{tabular}[c]{@{}l@{}}\textcolor[rgb]{0.18,0.196,0.22}{[ O, O, O, }\textcolor{blue}{B-LOC}\textcolor[rgb]{0.18,0.196,0.22}{, O, O, O, }\\\textcolor[rgb]{0.18,0.196,0.22}{O, O, O, O ]}\end{tabular}                                                                                                                                                                                                                                                                                                                                                  & \textcolor{red}{OntoNotes5.0-BN} & \textcolor{red}{OnBN}-\textcolor{blue}{B-LOC}                                                                                                                                                                                \\ 
\midrule
\begin{tabular}[c]{@{}l@{}}\textcolor[rgb]{0.18,0.196,0.22}{["In", "contrast", ",", "Taiwan", "alumni", "are", "noticeable", "by", "their", "absence",}\\\textcolor[rgb]{0.18,0.196,0.22}{~"from", "the", "SAR", "government", "in", "Hong", "Kong", "."]}\end{tabular} & \begin{tabular}[c]{@{}l@{}}\textcolor[rgb]{0.18,0.196,0.22}{[ O, O, O, }\textcolor{blue}{B-LOC}\textcolor[rgb]{0.18,0.196,0.22}{, O, O, O, O,~}\\\textcolor[rgb]{0.18,0.196,0.22}{O, O, O, }\textcolor{blue}{B-ORG}\textcolor[rgb]{0.18,0.196,0.22}{, }\textcolor{blue}{I-ORG}\textcolor[rgb]{0.18,0.196,0.22}{, }\textcolor{blue}{I-ORG}\textcolor[rgb]{0.18,0.196,0.22}{, }\\\textcolor[rgb]{0.18,0.196,0.22}{O, }\textcolor{blue}{B-LOC}\textcolor[rgb]{0.18,0.196,0.22}{, }\textcolor{blue}{I-LOC}\textcolor[rgb]{0.18,0.196,0.22}{, O ]}\end{tabular} & \textcolor{red}{OntoNotes5.0-MZ} & \textcolor{red}{OnMZ}-\textcolor{blue}{B-LOC}\textcolor[rgb]{0.18,0.196,0.22}{-}\textcolor{blue}{I-LOC}\textcolor[rgb]{0.18,0.196,0.22}{-}\textcolor{blue}{B-ORG}\textcolor[rgb]{0.18,0.196,0.22}{-}\textcolor{blue}{I-ORG}  \\
\bottomrule[2pt]
\end{tabular}}
\end{table}

\textbf{SQuADv1.1.} SQuADv1.1 is a question-answering dataset released by Stanford University. The dataset comprises an extensive collection of question-and-answer pairs, with the questions formulated based on paragraphs from Wikipedia articles, and the answers are extracted directly from these respective passages. SQuADv1.1 does not annotate any tags pertinent to the questions, hence, we employ k-means to generate classes for all training samples directly. The number of clusters chosen for this study is 30.

\textbf{MRQA.} The MRQA dataset emphasizes the generalizability of question-answering tasks. This dataset is an amalgamation of subsets from 18 existing QA datasets, carefully curated and transformed into the format as SQuAD. Of these 18 datasets, six are designated for training, another six for development, and the final six are reserved for testing purposes. In our approach, we simply employ the source datasets as the classes.

\subsection{Partitioning Non-IID Datas With Classes}
\label{subsec:partition}
We partition the non-IID data based on the Dirichlet distribution and the classes we have defined. Let $C$ denote the total number of classes, and $c_i$ represent the total number of data instances for the $i$-th class in the undivided dataset. We compute the distribution for classes as $\boldsymbol{m_c}=[m_{c,1}, m_{c,2}\ldots m_{c,C}], m_{c,i}=\frac{c_i}{\sum_{i=1}^{C}c_i}$. Assuming there are N clients, we sample the local data class distribution $X_i=[x_{i,1},x_{i,2}\ldots x_{i,N}]$ for the $i$-th client according to the Dirichlet distribution, denoted as $X_i\sim Dir(\alpha \boldsymbol{m_c})$, $x_{i,j}$ represents the proportion of the $j$-th class on the $i$-th client.

\subsection{Visualization of Data Heterogeneity}
\begin{figure}[!htb]
    \centering
    \begin{subfigure}{.24\textwidth}
        \includegraphics[width=\linewidth]{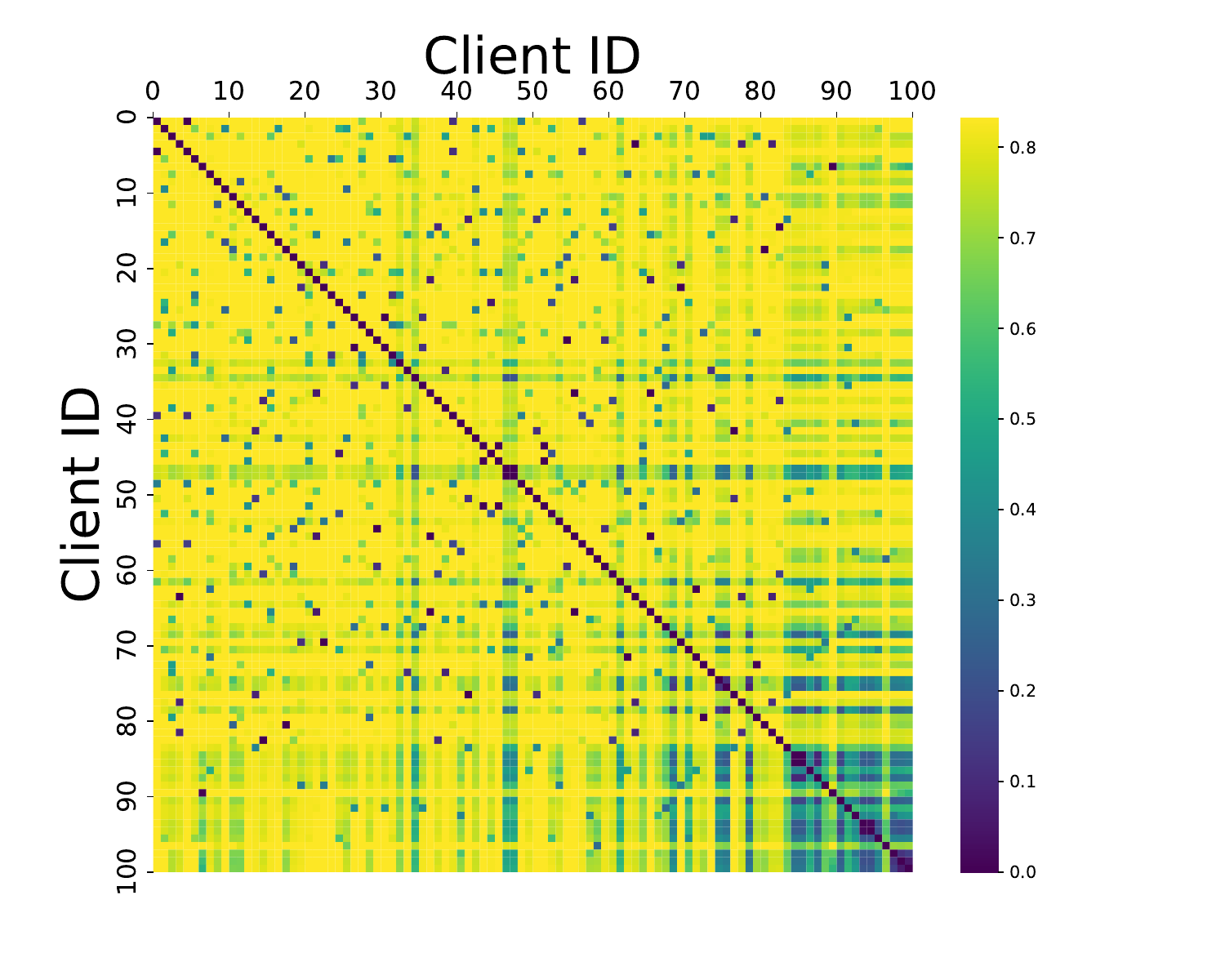}
        \caption{20News $\alpha$=0.1}
    \end{subfigure}\hfill
    \begin{subfigure}{.24\textwidth}
        \includegraphics[width=\linewidth]{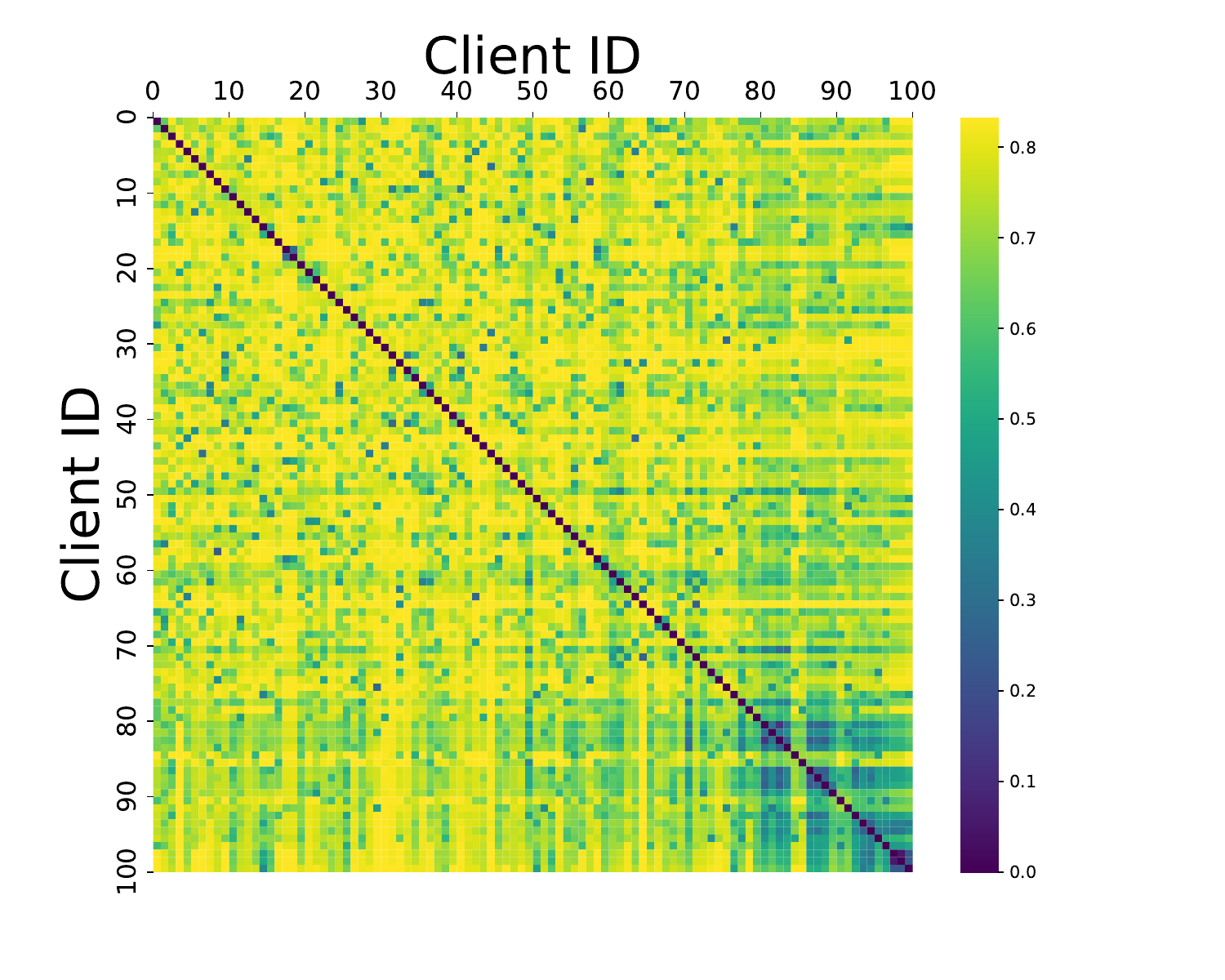}
        \caption{20News $\alpha$=1}
    \end{subfigure}\hfill
    \begin{subfigure}{.24\textwidth}
        \includegraphics[width=\linewidth]{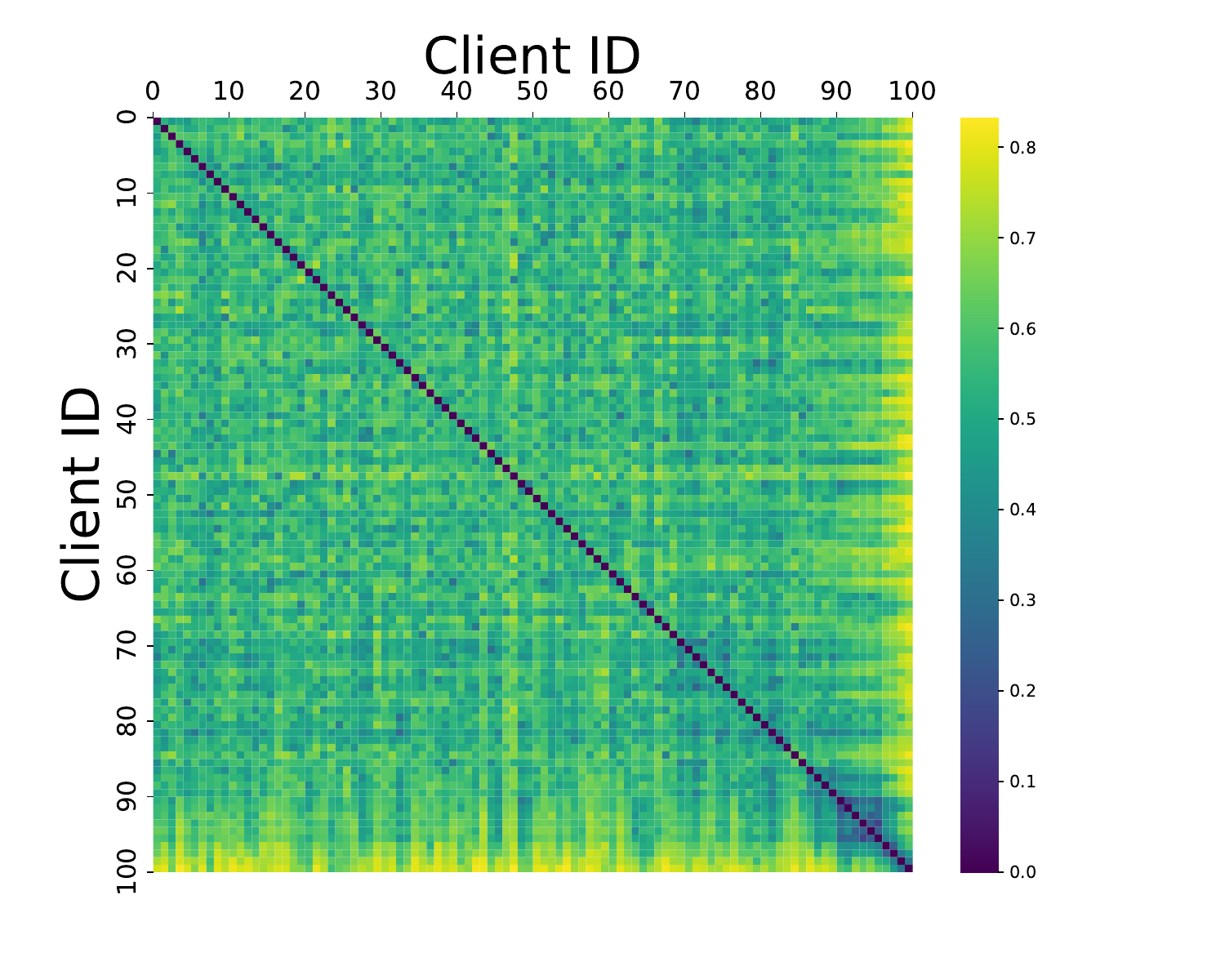}
        \caption{20News $\alpha$=10}
    \end{subfigure}\hfill
    \begin{subfigure}{.24\textwidth}
        \includegraphics[width=\linewidth]{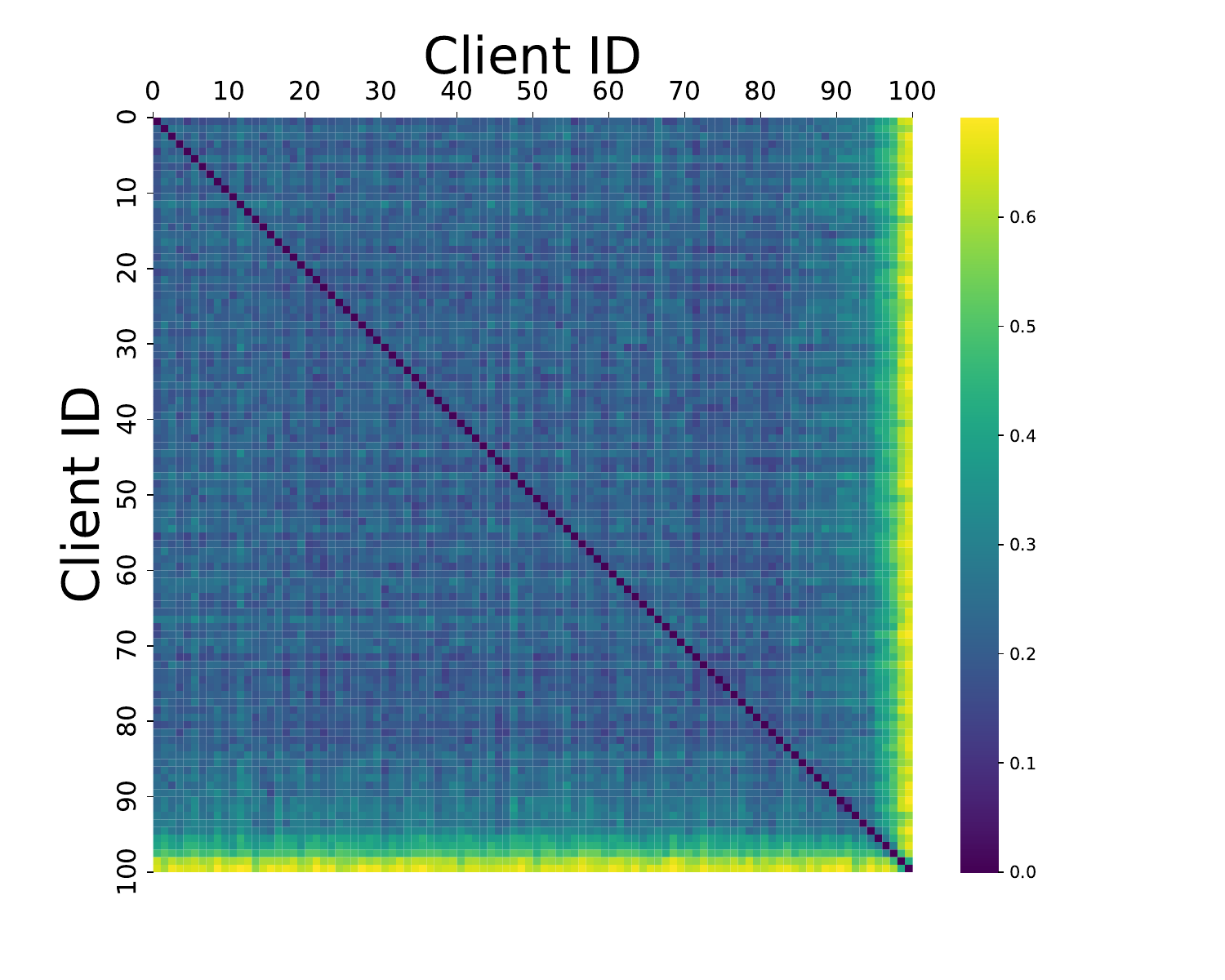}
        \caption{20News $\alpha$=100}
    \end{subfigure}

    \begin{subfigure}{.24\textwidth}
        \includegraphics[width=\linewidth]{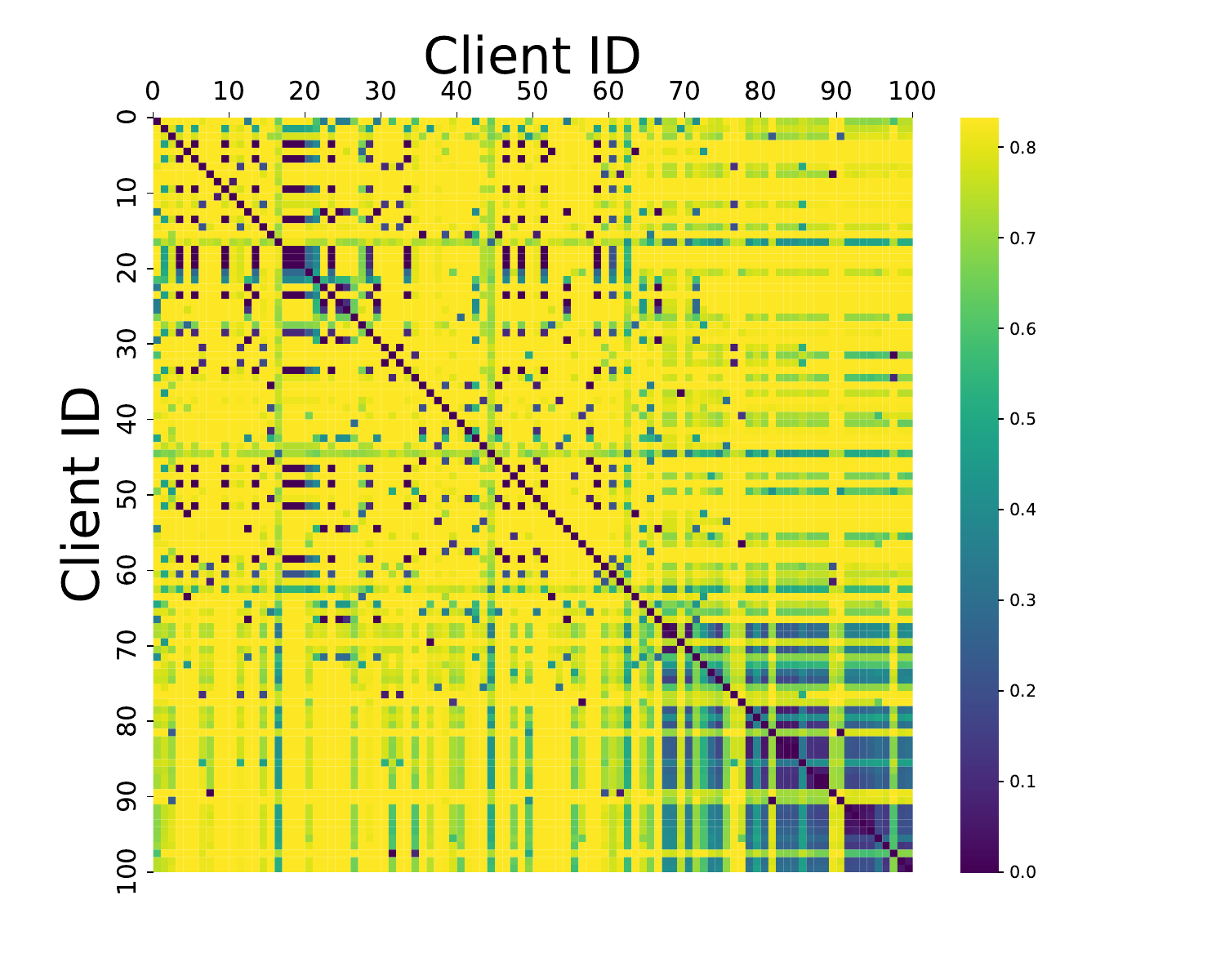}
        \caption{SemEval $\alpha$=0.1}
    \end{subfigure}\hfill
    \begin{subfigure}{.24\textwidth}
        \includegraphics[width=\linewidth]{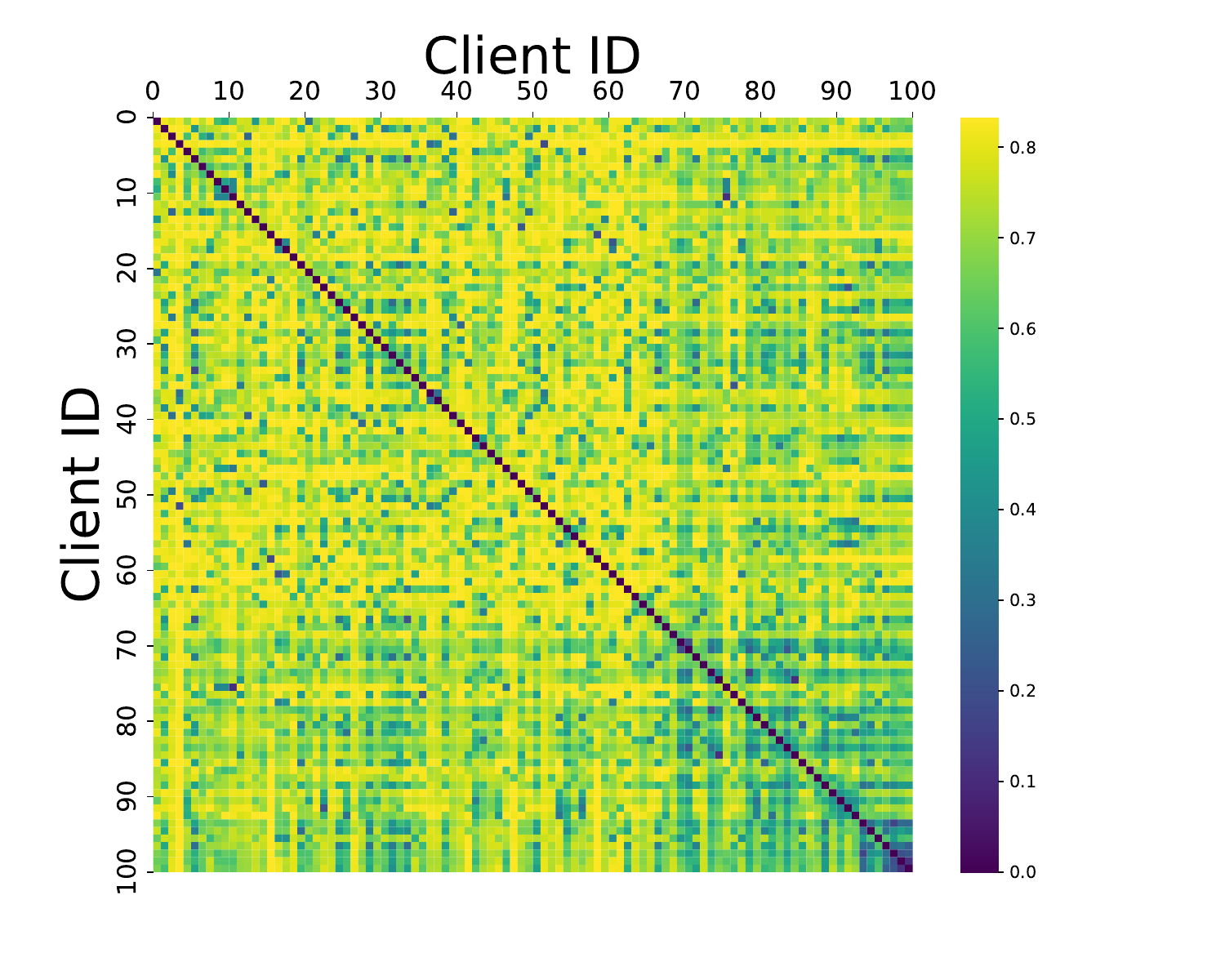}
        \caption{SemEval $\alpha$=1}
    \end{subfigure}\hfill
    \begin{subfigure}{.24\textwidth}
        \includegraphics[width=\linewidth]{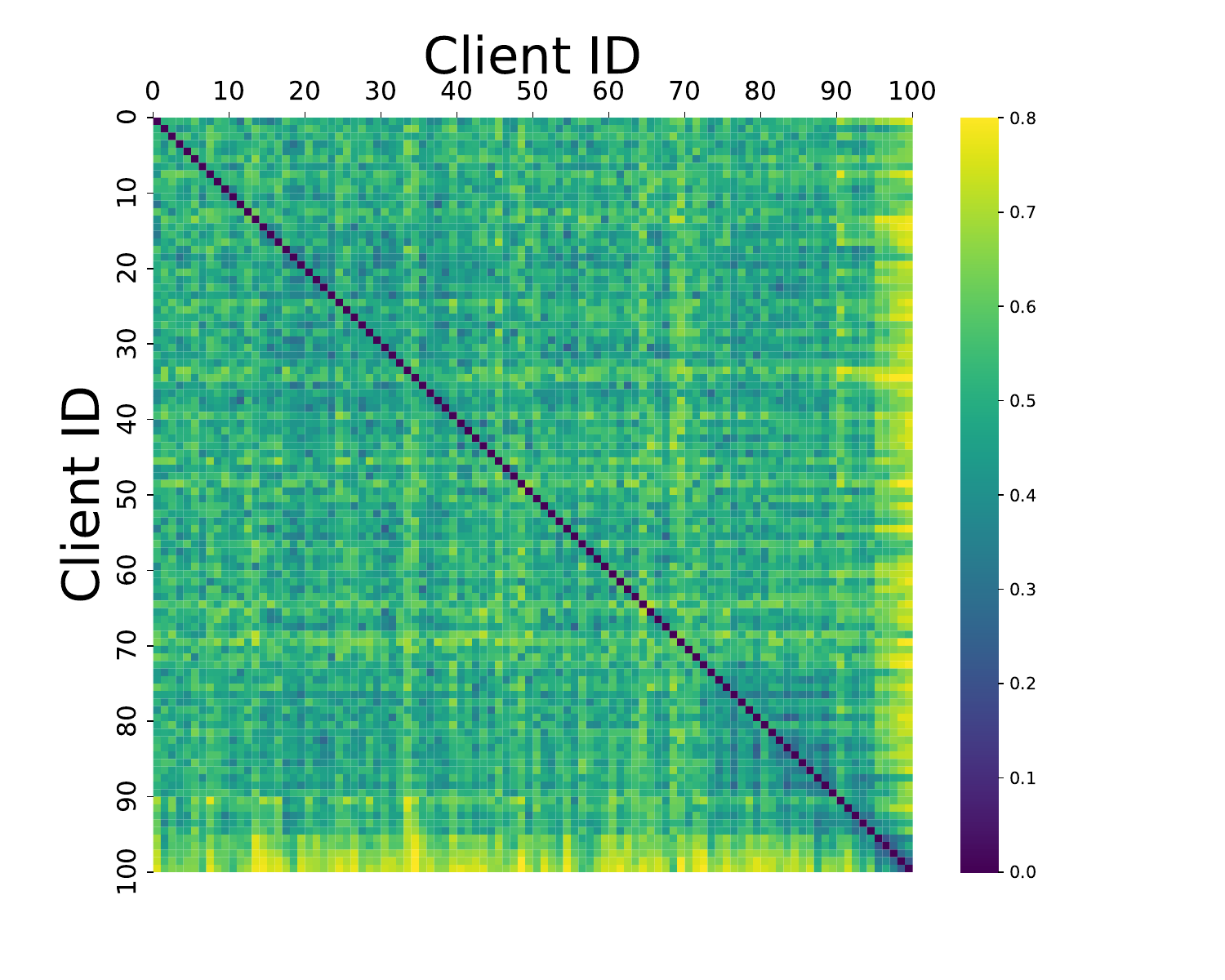}
        \caption{SemEval $\alpha$=10}
    \end{subfigure}\hfill
    \begin{subfigure}{.24\textwidth}
        \includegraphics[width=\linewidth]{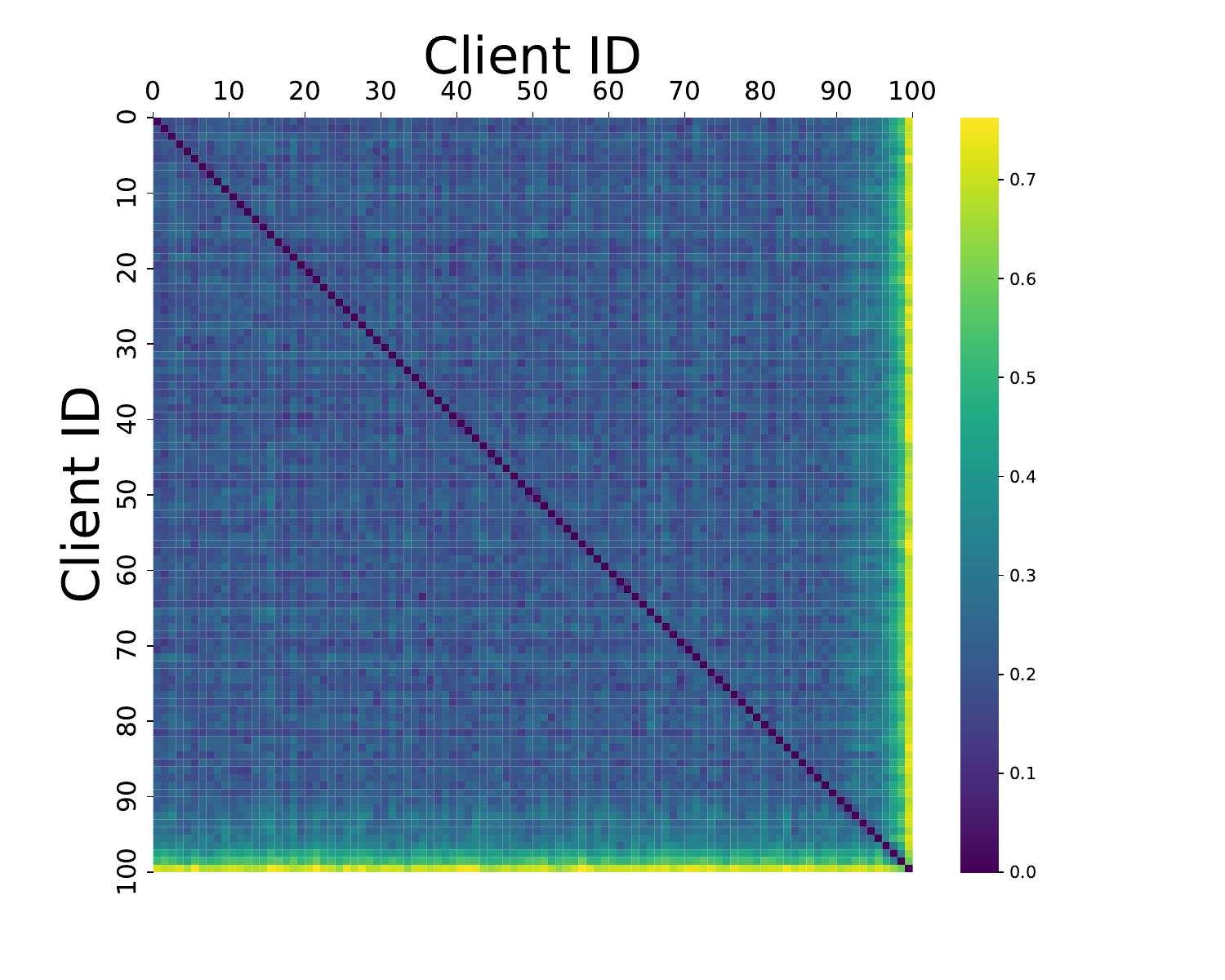}
        \caption{SemEval $\alpha$=100}
    \end{subfigure}

    \begin{subfigure}{.24\textwidth}
        \includegraphics[width=\linewidth]{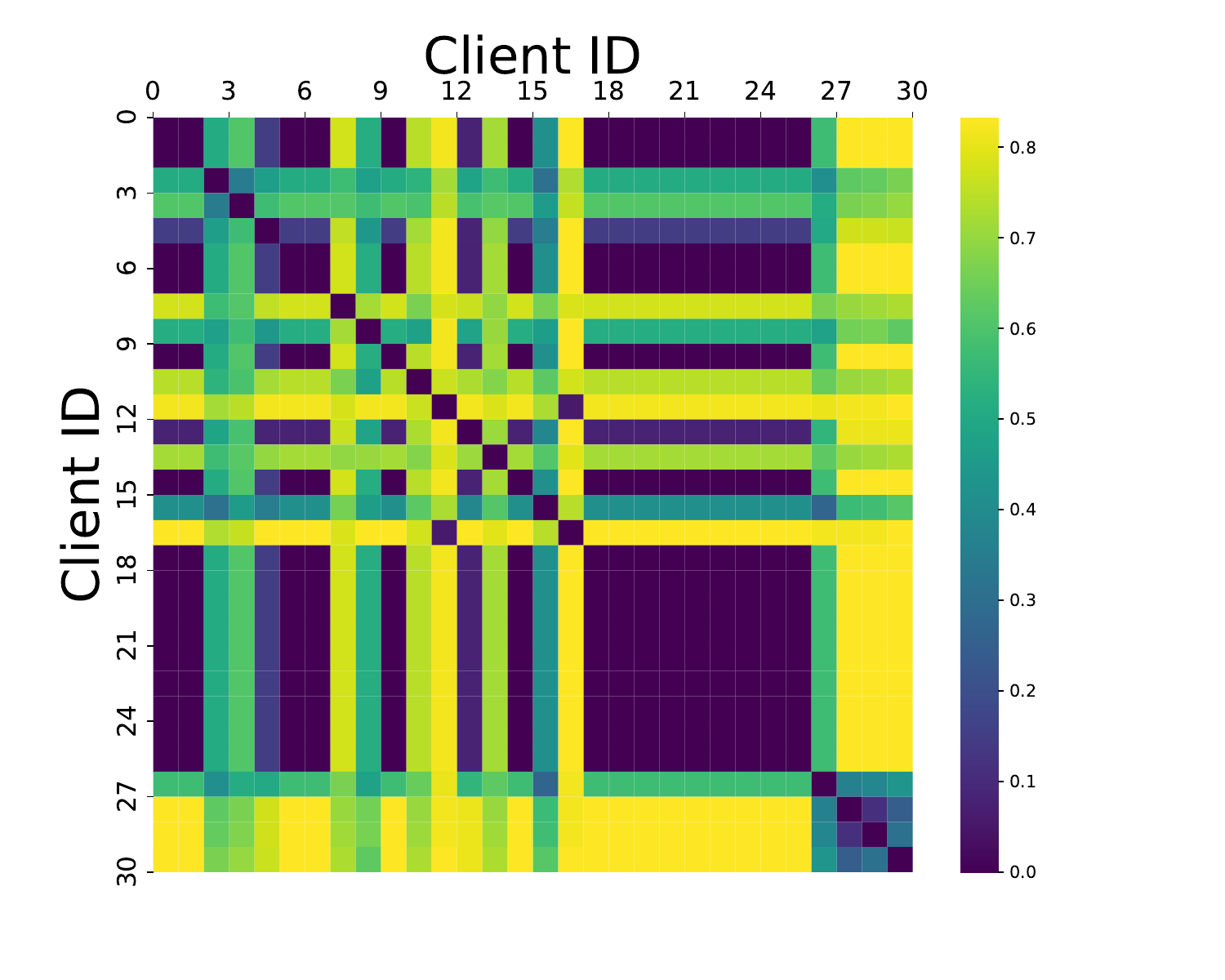}
        \caption{WNUT $\alpha$=0.1}
    \end{subfigure}\hfill
    \begin{subfigure}{.24\textwidth}
        \includegraphics[width=\linewidth]{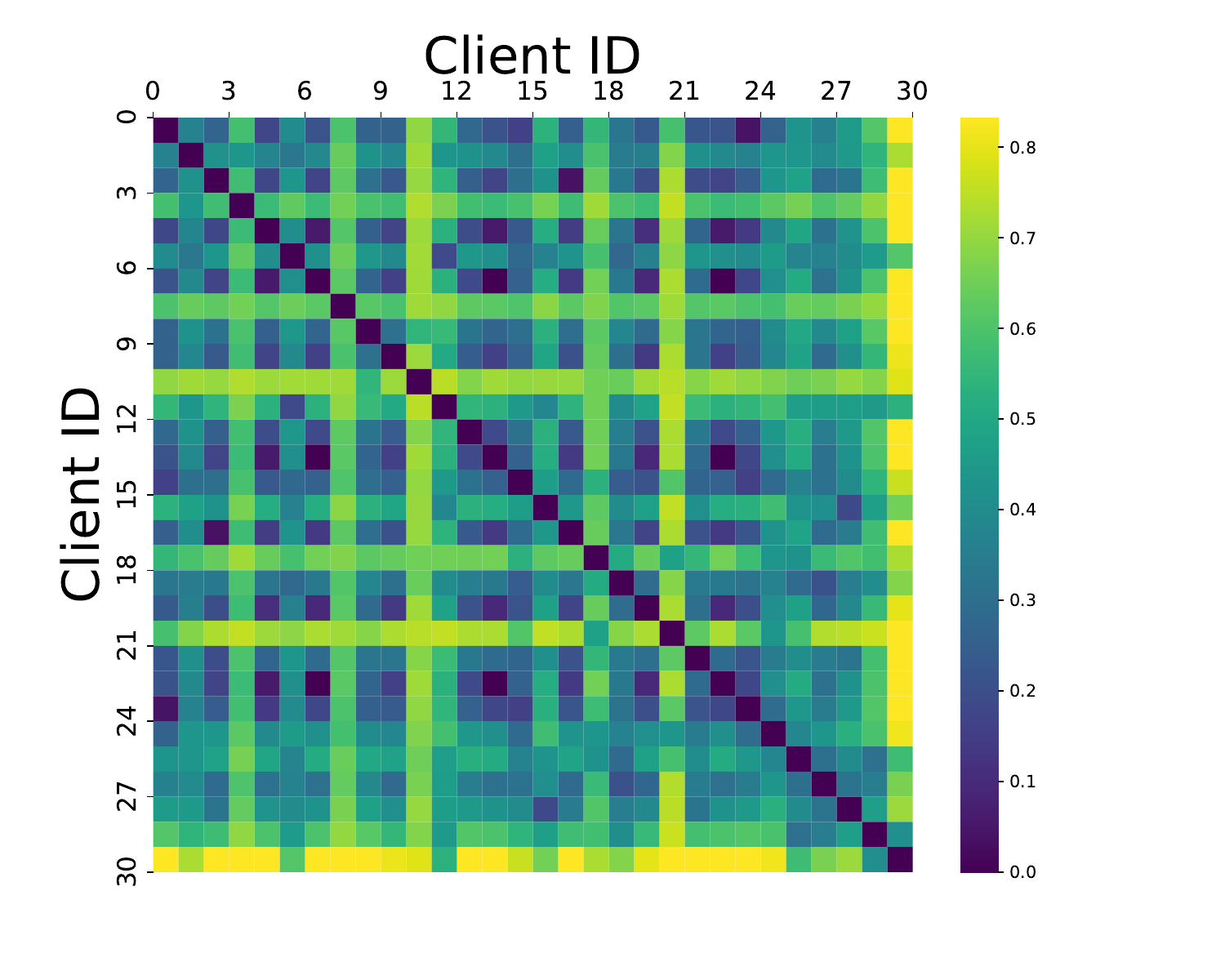}
        \caption{WNUT $\alpha$=1}
    \end{subfigure}\hfill
    \begin{subfigure}{.24\textwidth}
        \includegraphics[width=\linewidth]{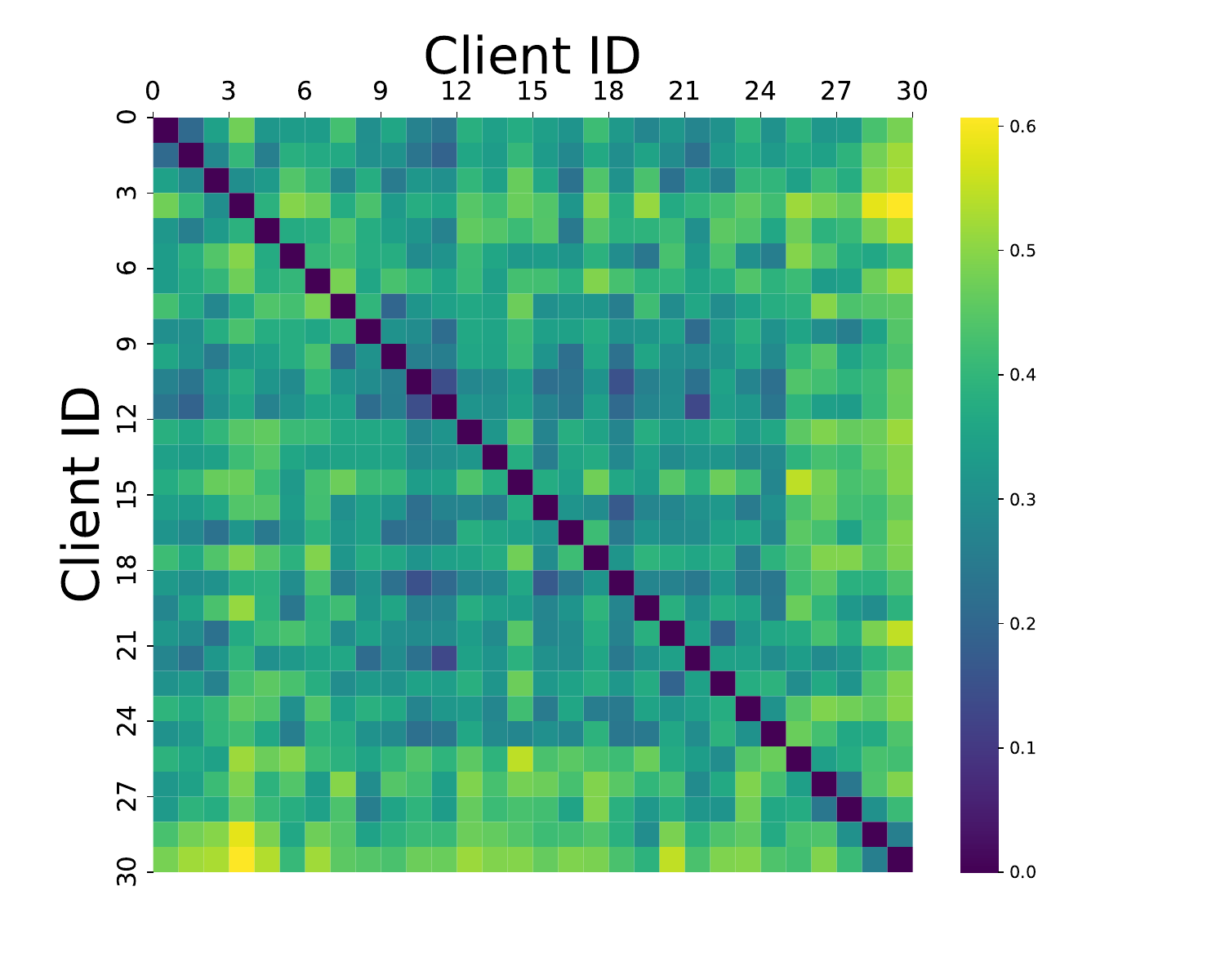}
        \caption{WNUT $\alpha$=10}
    \end{subfigure}\hfill
    \begin{subfigure}{.24\textwidth}
        \includegraphics[width=\linewidth]{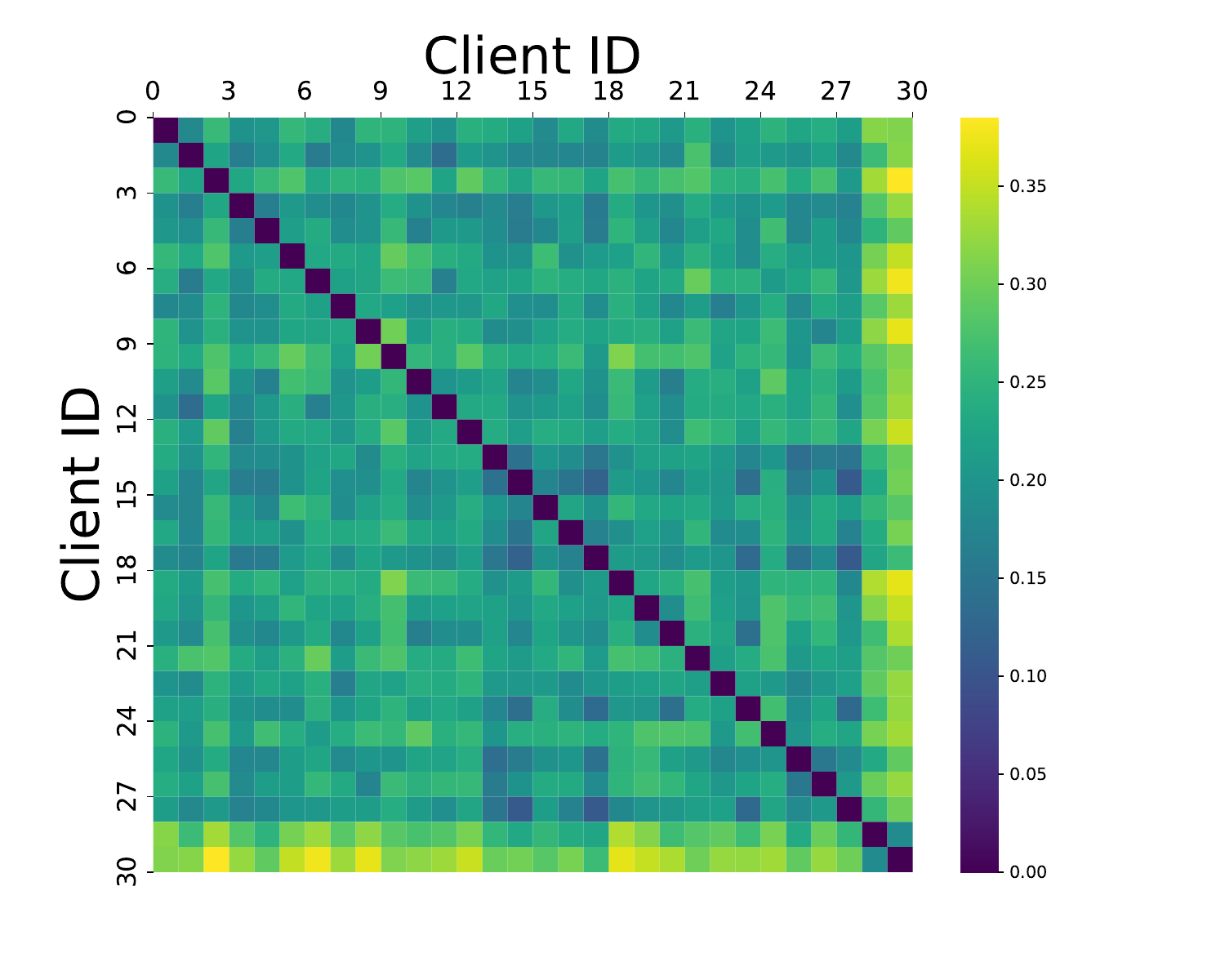}
        \caption{WNUT $\alpha$=100}
    \end{subfigure}

    \begin{subfigure}{.24\textwidth}
        \includegraphics[width=\linewidth]{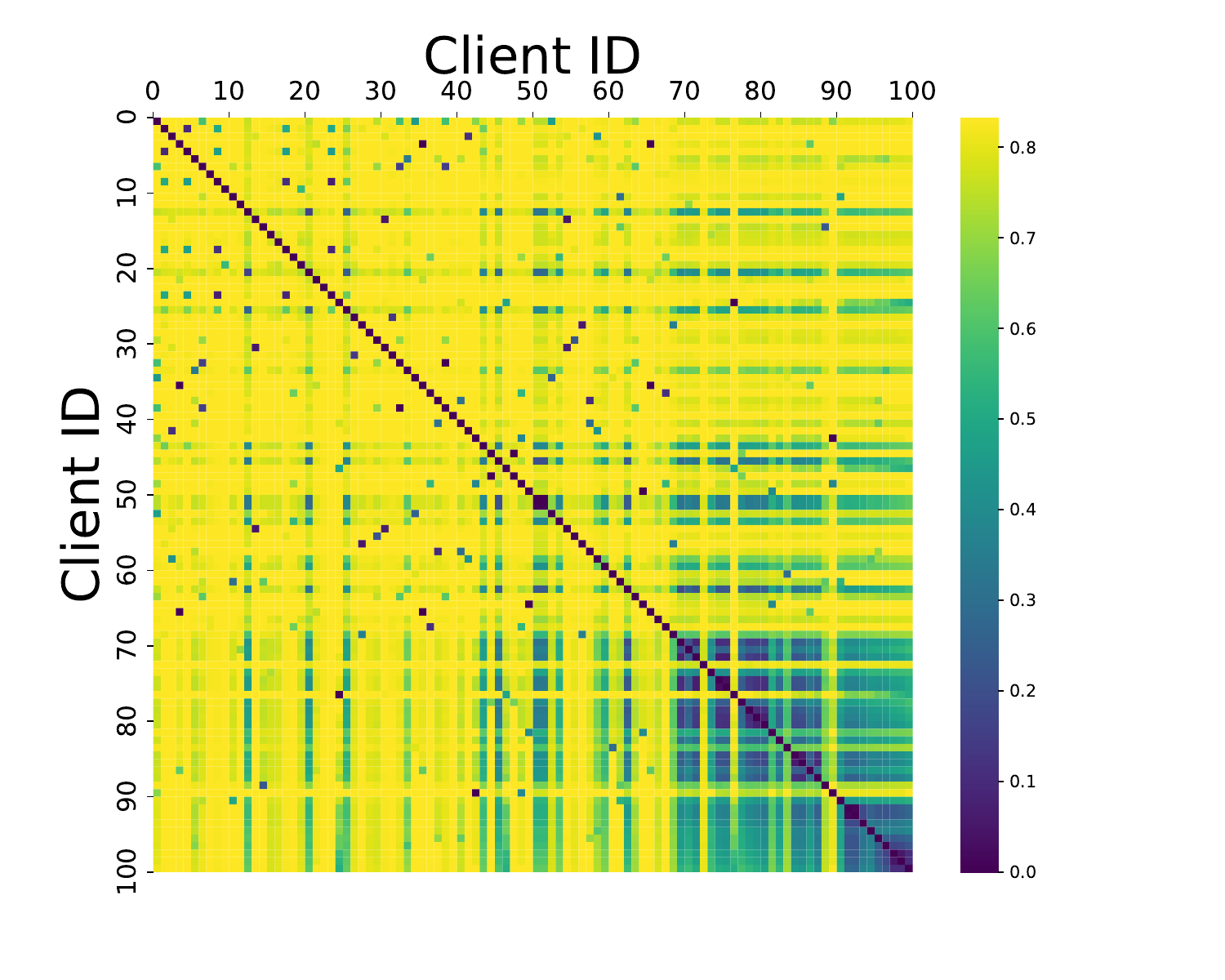}
        \caption{PLONER $\alpha$=0.1}
    \end{subfigure}\hfill
    \begin{subfigure}{.24\textwidth}
        \includegraphics[width=\linewidth]{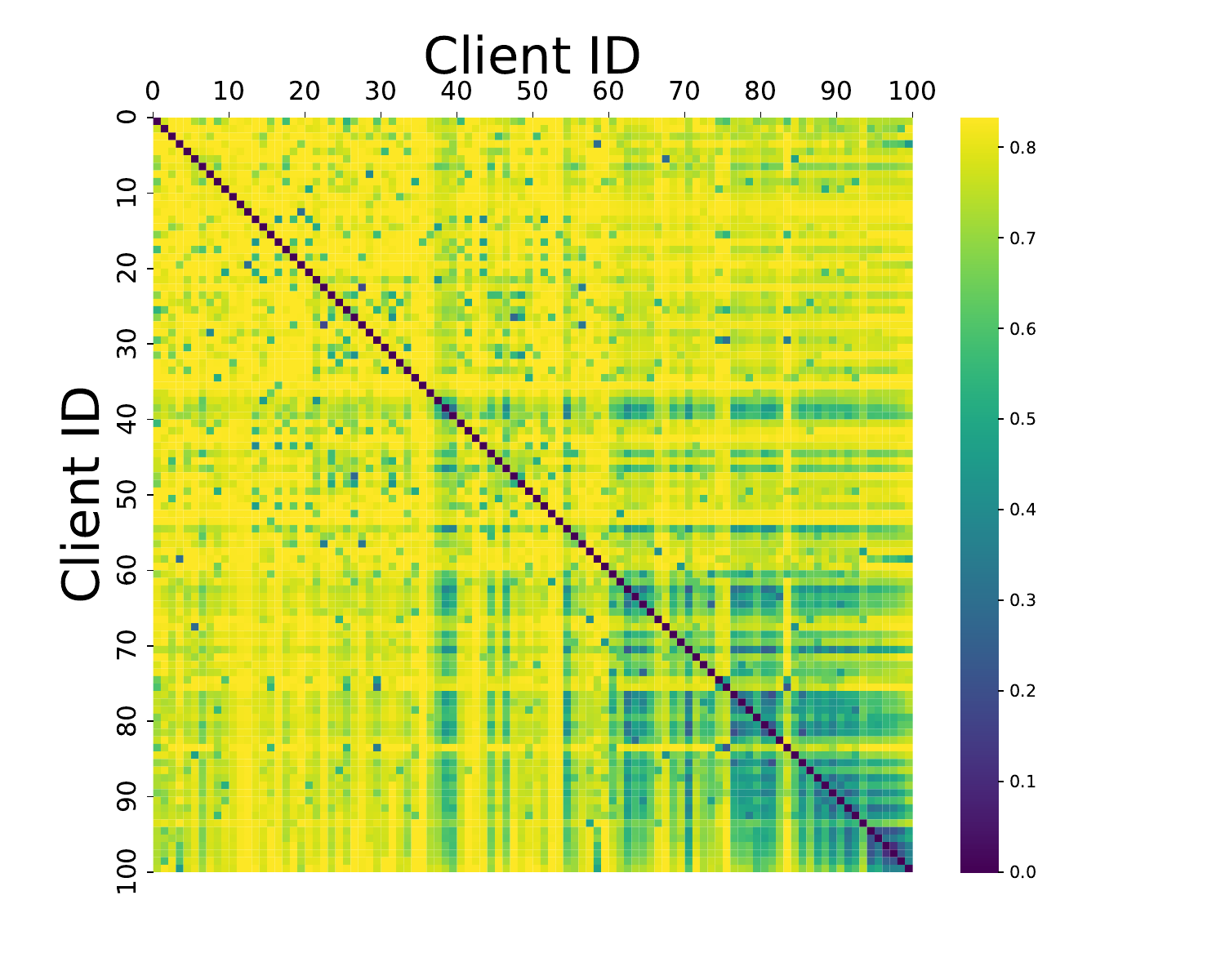}
        \caption{PLONER $\alpha$=1}
    \end{subfigure}\hfill
    \begin{subfigure}{.24\textwidth}
        \includegraphics[width=\linewidth]{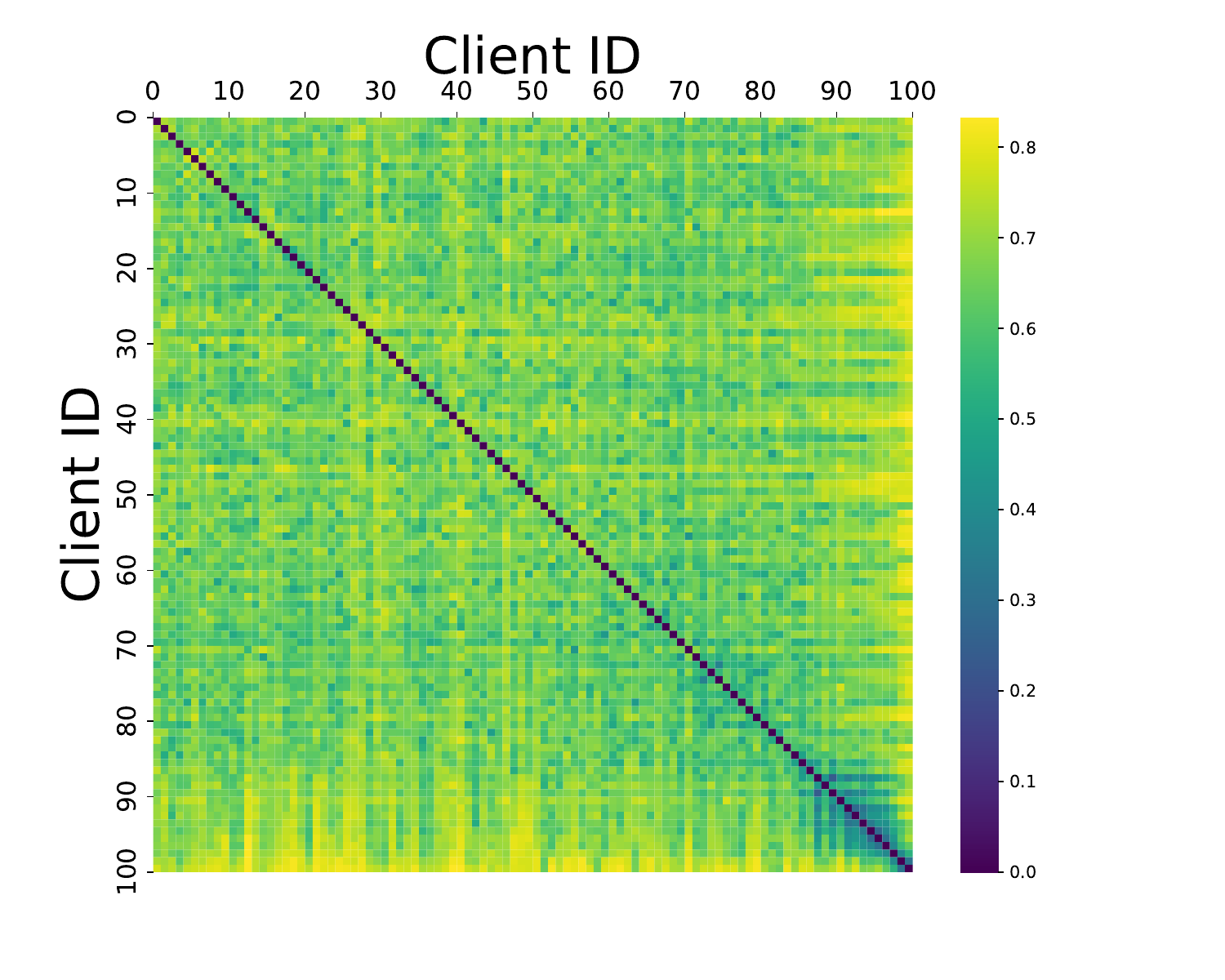}
        \caption{PLONER $\alpha$=10}
    \end{subfigure}\hfill
    \begin{subfigure}{.24\textwidth}
        \includegraphics[width=\linewidth]{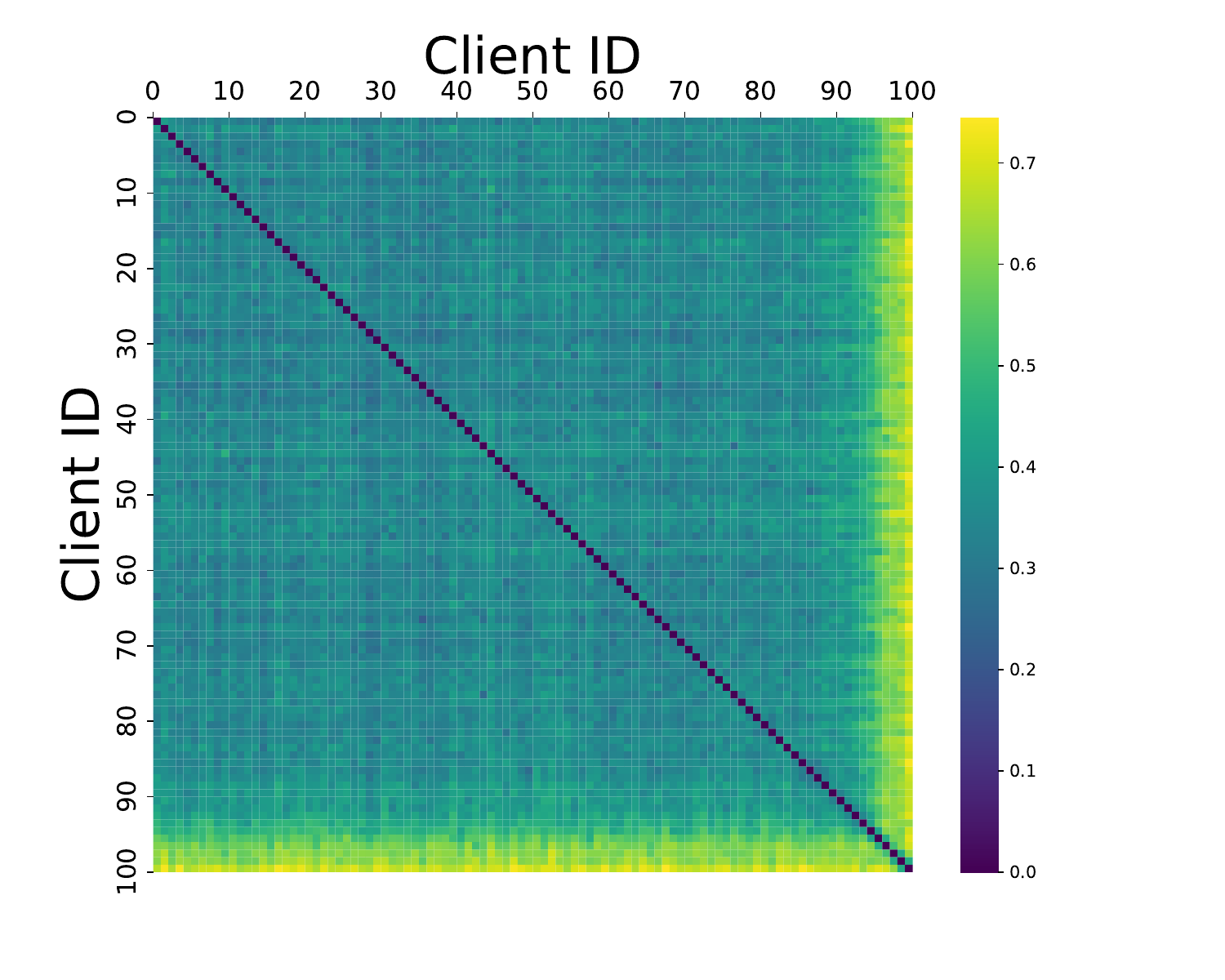}
        \caption{PLONER $\alpha$=100}
    \end{subfigure}

    \begin{subfigure}{.24\textwidth}
        \includegraphics[width=\linewidth]{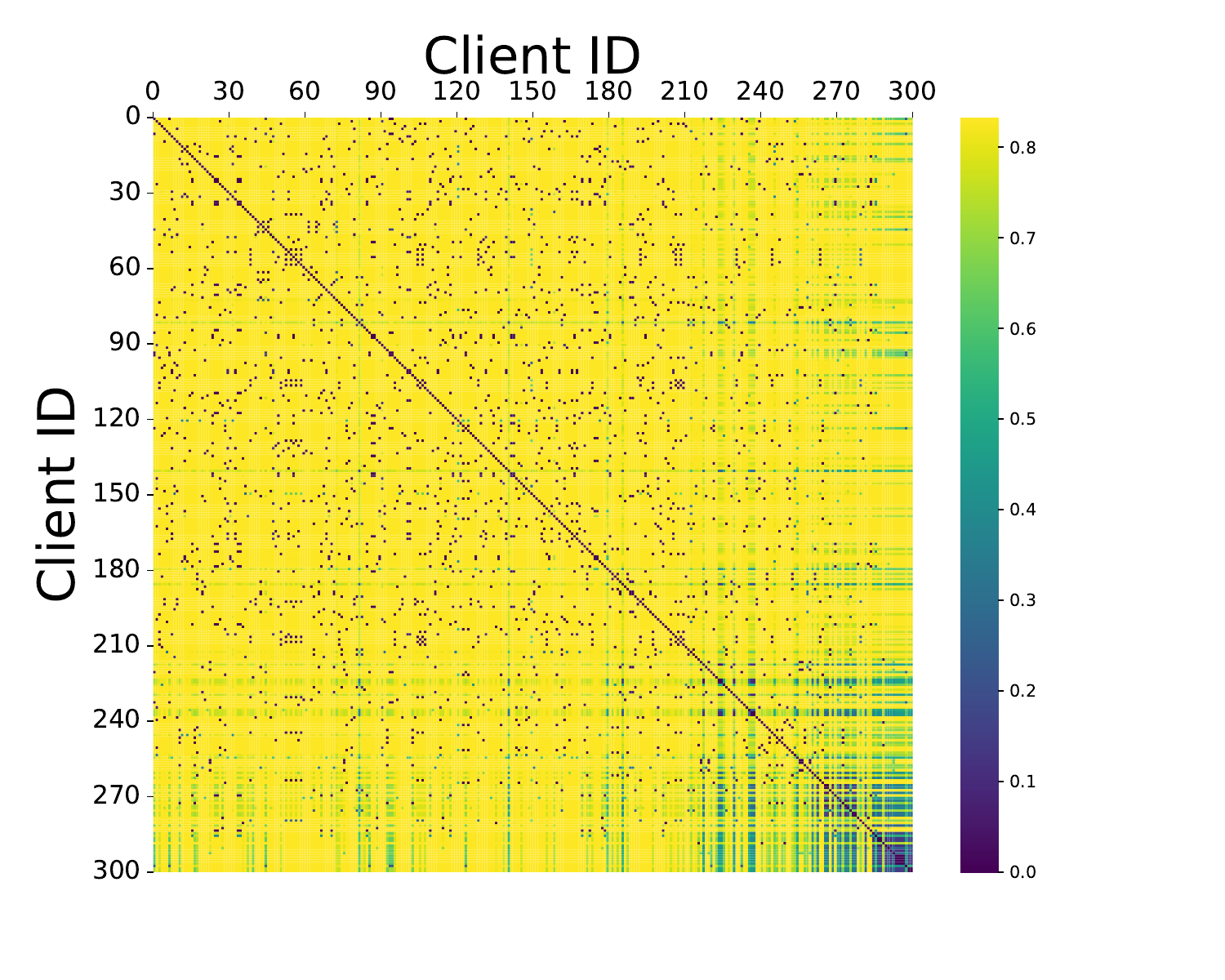}
        \caption{SQuAD $\alpha$=0.01}
    \end{subfigure}\hfill
    \begin{subfigure}{.24\textwidth}
        \includegraphics[width=\linewidth]{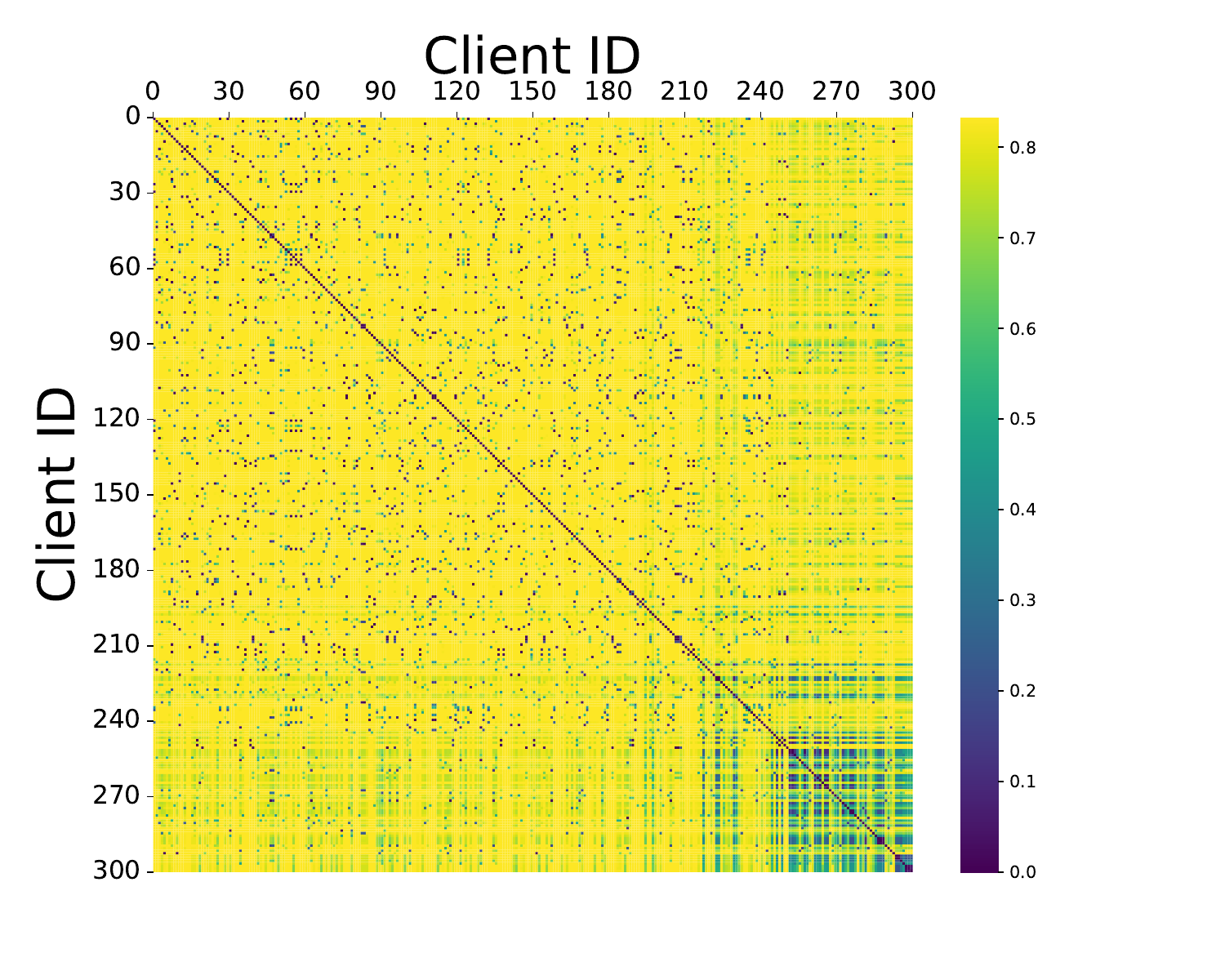}
        \caption{SQuAD $\alpha$=0.1}
    \end{subfigure}\hfill
    \begin{subfigure}{.24\textwidth}
        \includegraphics[width=\linewidth]{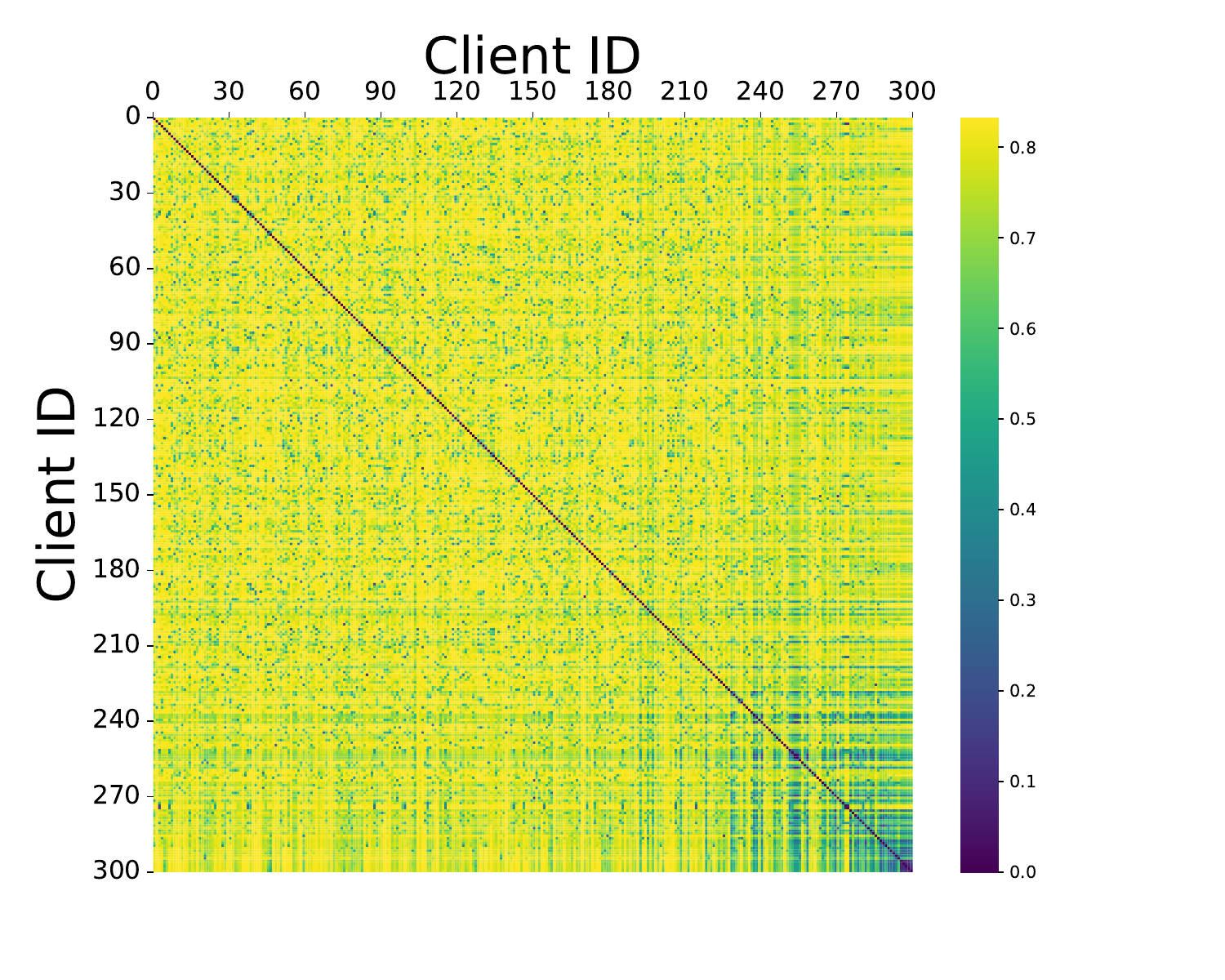}
        \caption{SQuAD $\alpha$=1}
    \end{subfigure}\hfill
    \begin{subfigure}{.24\textwidth}
        \includegraphics[width=\linewidth]{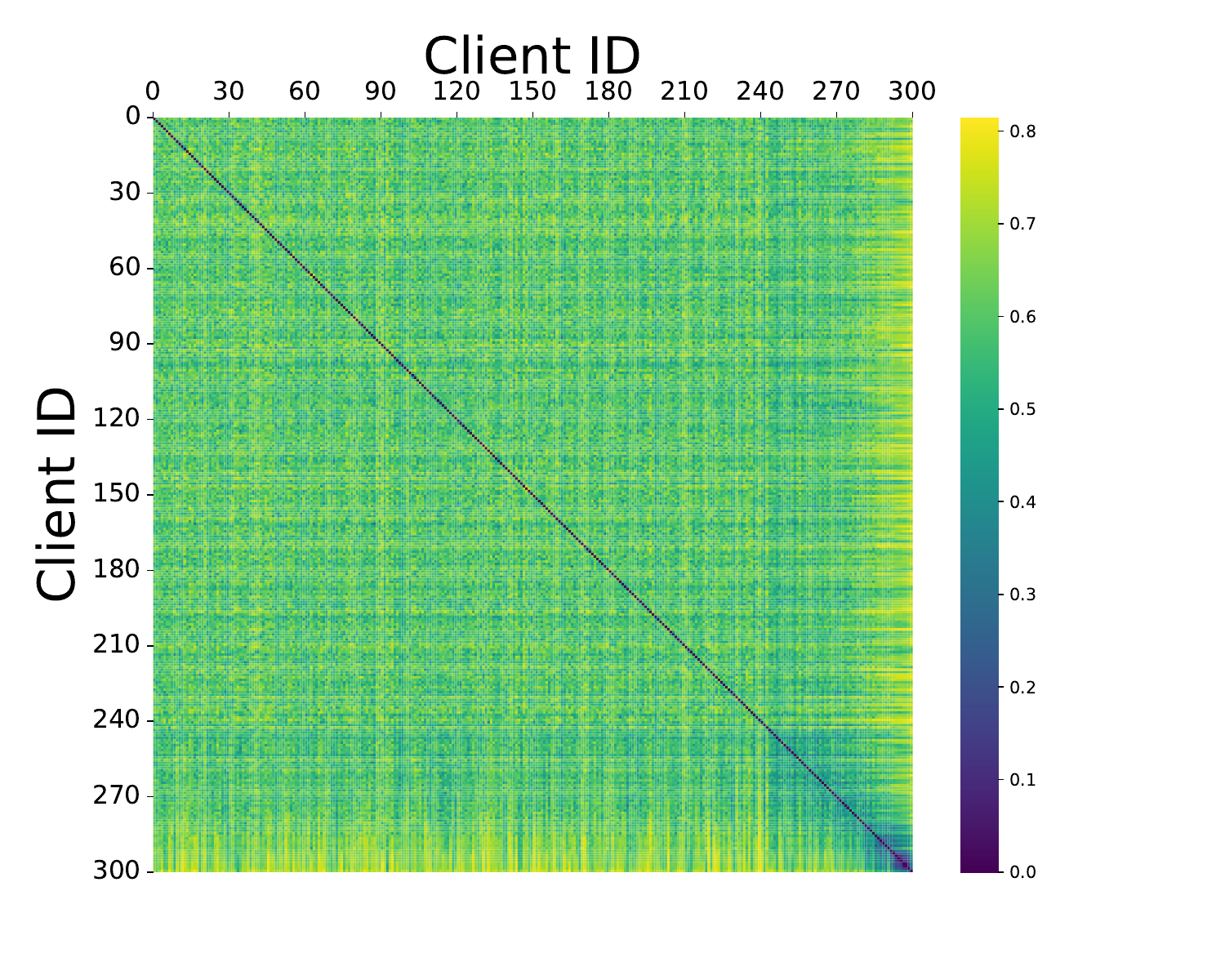}
        \caption{SQuAD $\alpha$=10}
    \end{subfigure}

    \begin{subfigure}{.24\textwidth}
        \includegraphics[width=\linewidth]{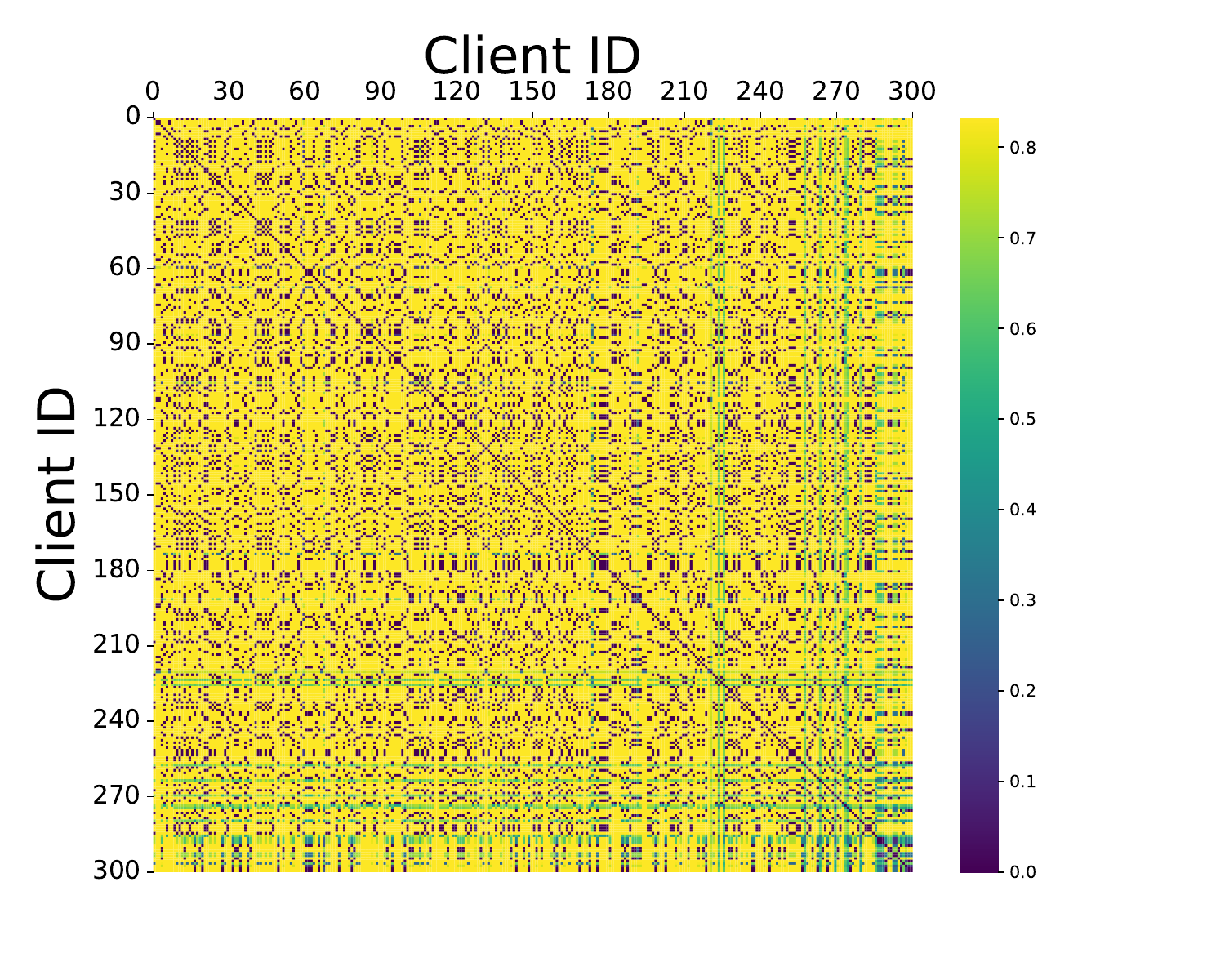}
        \caption{MRQA $\alpha$=0.01}
    \end{subfigure}\hfill
    \begin{subfigure}{.24\textwidth}
        \includegraphics[width=\linewidth]{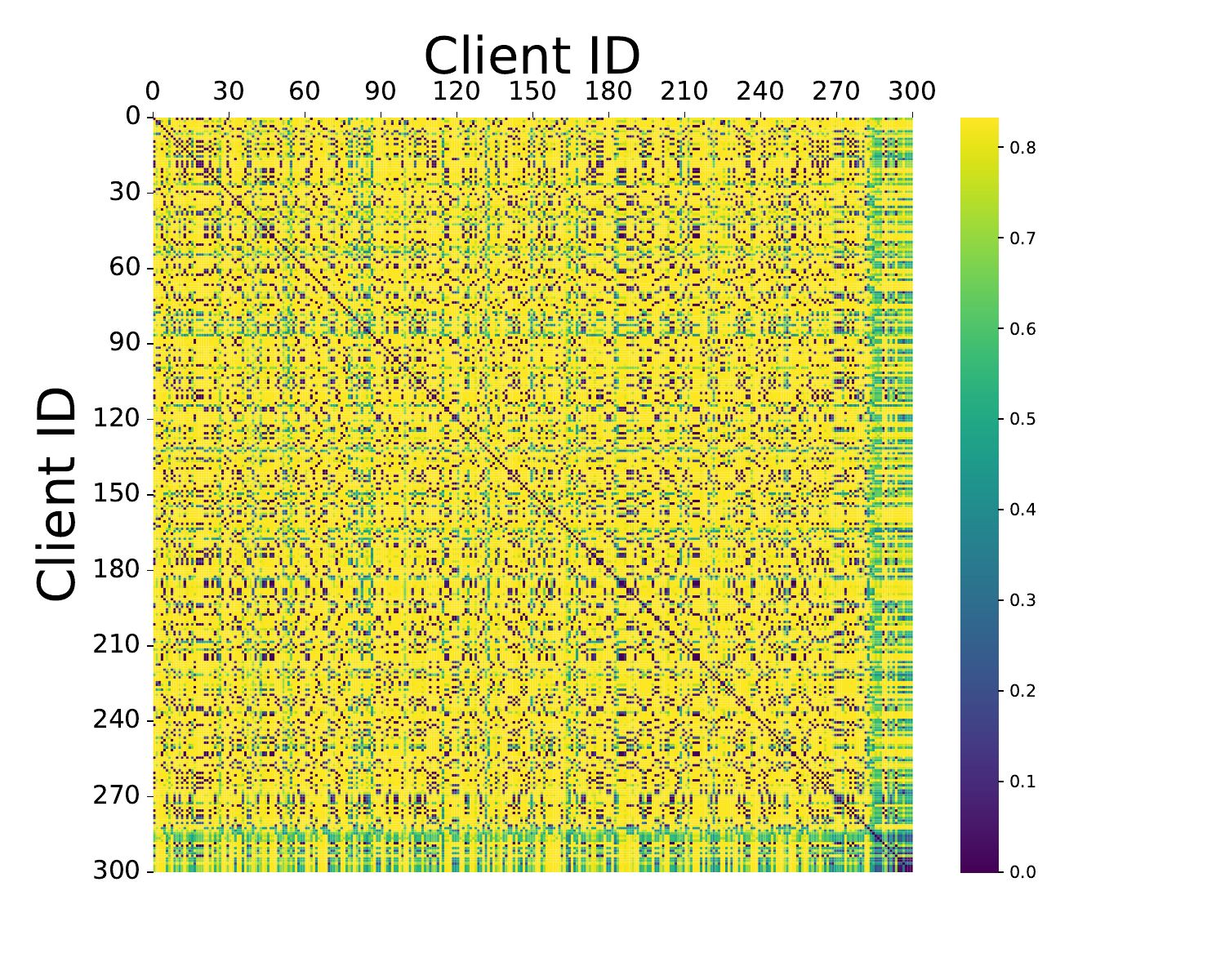}
        \caption{MRQA $\alpha$=0.1}
    \end{subfigure}\hfill
    \begin{subfigure}{.24\textwidth}
        \includegraphics[width=\linewidth]{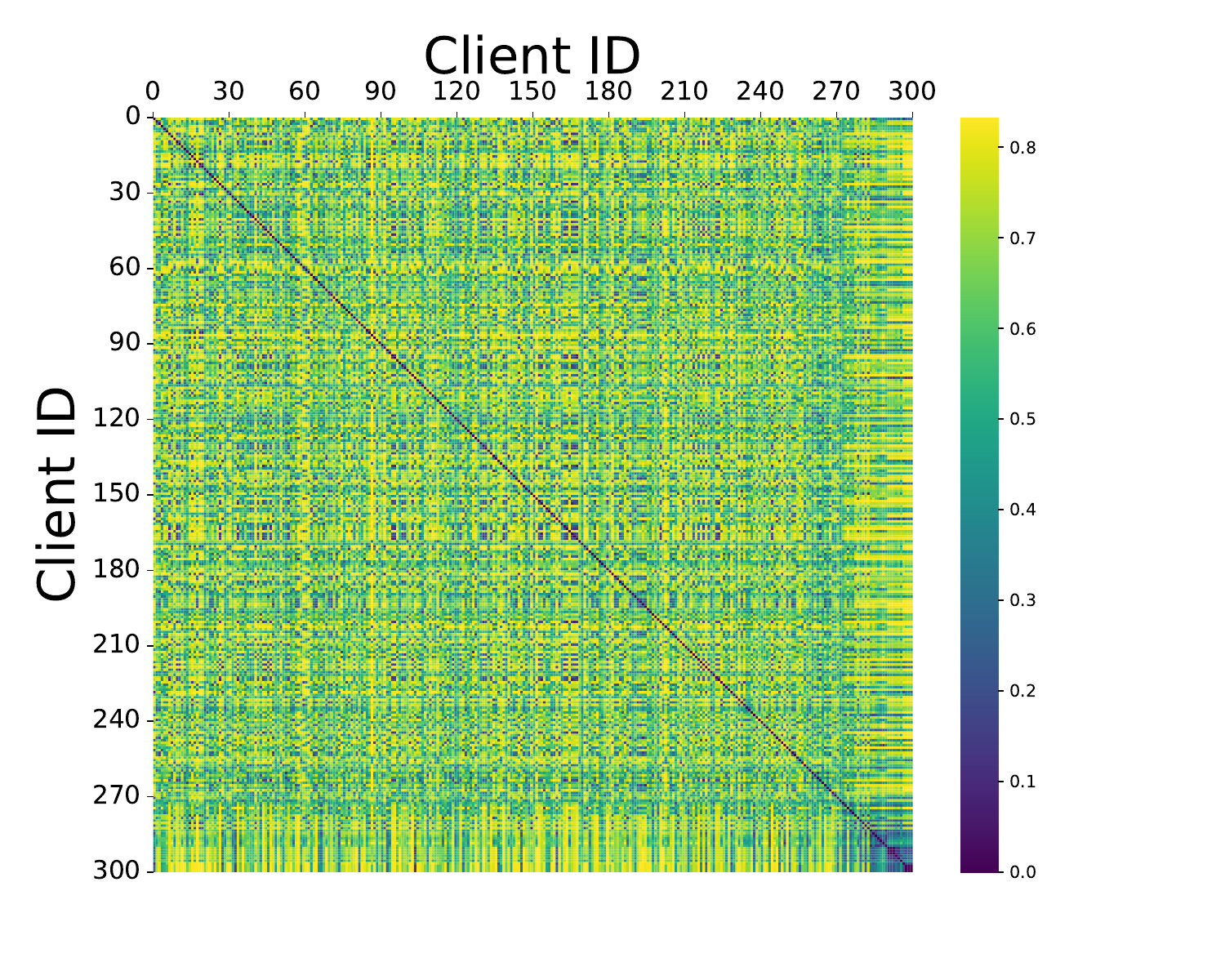}
        \caption{MRQA $\alpha$=1}
    \end{subfigure}\hfill
    \begin{subfigure}{.24\textwidth}
        \includegraphics[width=\linewidth]{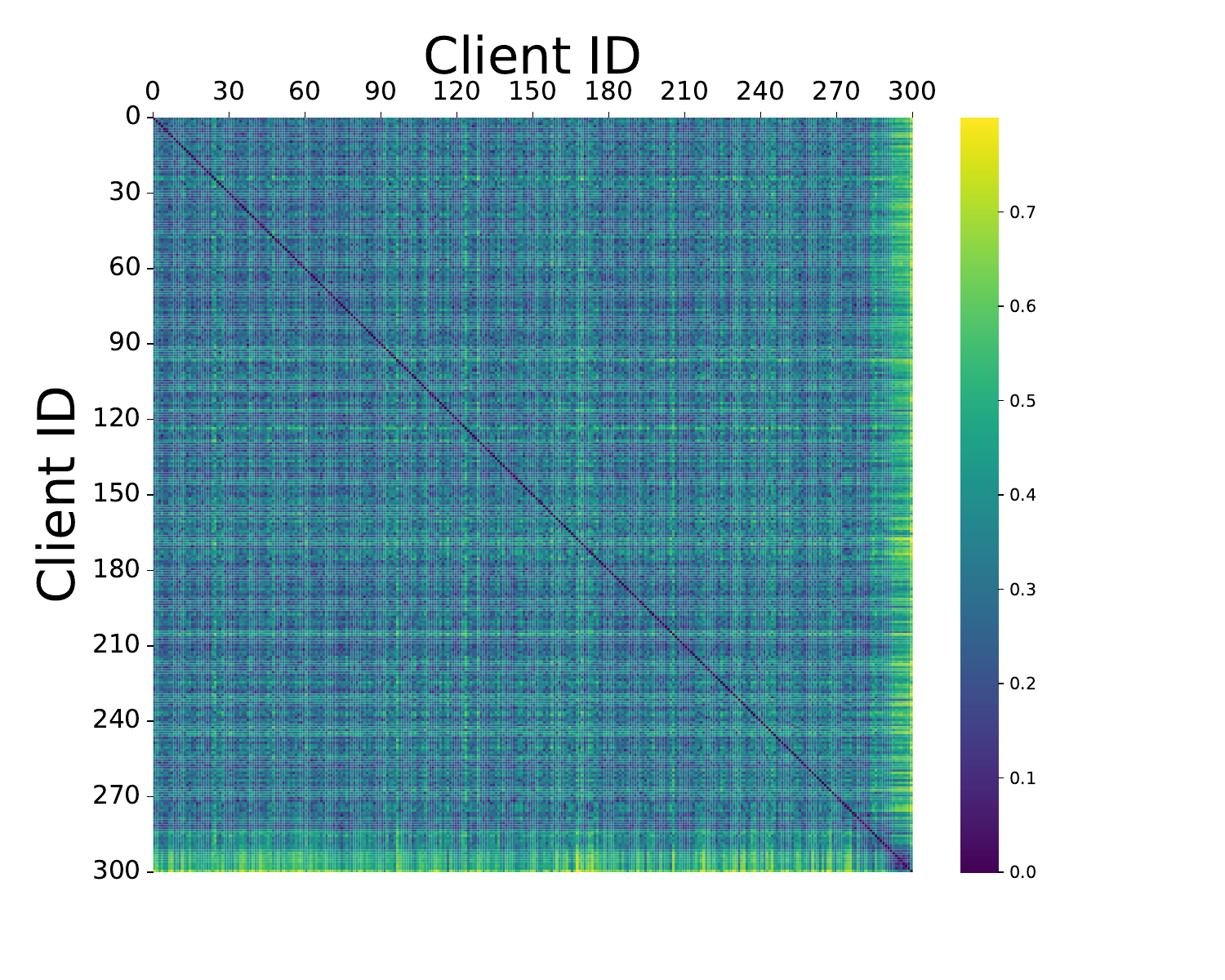}
        \caption{MRQA $\alpha$=10}
    \end{subfigure}
    
    \caption{Visualization of data heterogeneity across six datasets under various $\alpha$ settings.}
    \label{fig:visualization no-IID}
\end{figure}

We visualized the data heterogeneity among different clients partitioned according to \autoref{subsec:partition}. We employed the Jensen-Shannon divergence to characterize the disparities in the distributions of data across various clients:
\begin{equation*}
    JS(P||Q)=\frac{1}{2}\sum p_i\log\frac{p_i}{q_i} + \frac{1}{2}\sum q_i\log\frac{q_i}{p_i}
\end{equation*}
where $P$ and $Q$ denote two probability distributions defined in the same space. The value of Jensen-Shannon divergence ranges from 0 to 1, with larger values indicating a greater disparity between the two distributions. Visualization results are presented in \autoref{fig:visualization no-IID}, illustrating that the data we constructed exhibit a high degree of non-IID.

\section{Details of time cost}
\autoref{Tab:detail overheads} and \autoref{Tab:detail overhead params} presents the segmented timing based on federated learning in the experiments of \autoref{Tab:efficiency} and \autoref{fig:params}, including the local training time, communication time, and the time consumption for SVD decomposition. It can be observed from \autoref{Tab:detail overheads} that FedFT consumes more time than other methods in both local training and communication, particularly the cost of communication is considerably higher than other methods incorporating PEFT. This remains the reason why, even if FedFT converges faster and can achieve better performance, we still have to consider the potential of PEFT in federated learning. Even though FeDeRA introduces additional SVD time, this time cost is not high and only needs to be performed once during the entire training process, which is almost negligible compared to the full training time. \autoref{Tab:detail overhead params} illustrates that with the gradual increase in the amount of trainable parameters, both the client's training and communication time incrementally rise, guiding us in selecting the optimal training configuration amidst the trade-off between training communication time and convergence speed. 
\begin{table}[htbp]
\caption{The actual duration in different stages of federated learning, measured in seconds. \vspace{1mm}}
\label{Tab:detail overheads}
\resizebox{1\textwidth}{!}{
\begin{tabular}{c|c|ccccccccc} 
\toprule[2pt]
\multirow{2}{*}{\textbf{Model}}     & \multirow{2}{*}{\textbf{Dataset}}       & \multicolumn{2}{c}{\textbf{FedFT}}       & \multicolumn{2}{c}{\textbf{FedBF}}     & \multicolumn{2}{c}{\textbf{FedAP}}     & \multicolumn{2}{c}{\textbf{FedLR\&FeDeRA}} & \textbf{FeDeRA}                         \\
                                    &                                         & \textbf{Train} & \textbf{Comm.}          & \textbf{Train} & \textbf{Comm.}        & \textbf{Train} & \textbf{Comm.}        & \textbf{Train} & \textbf{Comm.}           & \multicolumn{1}{l}{\textbf{Decompose}}  \\ 
\midrule
\multirow{3}{*}{\textbf{RoBERTa}}   & \textbf{20NEWS}                         & 4.59           & \multirow{3}{*}{109.67} & 3.15           & \multirow{3}{*}{0.96} & 3.29           & \multirow{3}{*}{2.47} & 3.27           & \multirow{3}{*}{1.68}    & \multirow{3}{*}{8.62}                   \\
                                    & \textbf{WNUT}                           & 4.62           &                         & 3.31           &                       & 3.42           &                       & 3.33           &                          &                                         \\
                                    & \multicolumn{1}{l|}{\textbf{SQuADv1.1}} & 30.91          &                         & 21.83          &                       & 22.8           &                       & 22.37          &                          &                                         \\ 
\midrule
\multirow{3}{*}{\textbf{DeBERTaV3}} & \textbf{20NEWS}                         & 6.31           & \multirow{3}{*}{149.98} & 4.36           & \multirow{3}{*}{0.96} & 4.62           & \multirow{3}{*}{2.47} & 4.54           & \multirow{3}{*}{1.68}    & \multirow{3}{*}{8.62}                   \\
                                    & \textbf{WNUT}                           & 6.22           &                         & 4.49           &                       & 4.81           &                       & 4.62           &                          &                                         \\
                                    & \multicolumn{1}{l|}{\textbf{SQuADv1.1}} & 42.44          &                         & 29.51          &                       & 31.86          &                       & 30.46          &                          &                                         \\
\bottomrule[2pt]
\end{tabular}}
\end{table}

\begin{table}[!h]
\centering
\caption{The actual duration n different stages of federated learning under different trainable parameter budgets, measured in seconds. \vspace{1mm}}
\label{Tab:detail overhead params}
\resizebox{1\textwidth}{!}{
\begin{tabular}{c|cccccccccc} 
\toprule[2pt]
\multirow{3}{*}{\textbf{Method}} & \multicolumn{10}{c}{\textbf{Trainable Parameters}}                                                                                                        \\ 
\cmidrule{2-11}
                                 & \multicolumn{2}{c}{0.1M}     & \multicolumn{2}{c}{1.2M}     & \multicolumn{2}{c}{1.8M}     & \multicolumn{2}{c}{2.4M}     & \multicolumn{2}{c}{4.8M}      \\
                                 & Tain & Comm.                 & Tain & Comm.                 & Tain & Comm.                 & Tain & Comm.                 & Tain & Comm.                  \\ 
\midrule
\multicolumn{1}{l}{FedBF}        & 3.15 & \multirow{3}{*}{0.96} & -    & -                     & -    & -                     & -    & -                     & -    & -                      \\
\multicolumn{1}{l}{FedAP}        & 3.31 &                       & 3.25 & \multirow{2}{*}{1.68} & 3.29 & \multirow{2}{*}{2.47} & 4.05 & \multirow{2}{*}{3.24} & 4.17 & \multirow{2}{*}{4.55}  \\
\multicolumn{1}{l}{FedLR\&FeDeRA}  & 3.14 &                       & 3.22 &                       & 3.26 &                       & 3.95 &                       & 4.04 &                        \\
\bottomrule[2pt]
\end{tabular}}
\end{table}

\section{Additional Magnitude and direction Variation}
\begin{figure}[!htb]
    \centering
    \begin{subfigure}{.24\textwidth}
        \includegraphics[width=\linewidth]{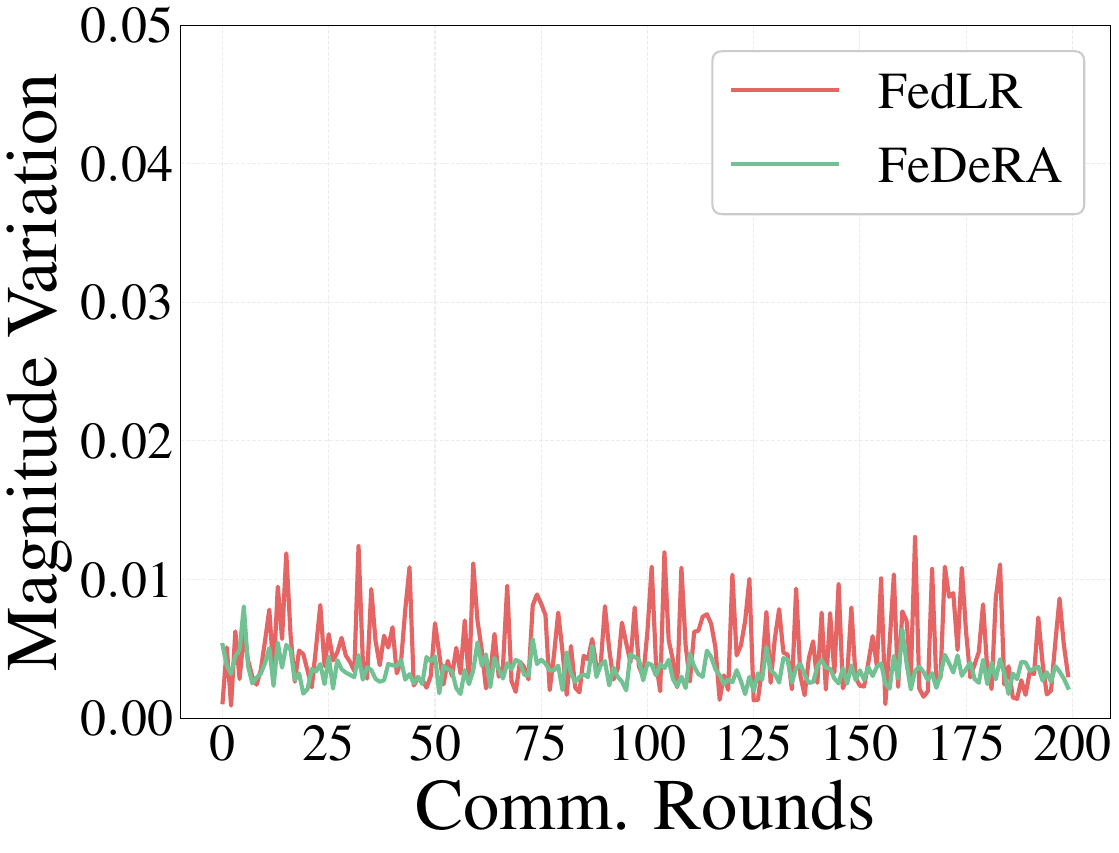}
        \caption{layer.1.q.lora\_A}
    \end{subfigure}\hfill
    \begin{subfigure}{.24\textwidth}
        \includegraphics[width=\linewidth]{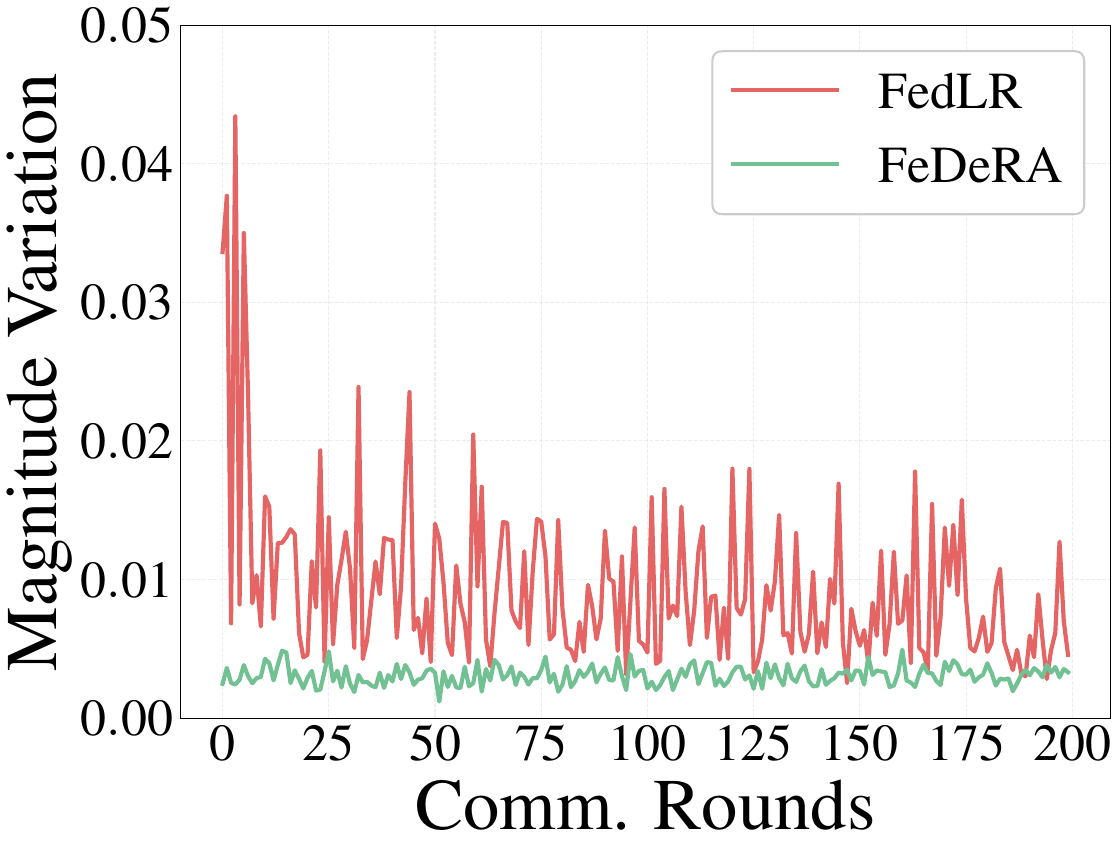}
        \caption{layer.1.q.lora\_B}
    \end{subfigure}\hfill
    \begin{subfigure}{.24\textwidth}
        \includegraphics[width=\linewidth]{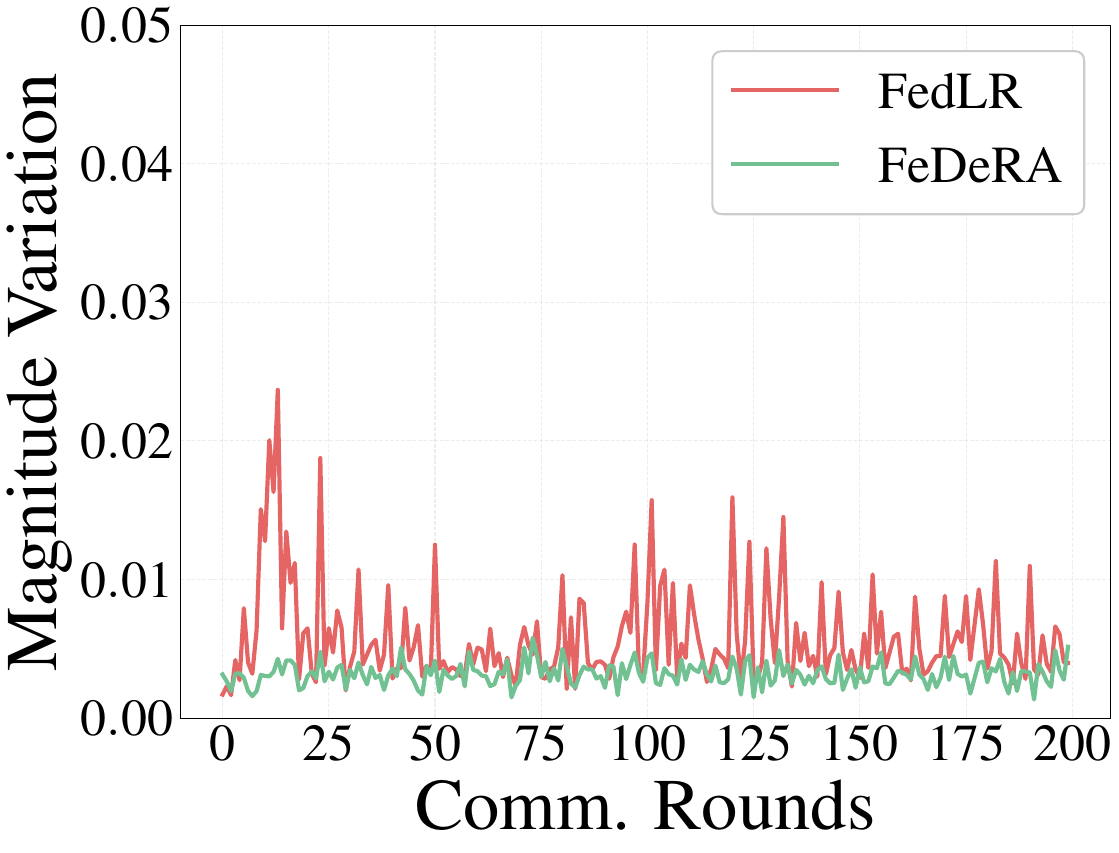}
        \caption{layer.1.q.lora\_A}
    \end{subfigure}\hfill
    \begin{subfigure}{.24\textwidth}
        \includegraphics[width=\linewidth]{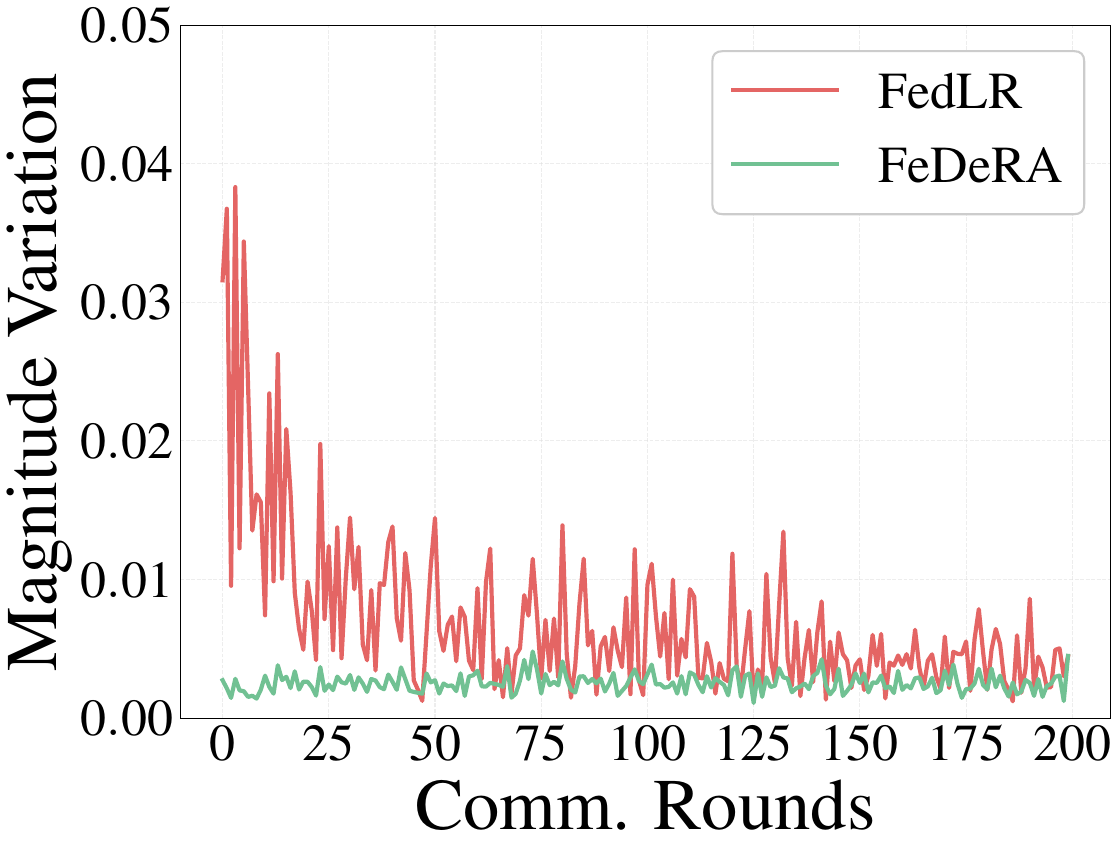}
        \caption{layer.1.v.lora\_B}
    \end{subfigure}

    \begin{subfigure}{.24\textwidth}
        \includegraphics[width=\linewidth]{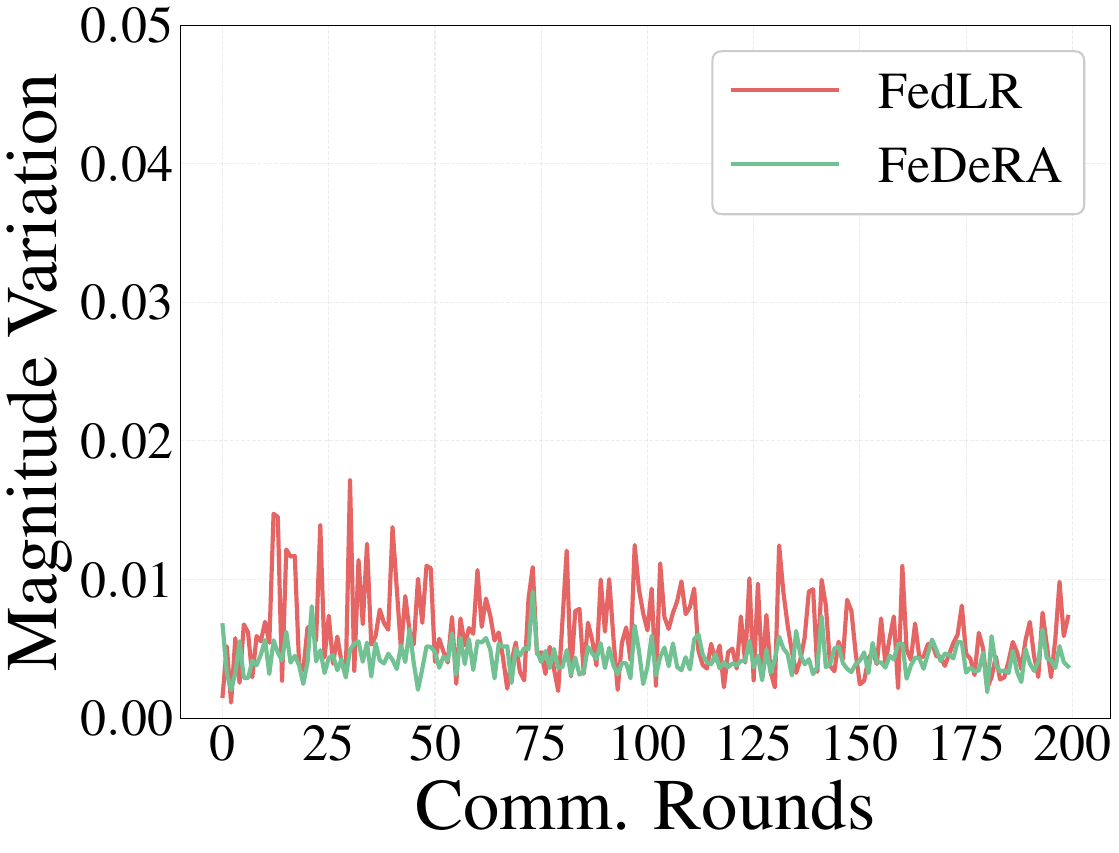}
        \caption{layer.2.q.lora\_A}
    \end{subfigure}\hfill
    \begin{subfigure}{.24\textwidth}
        \includegraphics[width=\linewidth]{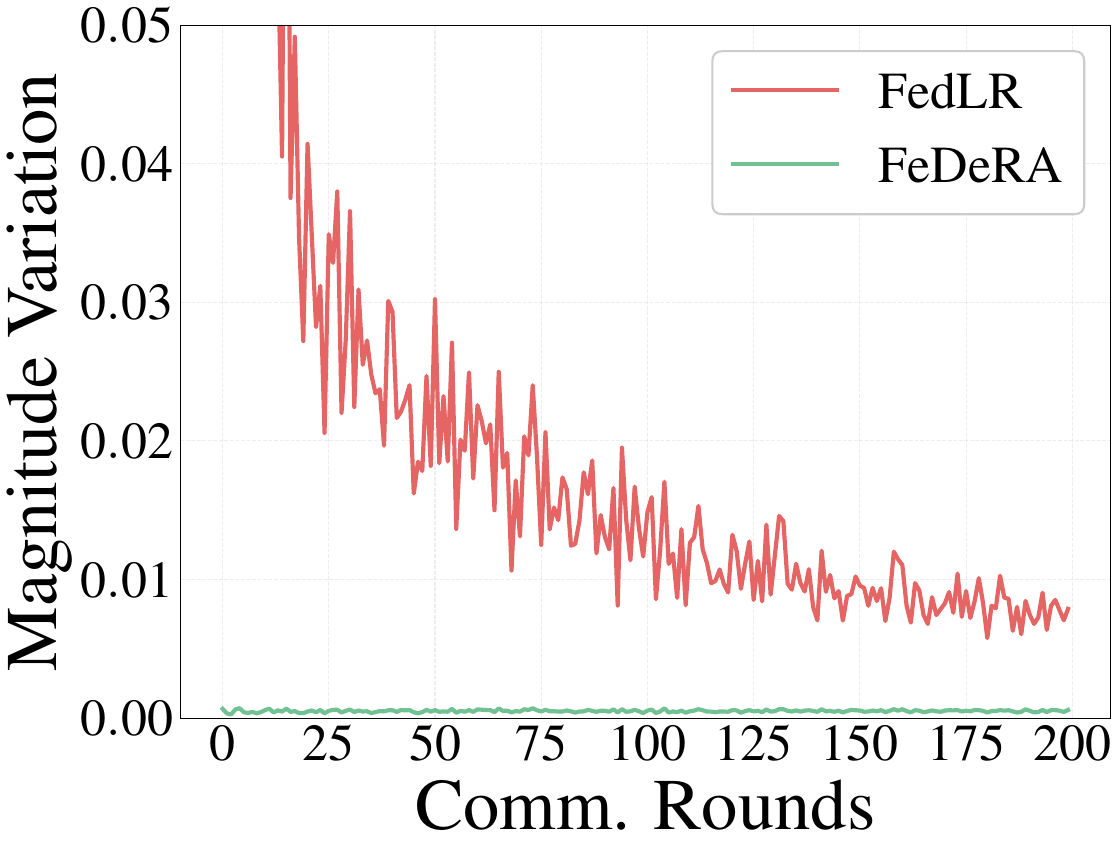}
        \caption{layer.2.q.lora\_B}
    \end{subfigure}\hfill
    \begin{subfigure}{.24\textwidth}
        \includegraphics[width=\linewidth]{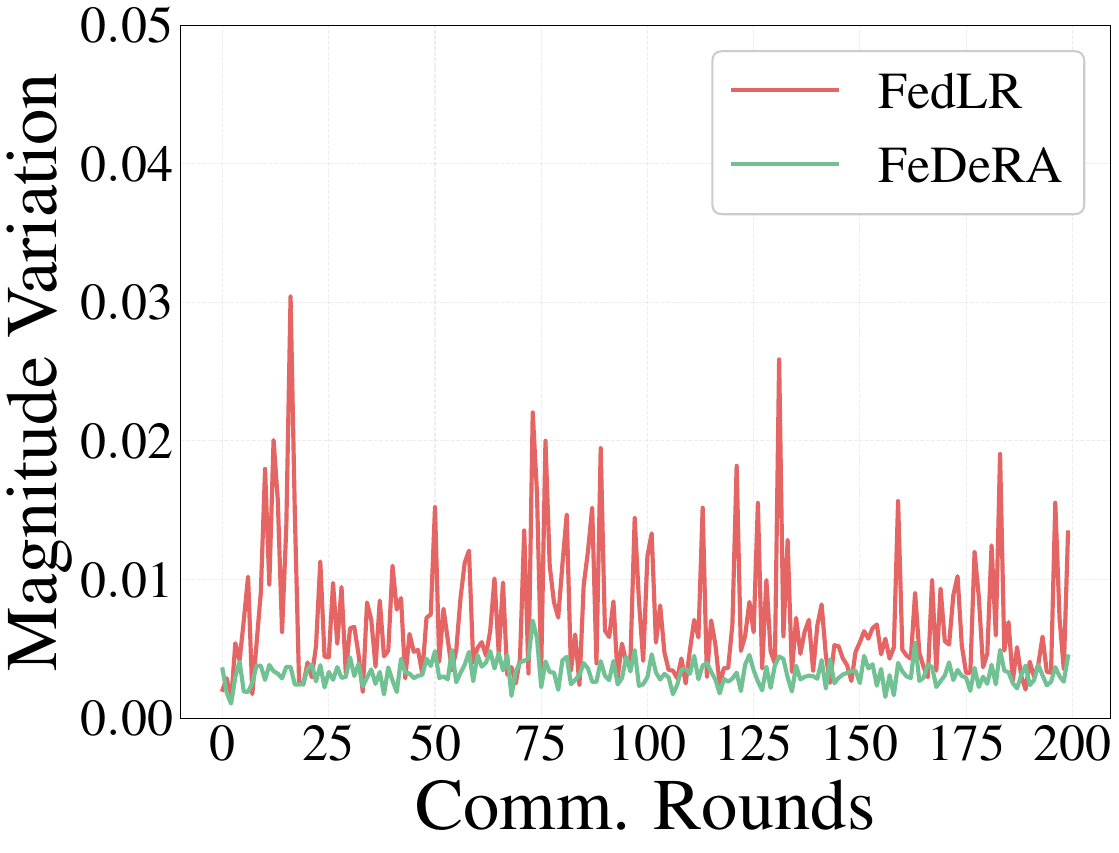}
        \caption{layer.2.q.lora\_A}
    \end{subfigure}\hfill
    \begin{subfigure}{.24\textwidth}
        \includegraphics[width=\linewidth]{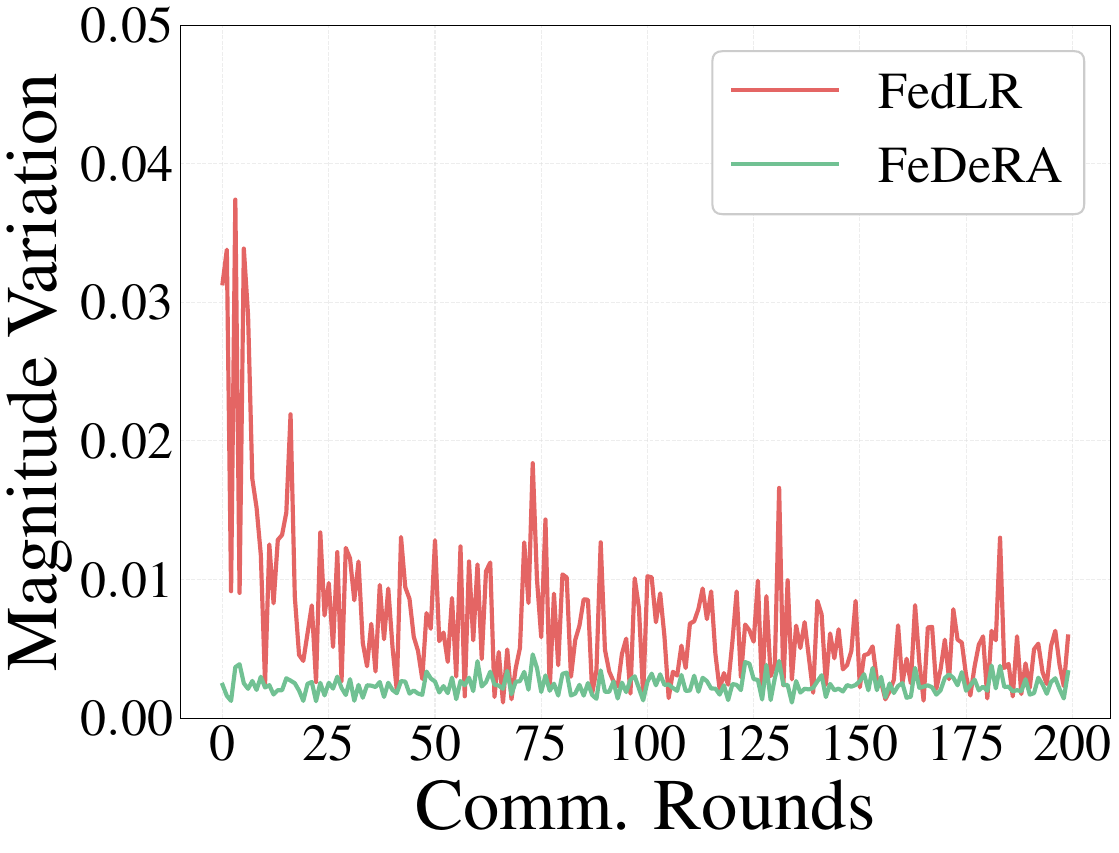}
        \caption{layer.2.v.lora\_B}
    \end{subfigure}

    \begin{subfigure}{.24\textwidth}
        \includegraphics[width=\linewidth]{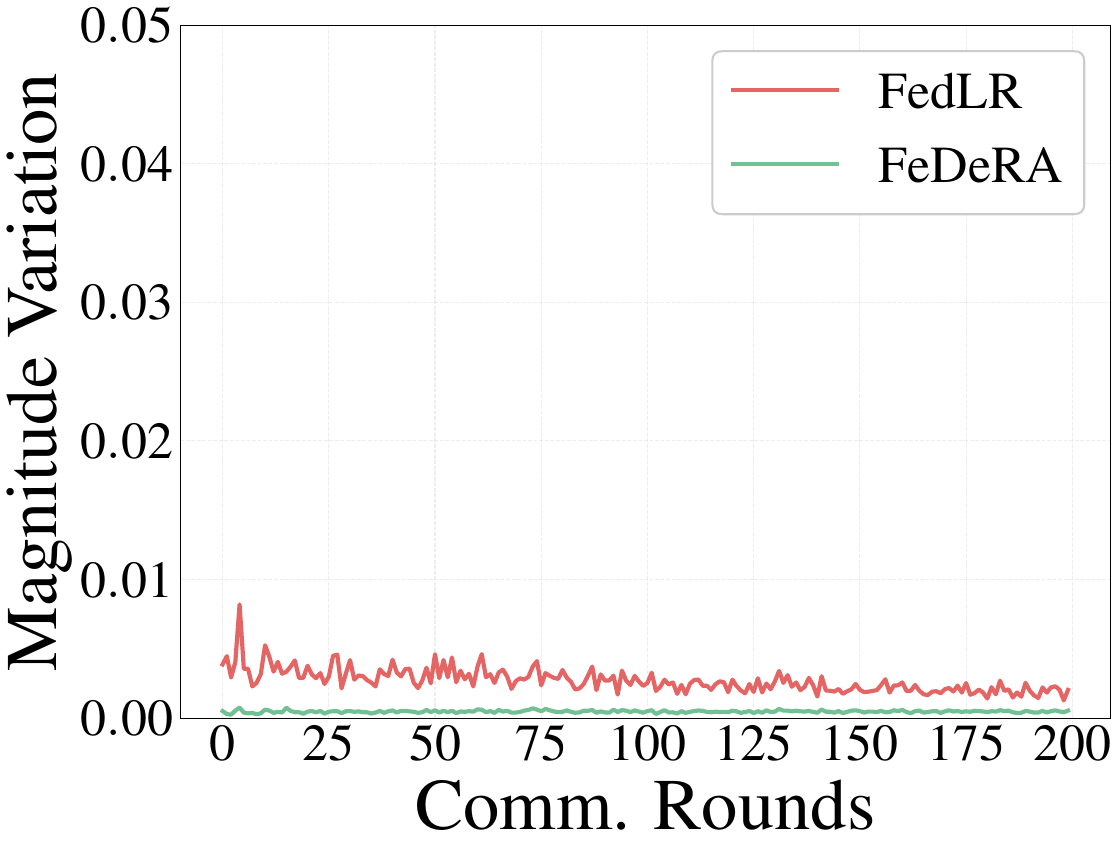}
        \caption{layer.3.q.lora\_A}
    \end{subfigure}\hfill
    \begin{subfigure}{.24\textwidth}
        \includegraphics[width=\linewidth]{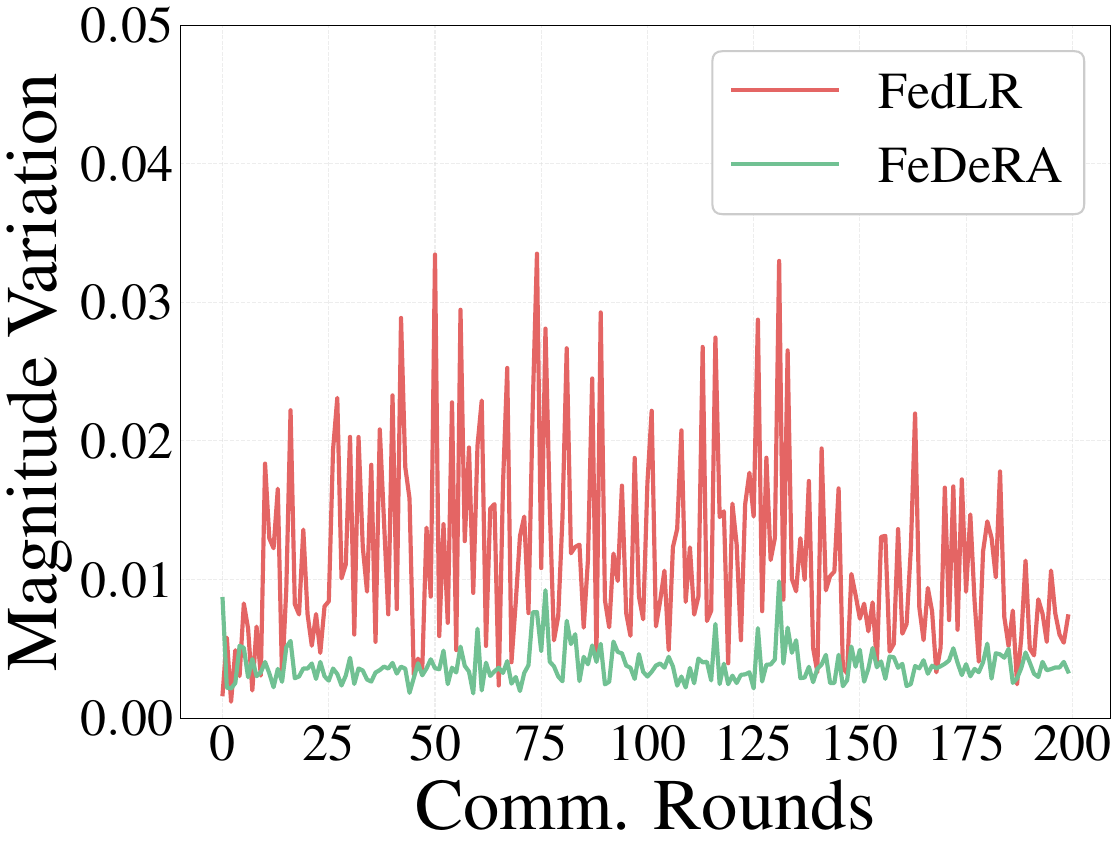}
        \caption{layer.3.q.lora\_B}
    \end{subfigure}\hfill
    \begin{subfigure}{.24\textwidth}
        \includegraphics[width=\linewidth]{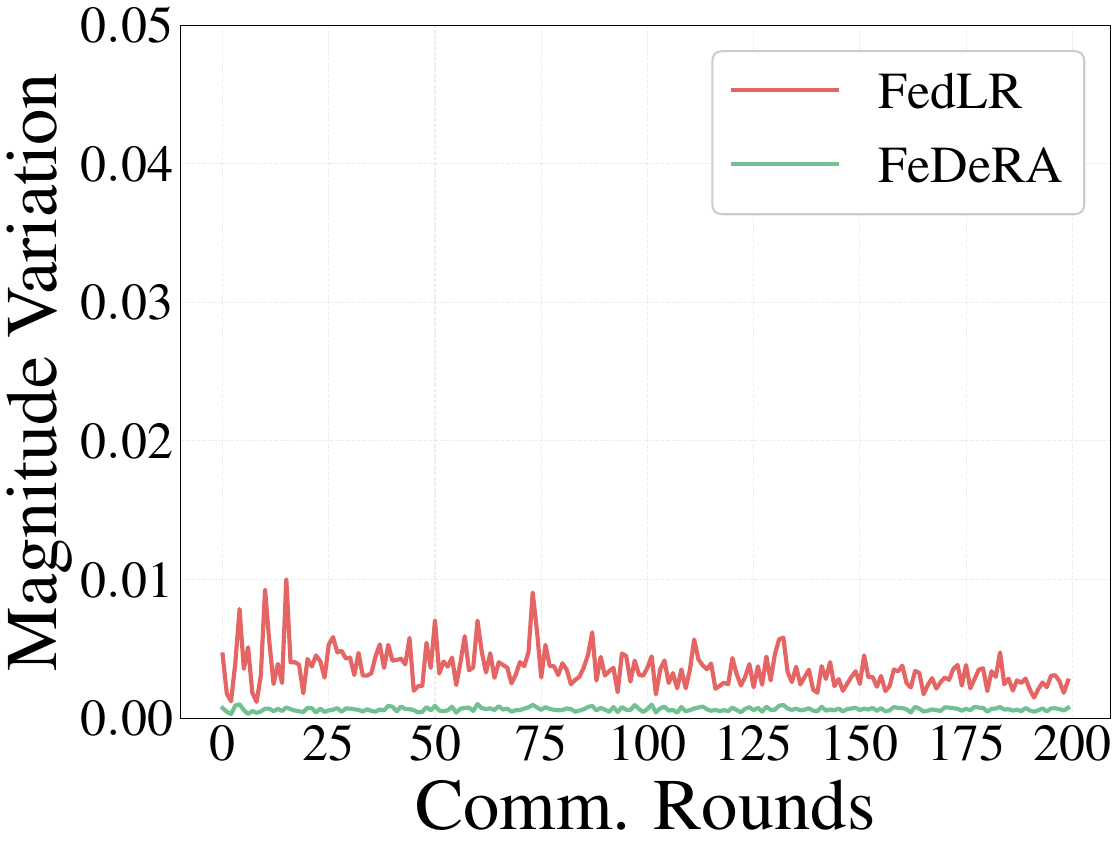}
        \caption{layer.3.q.lora\_A}
    \end{subfigure}\hfill
    \begin{subfigure}{.24\textwidth}
        \includegraphics[width=\linewidth]{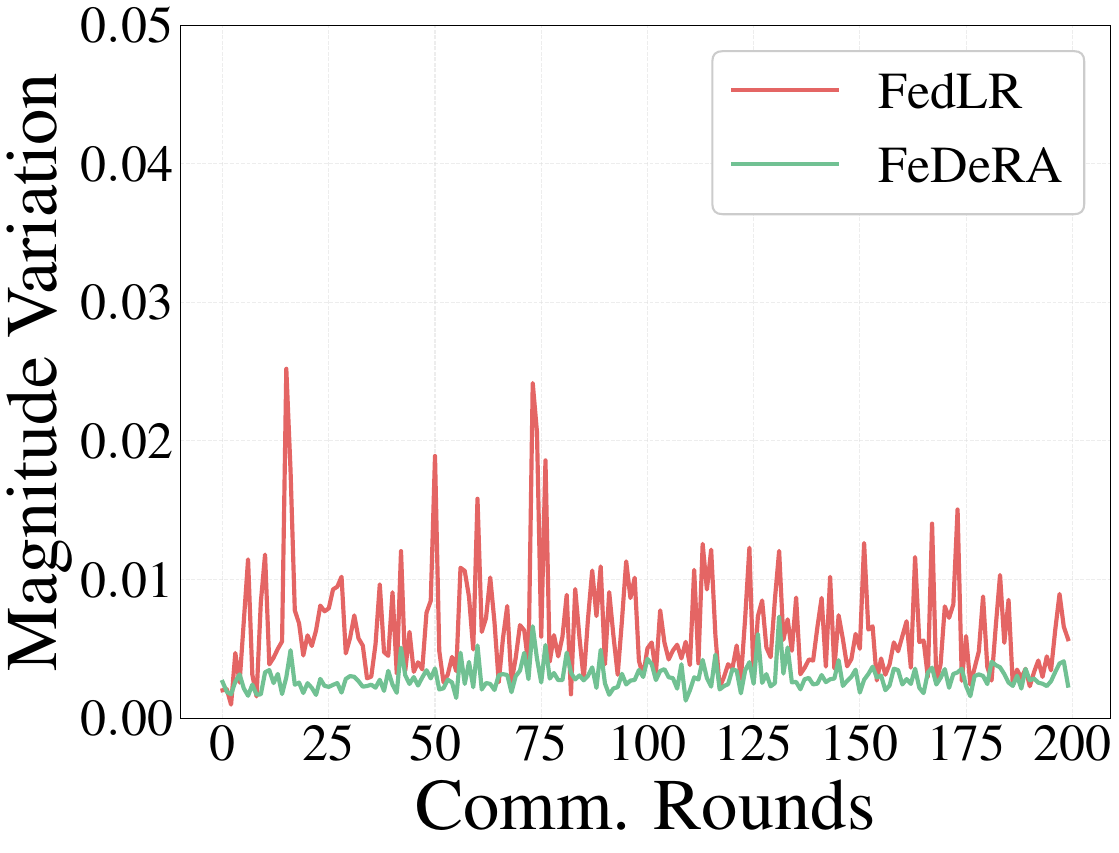}
        \caption{layer.3.v.lora\_B}
    \end{subfigure}

    \begin{subfigure}{.24\textwidth}
        \includegraphics[width=\linewidth]{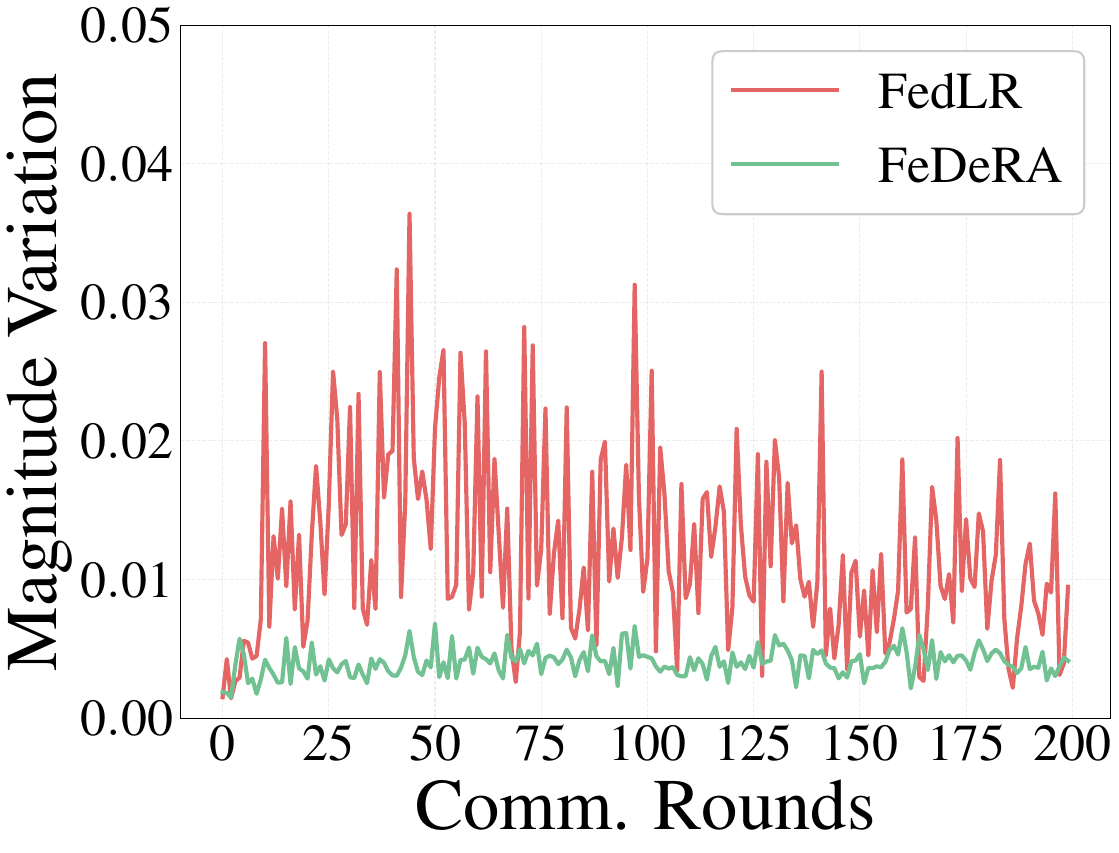}
        \caption{layer.4.q.lora\_A}
    \end{subfigure}\hfill
    \begin{subfigure}{.24\textwidth}
        \includegraphics[width=\linewidth]{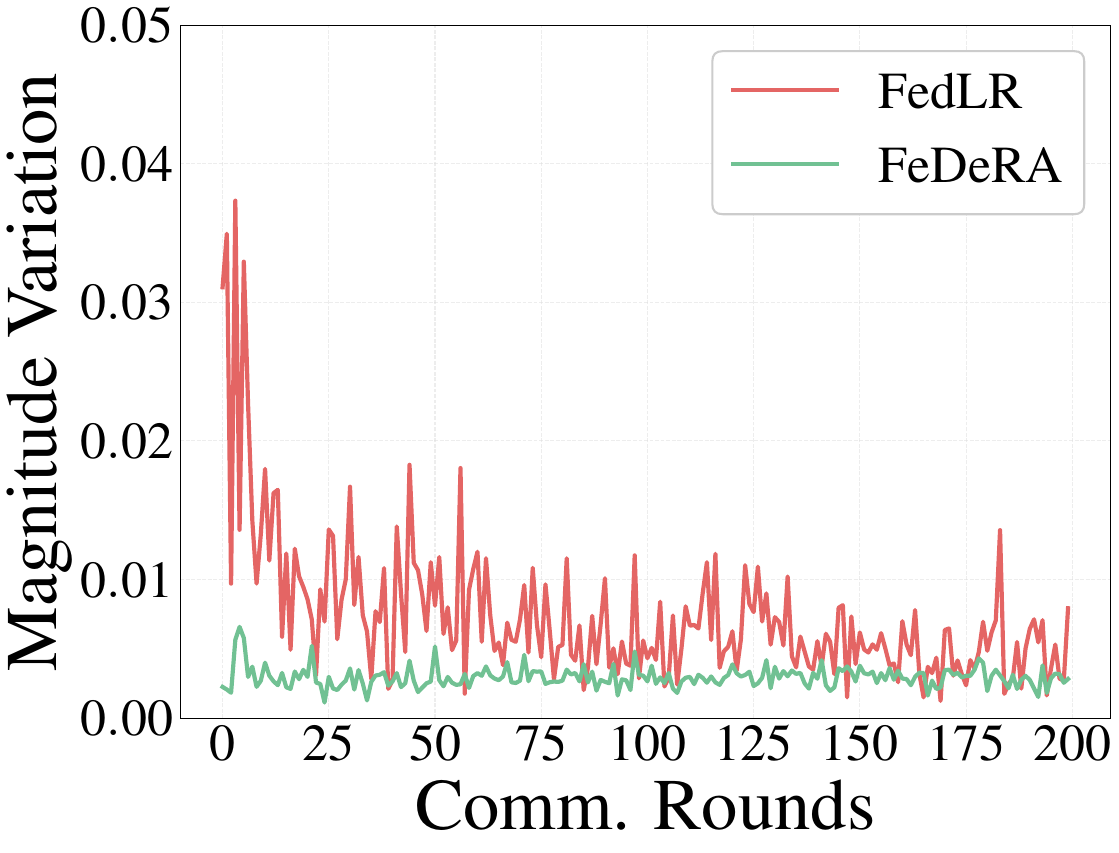}
        \caption{layer.4.q.lora\_B}
    \end{subfigure}\hfill
    \begin{subfigure}{.24\textwidth}
        \includegraphics[width=\linewidth]{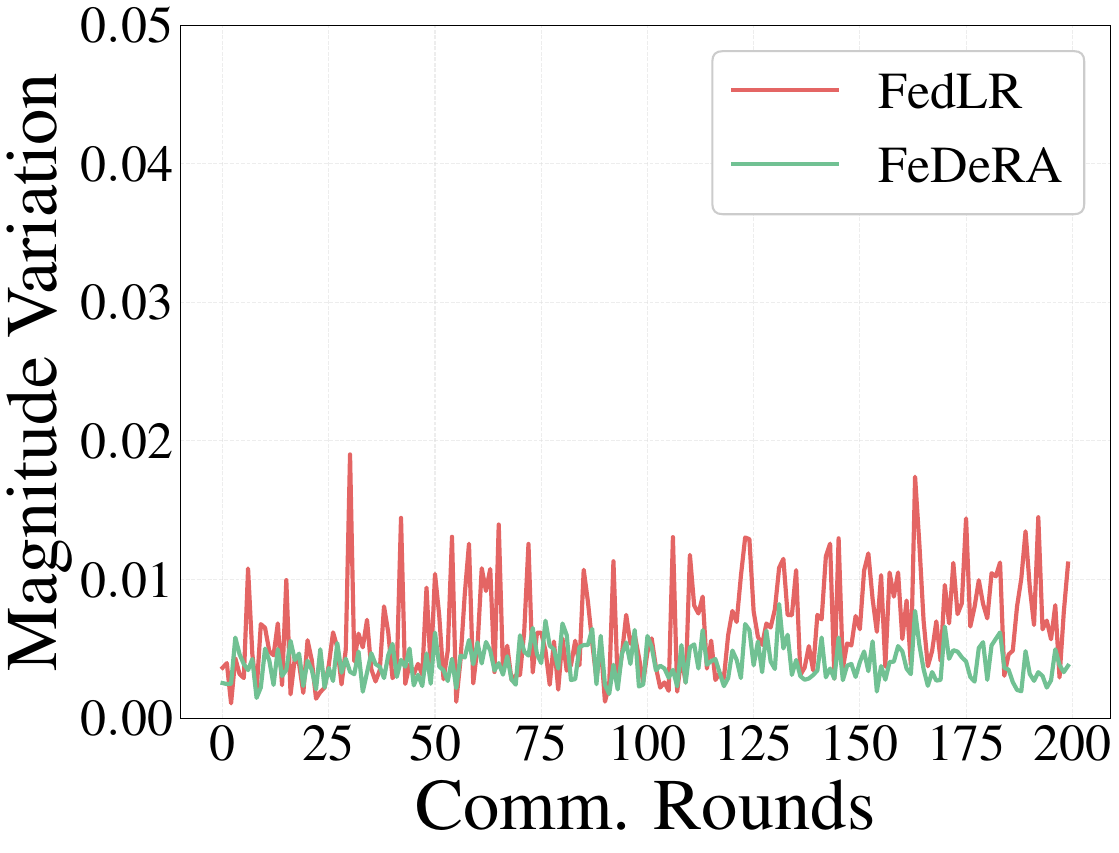}
        \caption{layer.4.q.lora\_A}
    \end{subfigure}\hfill
    \begin{subfigure}{.24\textwidth}
        \includegraphics[width=\linewidth]{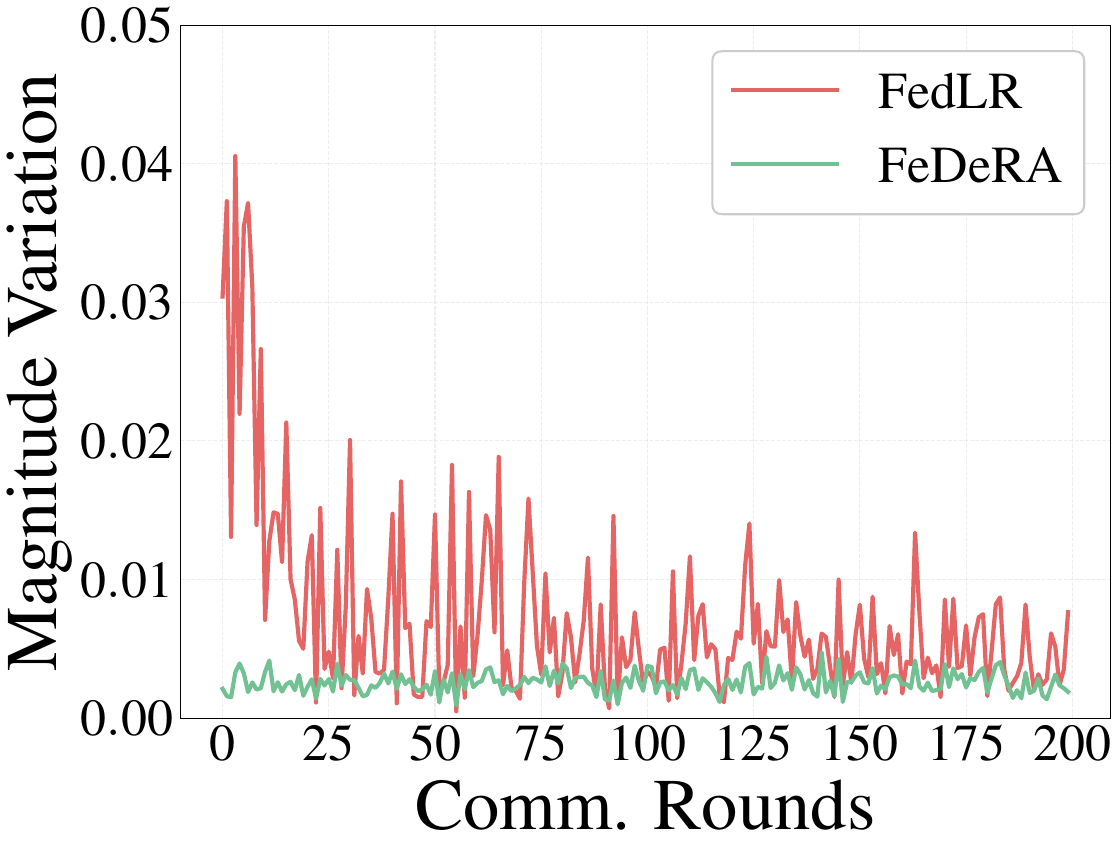}
        \caption{layer.4.v.lora\_B}
    \end{subfigure}

    \caption{Additonal magnitude variation of DistilBERT over 200 Communication Rounds on the 20Newsgroups dataset.}
    \label{fig:additional mag}
\end{figure}

\begin{figure}[!htb]
    \centering
    \begin{subfigure}{.24\textwidth}
        \includegraphics[width=\linewidth]{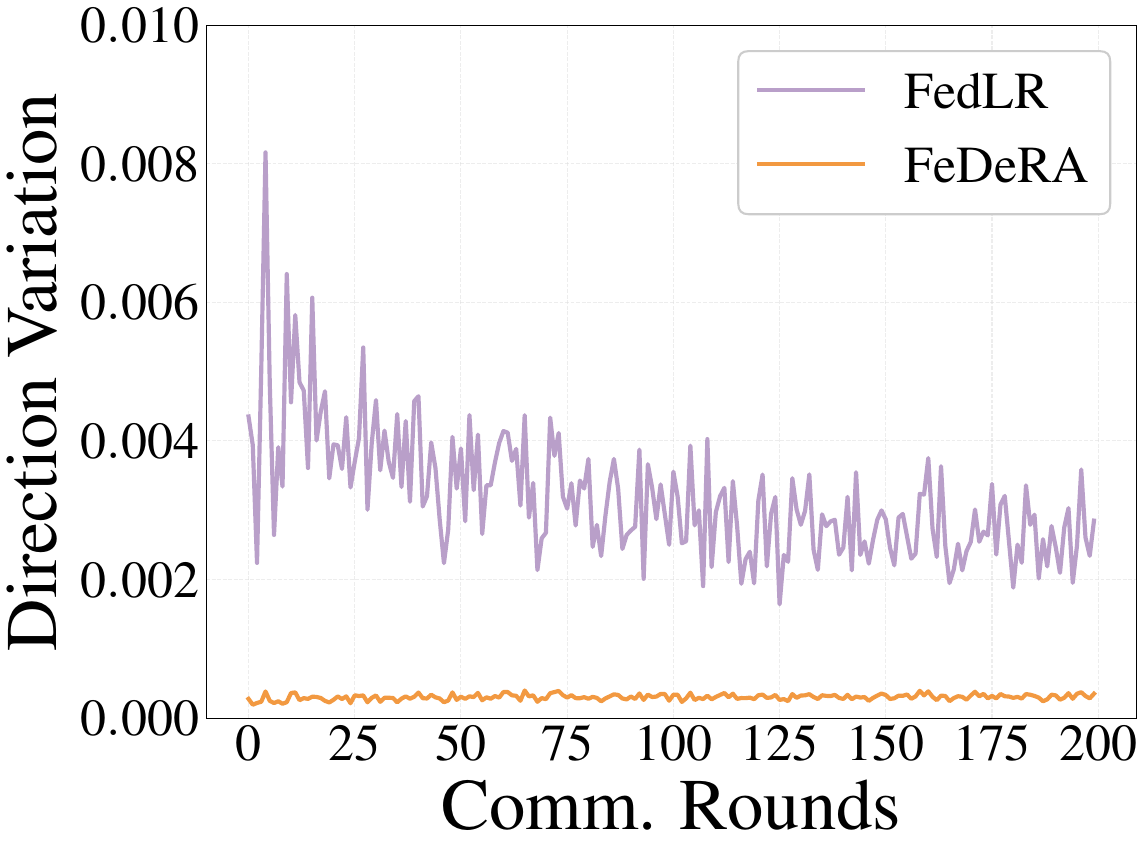}
        \caption{layer.1.q.lora\_A}
    \end{subfigure}\hfill
    \begin{subfigure}{.24\textwidth}
        \includegraphics[width=\linewidth]{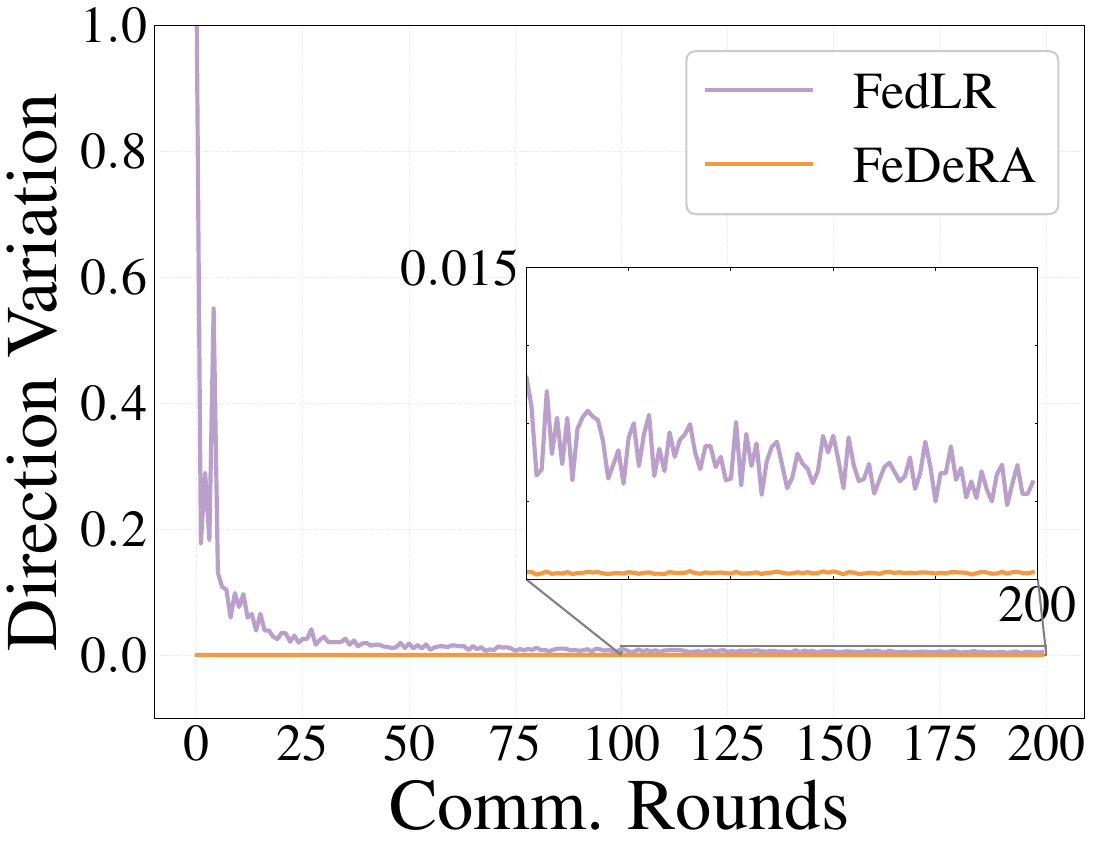}
        \caption{layer.1.q.lora\_B}
    \end{subfigure}\hfill
    \begin{subfigure}{.24\textwidth}
        \includegraphics[width=\linewidth]{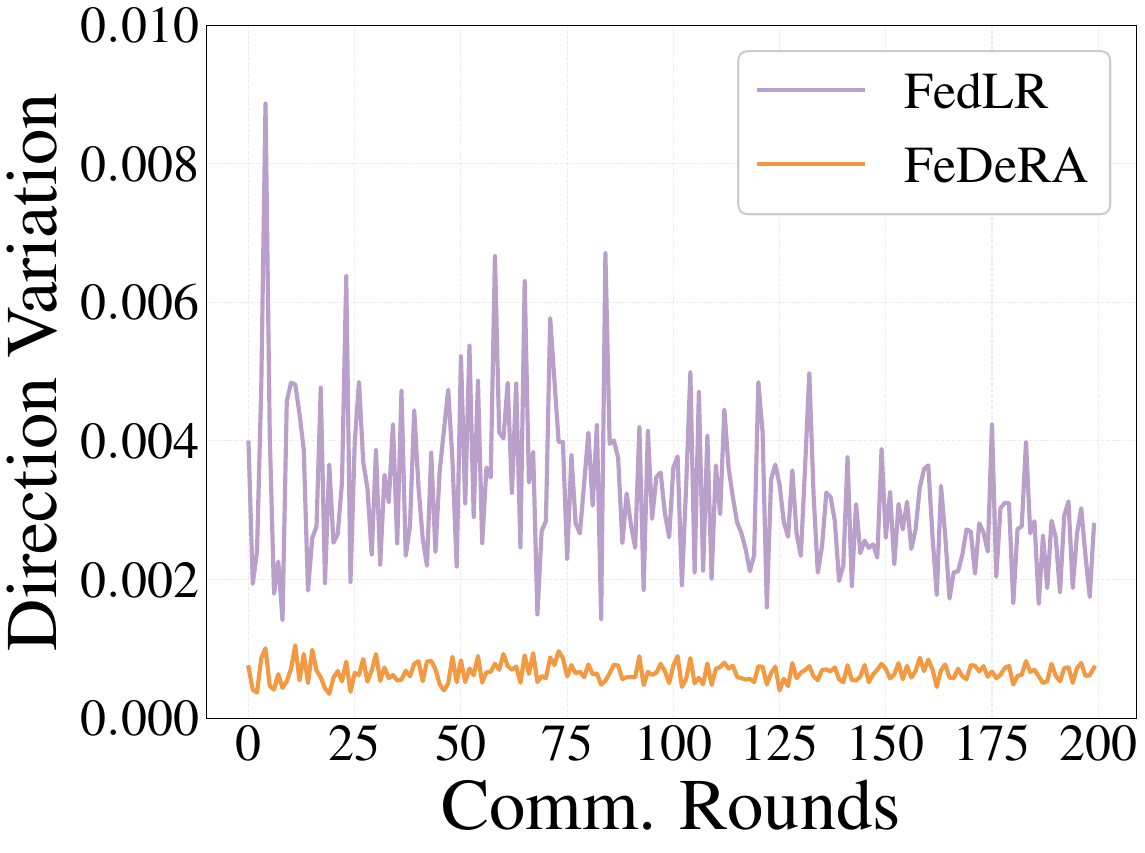}
        \caption{layer.1.q.lora\_A}
    \end{subfigure}\hfill
    \begin{subfigure}{.24\textwidth}
        \includegraphics[width=\linewidth]{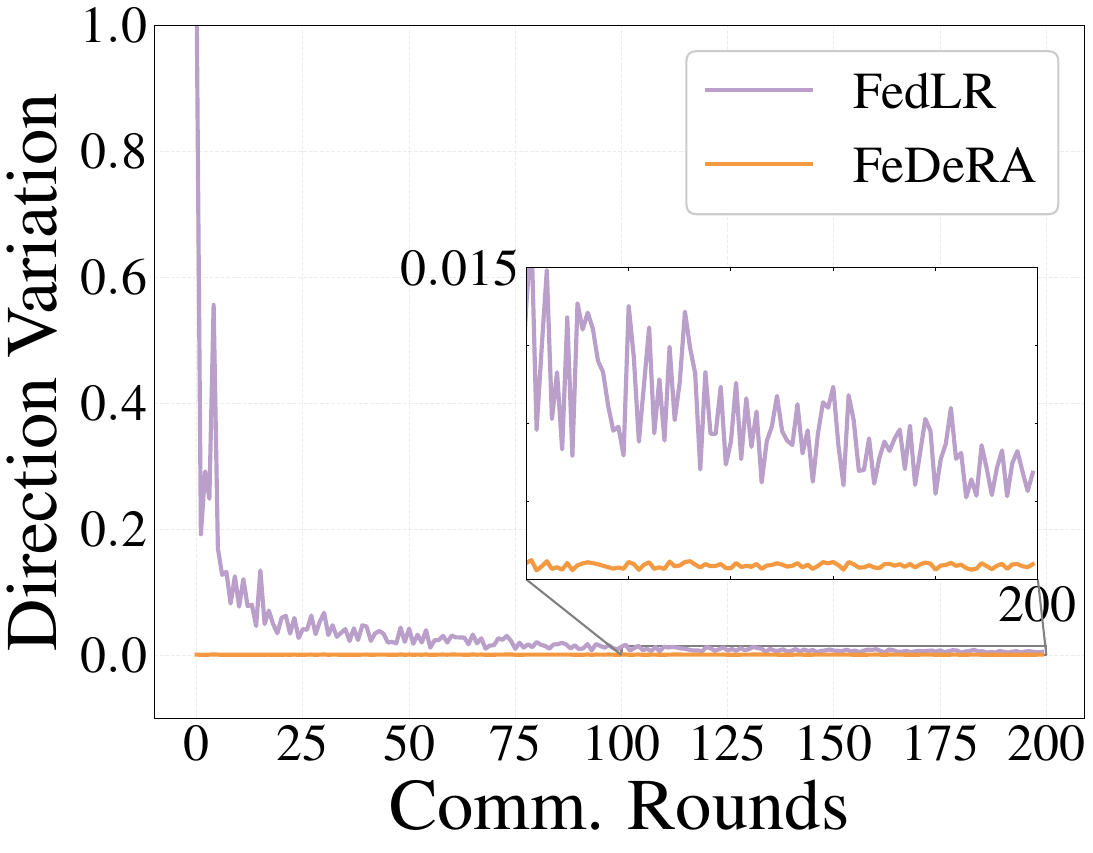}
        \caption{layer.1.v.lora\_B}
    \end{subfigure}

    \begin{subfigure}{.24\textwidth}
        \includegraphics[width=\linewidth]{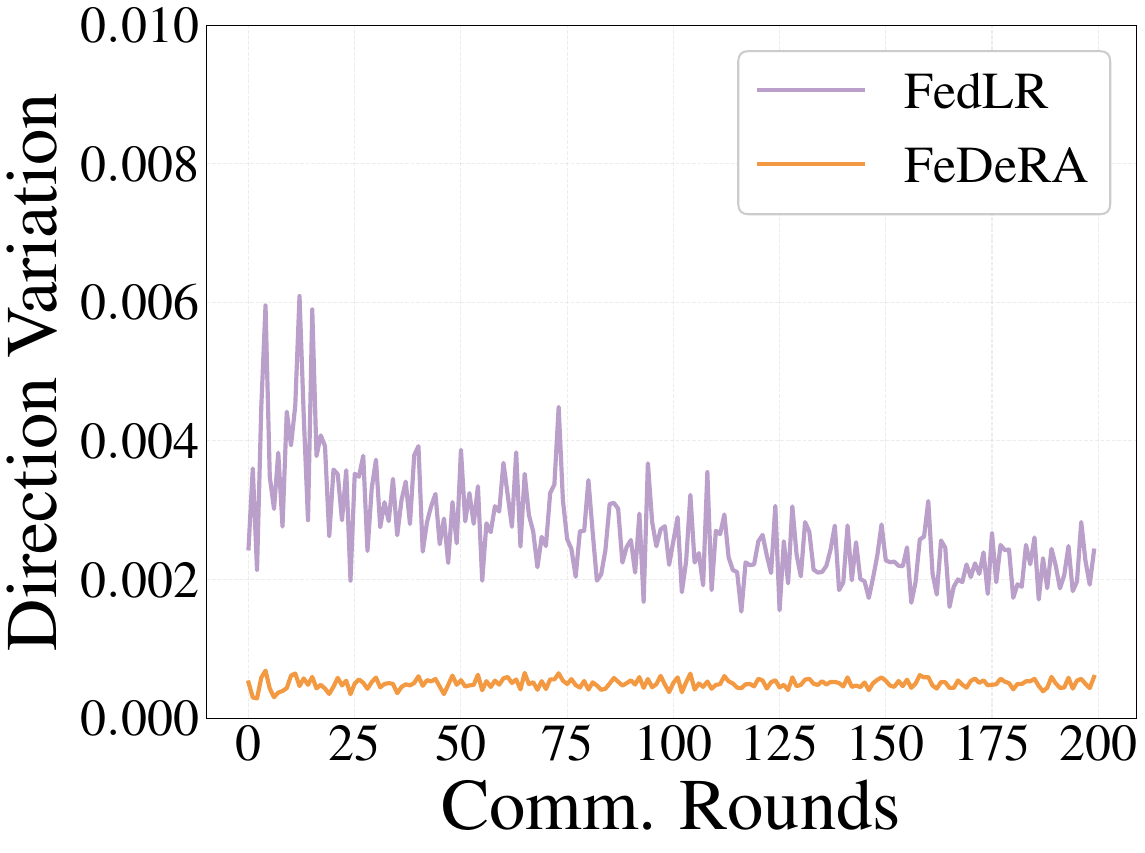}
        \caption{layer.2.q.lora\_A}
    \end{subfigure}\hfill
    \begin{subfigure}{.24\textwidth}
        \includegraphics[width=\linewidth]{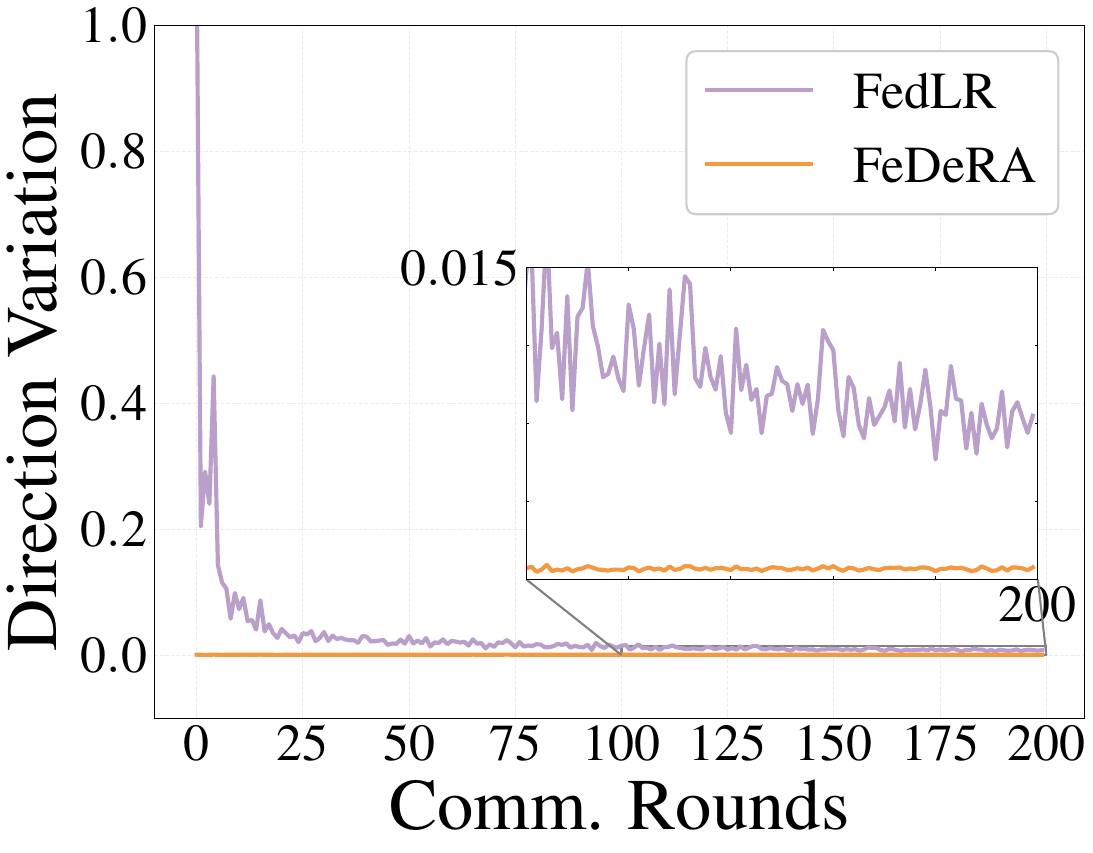}
        \caption{layer.2.q.lora\_B}
    \end{subfigure}\hfill
    \begin{subfigure}{.24\textwidth}
        \includegraphics[width=\linewidth]{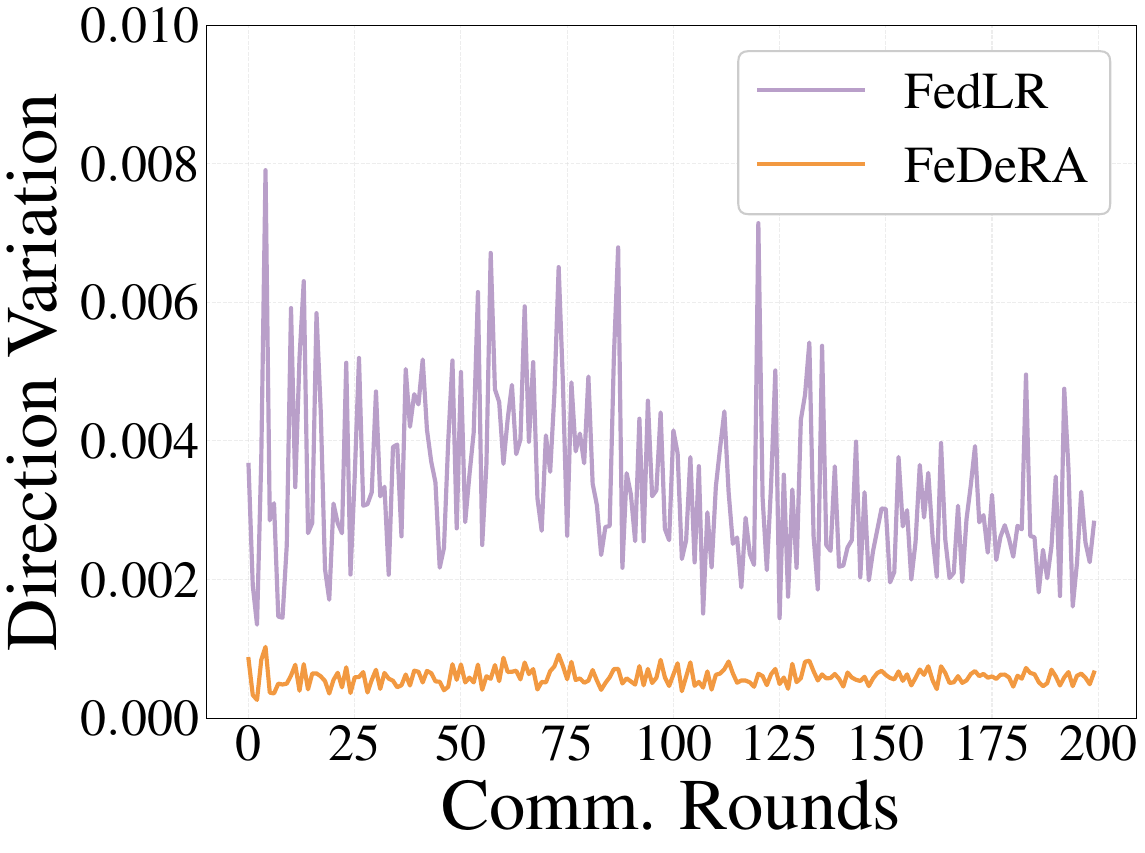}
        \caption{layer.2.q.lora\_A}
    \end{subfigure}\hfill
    \begin{subfigure}{.24\textwidth}
        \includegraphics[width=\linewidth]{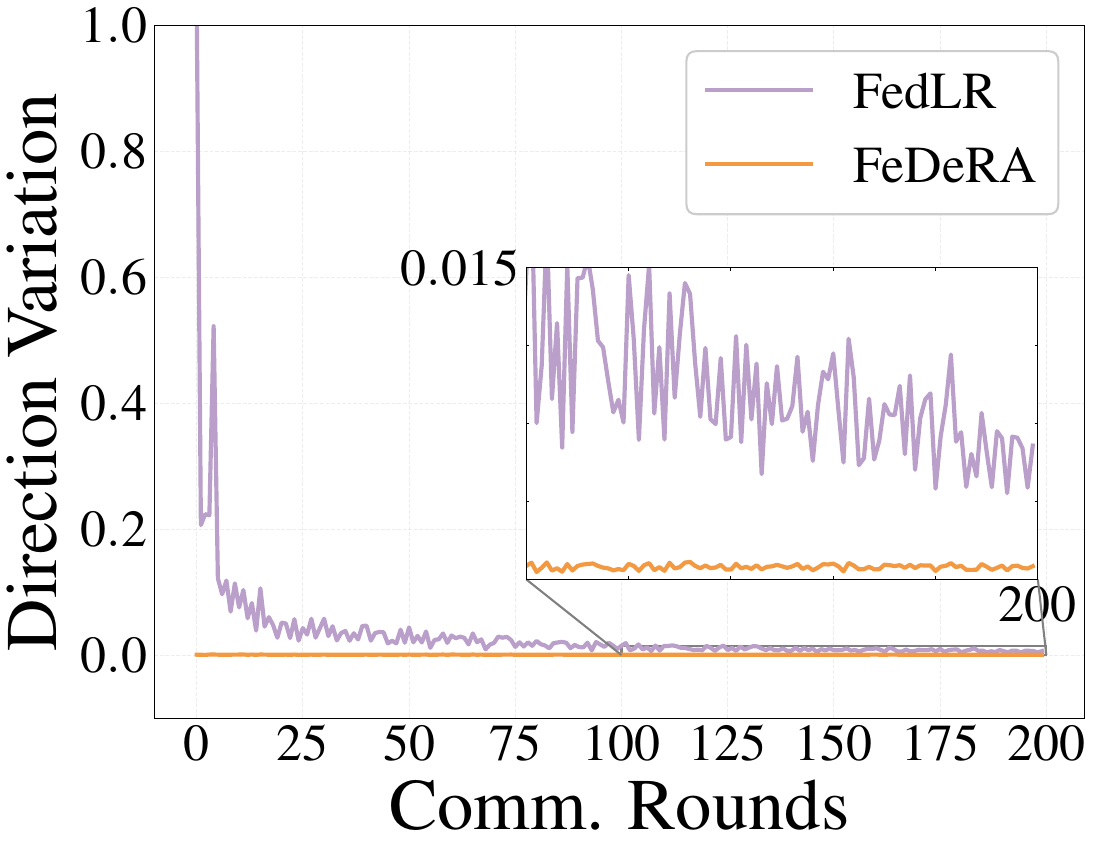}
        \caption{layer.2.v.lora\_B}
    \end{subfigure}

    \begin{subfigure}{.24\textwidth}
        \includegraphics[width=\linewidth]{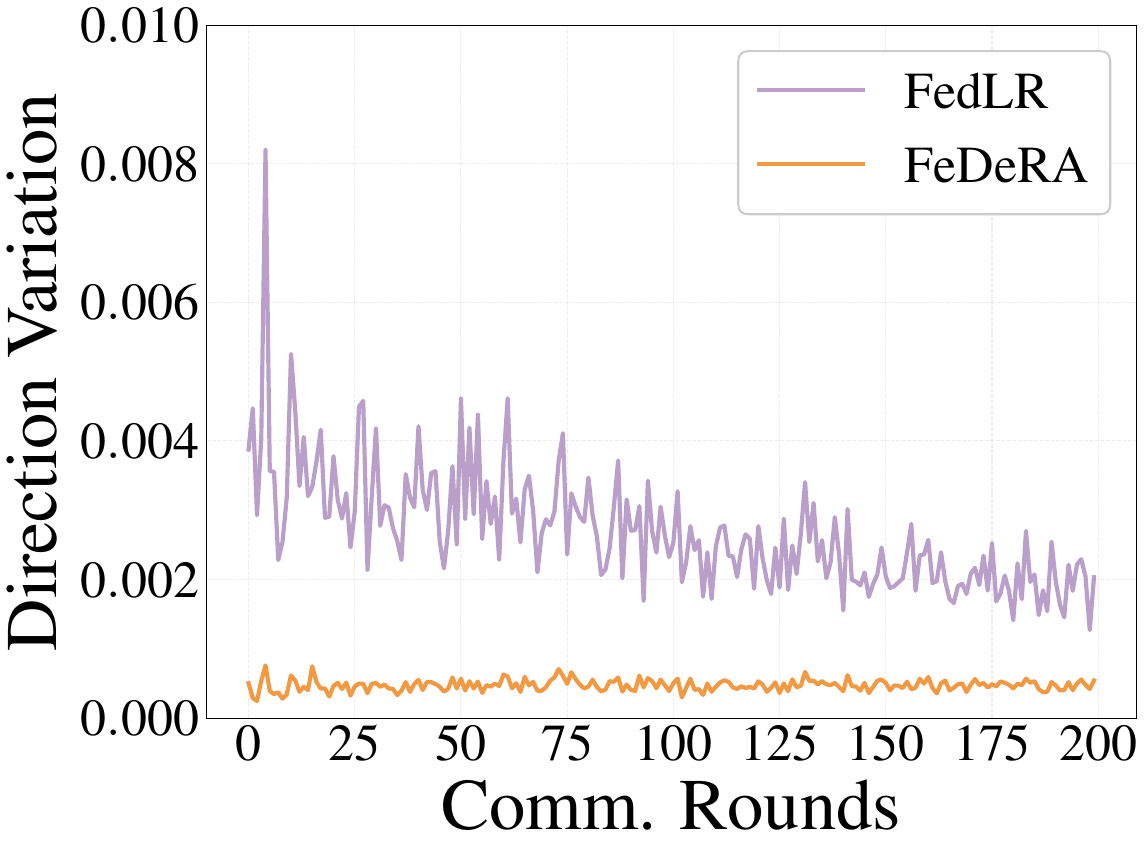}
        \caption{layer.3.q.lora\_A}
    \end{subfigure}\hfill
    \begin{subfigure}{.24\textwidth}
        \includegraphics[width=\linewidth]{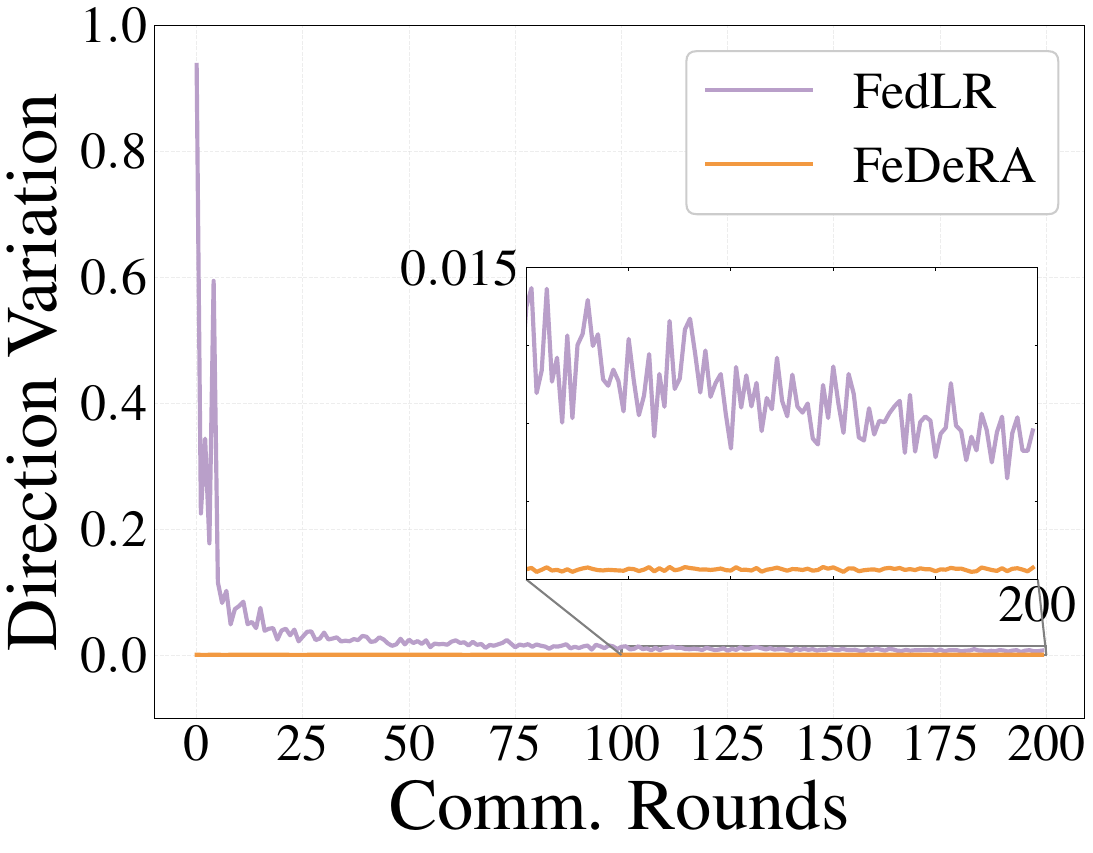}
        \caption{layer.3.q.lora\_B}
    \end{subfigure}\hfill
    \begin{subfigure}{.24\textwidth}
        \includegraphics[width=\linewidth]{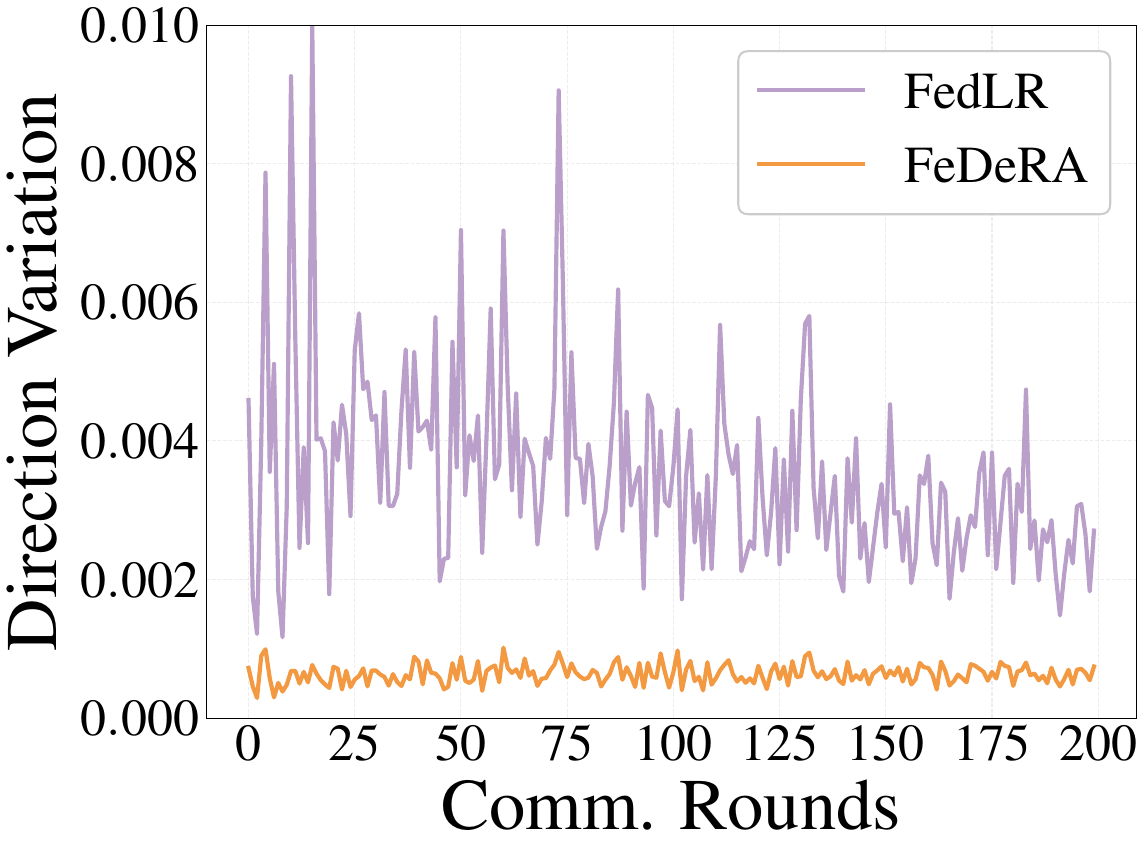}
        \caption{layer.3.q.lora\_A}
    \end{subfigure}\hfill
    \begin{subfigure}{.24\textwidth}
        \includegraphics[width=\linewidth]{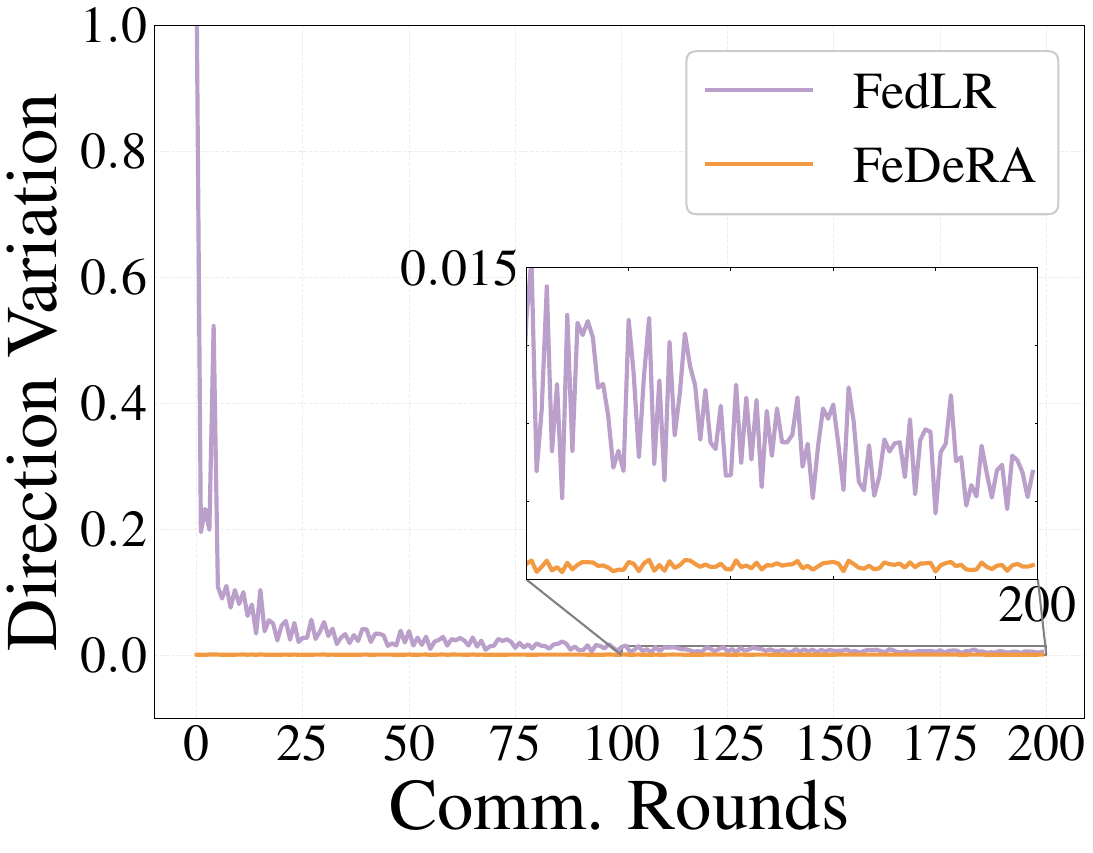}
        \caption{layer.3.v.lora\_B}
    \end{subfigure}

    \begin{subfigure}{.24\textwidth}
        \includegraphics[width=\linewidth]{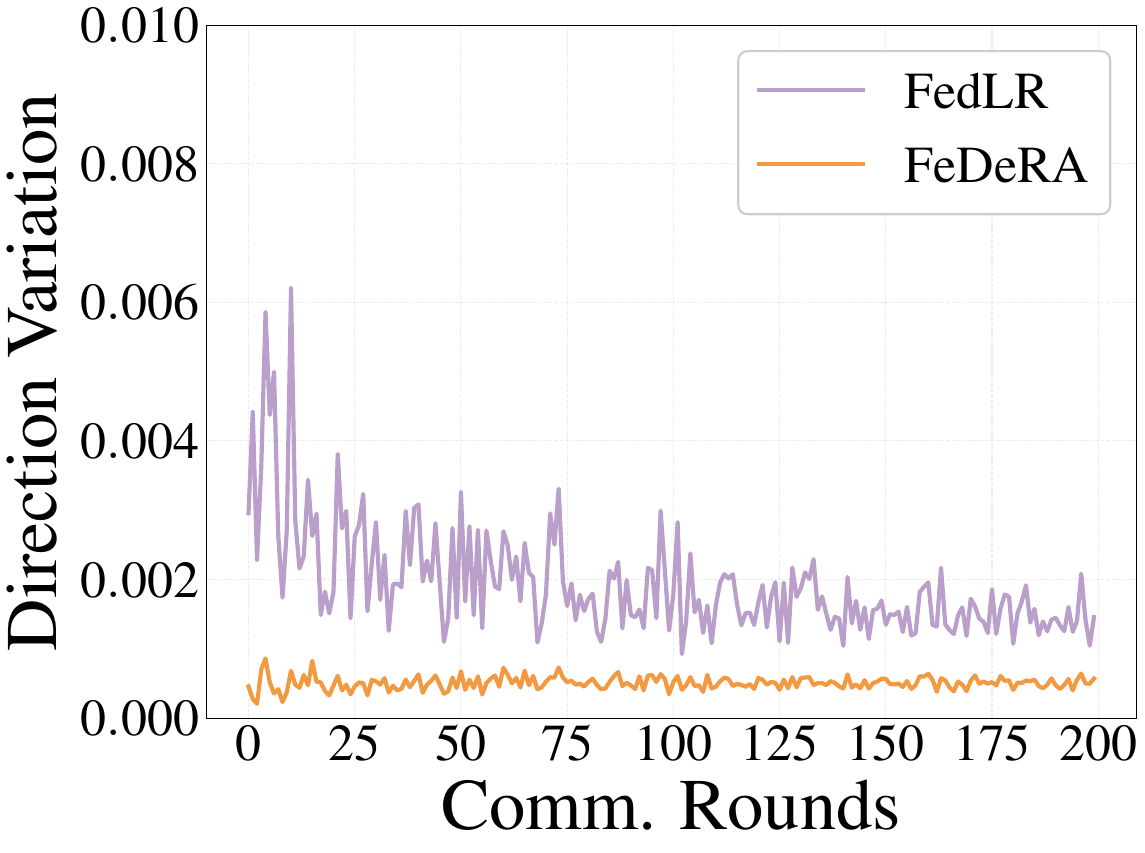}
        \caption{layer.4.q.lora\_A}
    \end{subfigure}\hfill
    \begin{subfigure}{.24\textwidth}
        \includegraphics[width=\linewidth]{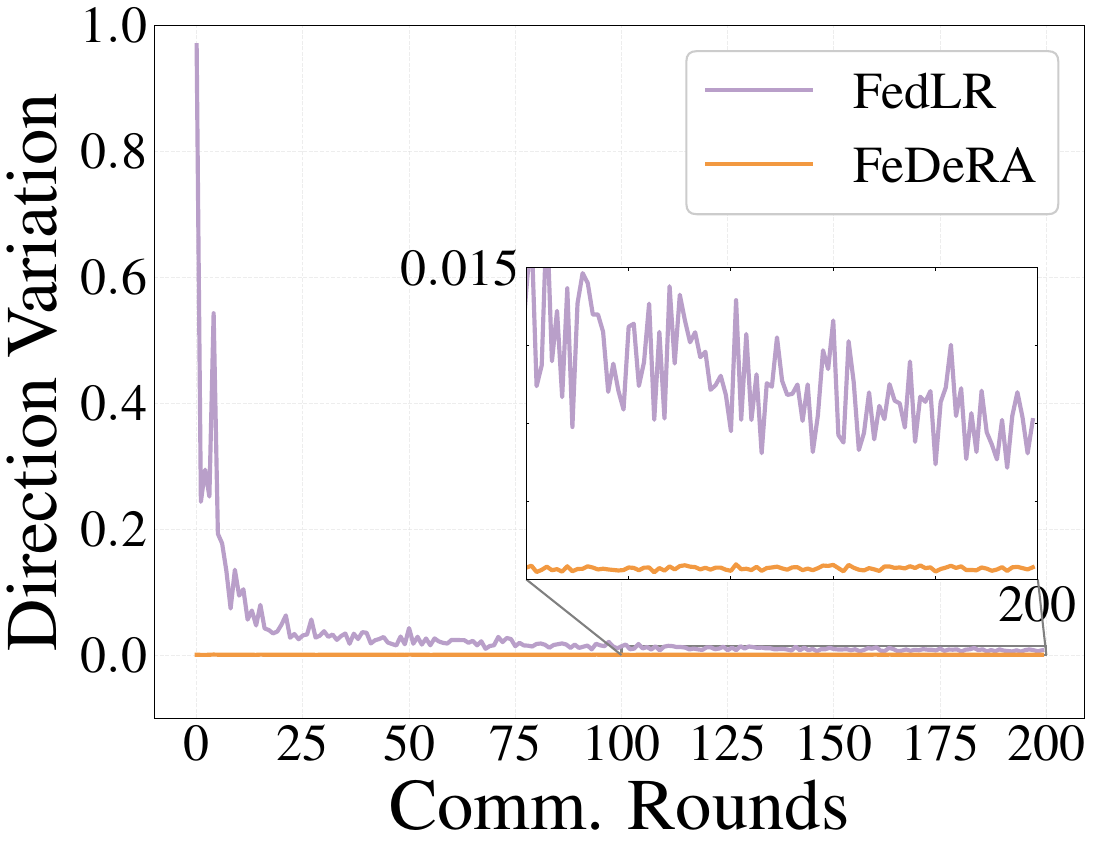}
        \caption{layer.4.q.lora\_B}
    \end{subfigure}\hfill
    \begin{subfigure}{.24\textwidth}
        \includegraphics[width=\linewidth]{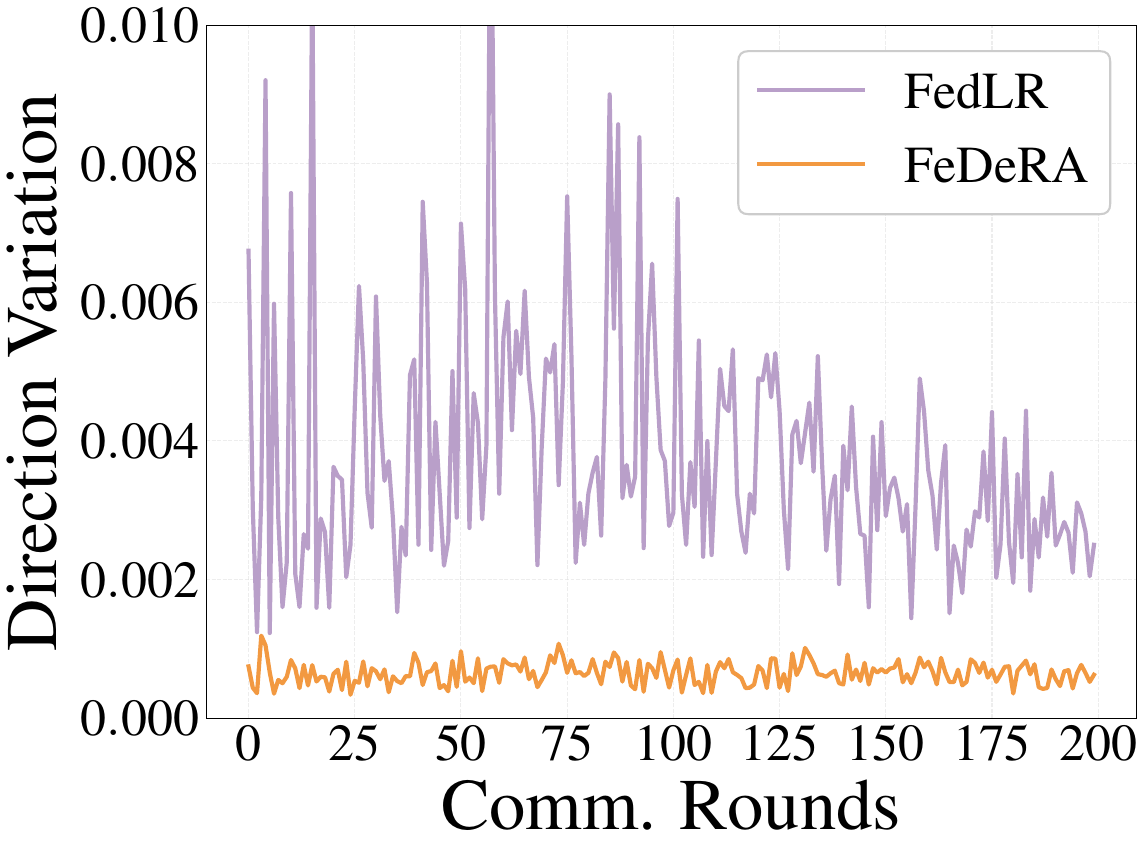}
        \caption{layer.4.q.lora\_A}
    \end{subfigure}\hfill
    \begin{subfigure}{.24\textwidth}
        \includegraphics[width=\linewidth]{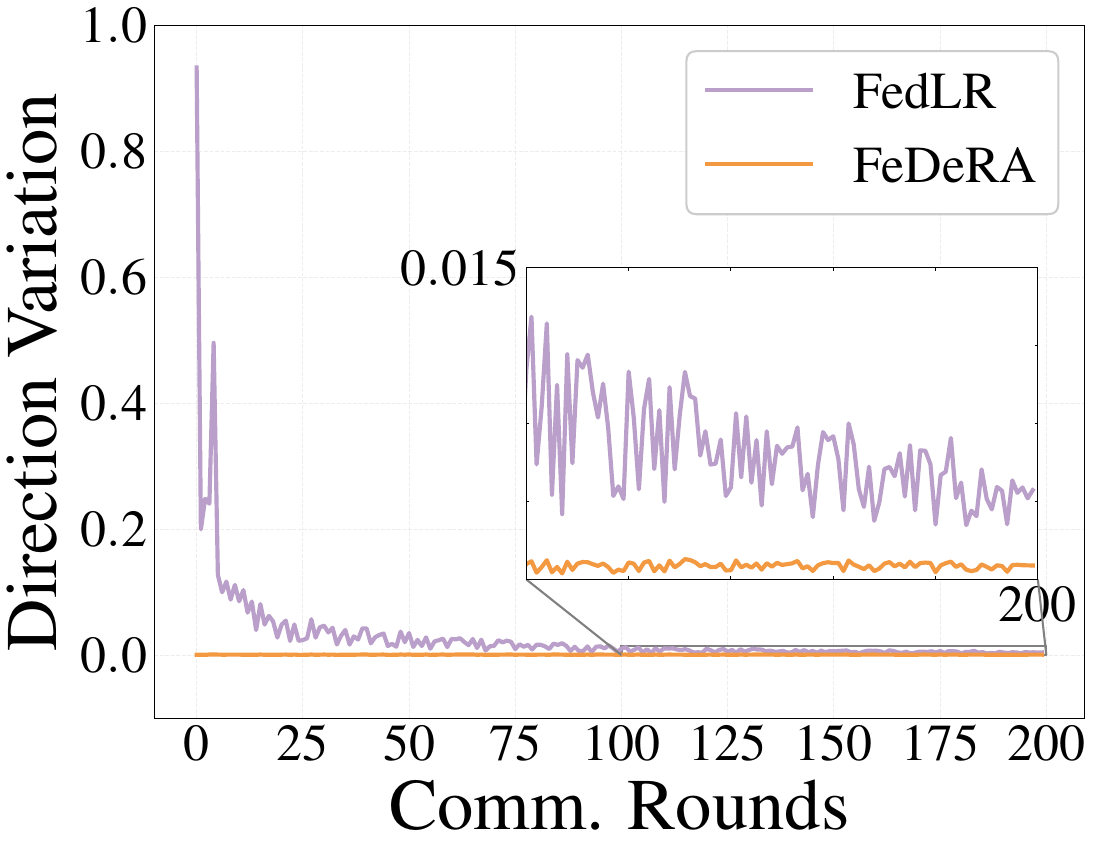}
        \caption{layer.4.v.lora\_B}
    \end{subfigure}

    \caption{Additional direction variation of DistilBERT over 200 communication rounds on the 20Newsgroups dataset.}
    \label{fig:additional dir}
\end{figure}

We present the results of magnitude variation and direction variation not displayed in \autoref{subsec:analysis} here, as illustrated in \autoref{fig:additional mag} and \autoref{fig:additional dir}. From this, we can draw the same conclusion across all layers of DistilBERT, namely that our proposed FeDeRA is more stable in weight updates and consequently easier to converge compared to the original LoRA method.

\end{document}